\documentclass[sigconf]{acmart}
\AtBeginDocument{%
  \providecommand\BibTeX{{%
    \normalfont B\kern-0.5em{\scshape i\kern-0.25em b}\kern-0.8em\TeX}}}

\usepackage{subcaption}
\usepackage{graphicx}
\usepackage{cleveref}
\usepackage{tabularx}
\usepackage{makecell}
\usepackage{tikz}
\usetikzlibrary{shapes.geometric}
\usetikzlibrary{shapes.multipart}
\usetikzlibrary{patterns}
\usetikzlibrary{arrows}
\usetikzlibrary{fit}

\usepackage{multicol}
\usepackage{todonotes}
\usepackage{color, colortbl}
\definecolor{Gray}{gray}{0.9}
\usepackage{tabularx}
\newcolumntype{Y}{>{\raggedleft\arraybackslash}X}
\newcolumntype{Z}{>{\centering\arraybackslash}X}
\newcolumntype{B}{>{\centering\arraybackslash\hsize=1.5\hsize}X}
\newcolumntype{b}{>{\centering\arraybackslash\hsize=1.1\hsize}X}
\newcolumntype{S}{>{\centering\arraybackslash\hsize=.5\hsize}X}
\newcolumntype{s}{>{\centering\arraybackslash\hsize=.7\hsize}X}
\newcolumntype{l}{>{\hsize=.8\hsize}X}

\setcopyright{acmcopyright}
\copyrightyear{2022}
\acmYear{2022}
\acmDOI{XXXXXXX.XXXXXXX}

\acmConference[XKDD '22]{ECML PKDD International Workshop on
eXplainable Knowledge Discovery in Data Mining}{September 19th 2022}{Grenoble, France}
\acmBooktitle{Grenoble 2022, ECML PKDD International Workshop on
eXplainable Knowledge Discovery in Data Mining
 September 19, 2022, Grenoble, France} 
\acmPrice{15.00}
\acmISBN{978-1-4503-XXXX-X/18/06}

\begin{document}

\title{Is Attention Interpretation? A Quantitative Assessment On Sets}

\author{Jonathan Haab}
\email{jonathan.haab@ibm.com}
\orcid{1234-5678-9012}
\author{Nicolas Deutschmann}
\email{deu@zurich.ibm.com}
\author{Mar\'ia Rodr\'iguez Mart\'inez}
\email{mrm@zurich.ibm.com}
\affiliation{%
  \institution{IBM Research Europe}
  \streetaddress{Sa\"umerstrasse 3}
  \city{Z\"urich}
  \country{Switzerland}
  \postcode{8803}
}


\begin{abstract}
  The debate around the interpretability of attention mechanisms is centered on whether attention scores can be used as a proxy for the relative amounts of signal carried by sub-components of data.
  We propose to study the interpretability of attention in the context of set machine learning, where each data point is composed of an unordered collection of instances with a global label.
  For classical multiple-instance-learning problems and simple extensions, there is a well-defined ``importance'' ground truth that can be leveraged to cast interpretation as a binary classification problem, which we can quantitatively evaluate.
  By building synthetic datasets over several data modalities, we perform a systematic assessment of attention-based interpretations. We find that attention distributions are indeed often reflective of the relative importance of individual instances, but that silent failures happen where a model will have high classification performance but attention patterns that do not align with expectations.
  Based on these observations, we propose to use ensembling to minimize the risk of misleading attention-based explanations.
\end{abstract}



\maketitle

\section{Introduction}
\label{sec:introduction}
Attention mechanisms have become a popular tool in multiple areas of machine learning, in particular in natural language processing (NLP) where their introduction significantly increased performance~\cite{Devlin2018BERT}. Attention-based models have also been successful in the context of computer vision~\cite{dosovitskiyImageWorth16x162020} and have in particular been attractive in digital histopathology applications (cancer diagnosis based on stained microscopy images)~\cite{ilseAttentionbasedDeepMultiple2018,redekopAttentionGuidedProstateLesion2021,luDataefficientWeaklySupervised2021,tourniaireAttentionbasedMultipleInstance2021}, where a patch-based approach is particularly well-adapted to analyse the large whole-slide images (WSIs) with corrupting artefacts typically exploited in this field.

Besides the performance gain provided by attention mechanisms in many applications, one of their alluring aspects is the promise of interpretability: attention relies on a dynamically weighted average of representations of data subcomponents, and it feels natural that these weights should be informative of the relative importance of these subcomponents for the final prediction.
This potential interpretability is particularly attractive for biomedical applications, both in a clinical setting and for research. Indeed, insights into automatic diagnostic tools is both a regulatory requirement~\cite{selbstMeaningfulInformationRight2017} and a necessary safeguard to understand and diagnose failure modes for critical decisions~\cite{cluzeauDesignAssurance}. In a biomedical research context, attention-based interpretability could lead to new breakthroughs in understanding the mechanisms that underlie diseases and help find new targets for diagnosis and therapy.

While intuitively promising, there is still no clear understanding of the extent to which attention distributions provide meaningful information about the amount of signal carried by data subcomponents. This has been the object of a debate within the context of NLP~\cite{Jain2019Attention, Wiegreffe2019Attention}, which started at a conceptual level but was then moved forward by experimental assessments~\cite{serranoAttentionInterpretable2019,vashishthAttentionInterpretabilityNLP2019}. These studies found imperfect and task-dependent agreement between attention and other importance attribution metrics, but are limited by the constraints inherent to NLP: the difficulty of building robust ground truths and evaluation metrics for token importance~\cite{madsenPosthocInterpretabilityNeural2022}.

Given the recent interest in using attention in the context of biomedical applications, we propose to study the quality of attention-based explanations of instance importance in a simpler context, where we can conceive synthetic tasks with a well-defined ground truth, therefore allowing more control on the evaluation. Indeed, histopathological (and biomolecular) applications of attention can be characterised as multiple-instance learning (MIL) problems or simple extensions thereof. The goal of this work is to establish synthetic analogies for the MIL-like problems encountered in biomedical applications, with well-defined instance-level importance labels, and to quantitatively assess the quality of attention-based explanations, how frequently they are misleading, and potential solutions.

Our manuscript is organised as follows: we first introduce MIL as an abstract set classification problem, as well as some multi-population extensions. We show how these problems permit a quantitative assessment of instance importance attributions and why they map satisfyingly to some biomedical problems. We then describe the synthetic datasets we constructed as analogies and the attention-based models used to classify them, and conduct experiments to show to which extent attention-based explanations can be trusted. Finally, we argue for an ensemble-based solution to respond to the potential weaknesses of single-model explanations.


\section{Importance attribution as a binary classification task}
\label{sec:import-attr-as}
\subsection{Multiple-instance learning and its extensions}

\subsubsection{Problem Formulation}
\label{sec:problem-formulation}
Multiple-instance learning (MIL) is a classical weakly-supervised learning binary classification problem~\cite{maronFrameworkMultipleInstanceLearning1997,dietterichSolvingMultipleInstance1997,Oquab15} in which data points $X_{i}$ are made of unordered collections of vectors $X_{i} = \left\{x_{i1},\dots,x_{iM_{i}}\right\}$.
The individual vectors $x_{im}$ are referred to as ``instances'', while the data points $X_{i}$ are called bags of instances.
Each instance $x_{im}$ has a binary label $y_{im}\in \left\{1,0\right\}$ (also referred to as positive and negative), which is not available at training time, but defines the label $Y_{i}$ of the bag $X_{i}$ as:
\begin{equation}
  \label{eq:1}
  Y_{i} = \min \left(1, \sum\limits_{m=1}^{M_{i}} y_{im} \right),
\end{equation}
which simply means that $Y_{i}$ is 1 if at least one of the $y_{im}$ is 1, and is 0 otherwise.

This is a formalization of classification problems used in multiple biomedical applications, such as patient diagnosis from histopathology images.  Images are typically processed as collections of patches, of which only a few might contain clinically relevant regions. 
Another interesting medical application of MIL is the classification of tumors using single-cell molecular profiles. In this case, samples are a mixture of healthy and cancerous cell profiles, but only patient-level labels are available.


\subsubsection{Multi-Population MIL}
\label{sec:multi-population-mil}
Inspired by the biological applications of MIL, especially in the context of cancer, we propose to extend MIL to a multi-population setting with non-trivial interactions, which we can formalise as logical problems.

\paragraph{Multi-population AND}
\begin{itemize}
  \item There are three instance populations with three instance labels: $y_{im} \in \left\{0, 1, 2\right\}$. 
  \item Bags have a binary label $Y_{i}$ given as ``the set of $
        \left\{
        y_{im}
        \right\}$ contains 1 AND contains 2''. Namely, $Y_{i}$ is one only if it contains population 1 and 2, but 0 if only one of the two is present. Population 0 is irrelevant. 
\end{itemize}
This problem can model tumours where multiple cell communities can develop and support each other's growth by collaboration: the presence of both cellular communities accelerates disease progression and leads to worse prognosis~\cite{tabassumTumorigenesisItTakes2015}. In this case, population 0 would correspond to uninformative cells such as healthy cells in the tumour microenvironment while populations 1 and 2 would represent two cancerous populations that can collaborate.

\paragraph{Multi-population XOR}
\begin{itemize}
  \item There are three instance populations with three instance labels: $y_{im} \in \left\{0, 1, 2\right\}$.
  \item Bags have a binary label $Y_{i}$ given as ``the set of $
        \left\{
        y_{im}
        \right\}$ contains 1 XOR contains 2'', \textit{i.e.} $Y_{i}$ is one only if it contains population 1 but not 2 or 2 but not 1. Population 0 is irrelevant.
\end{itemize}
This problem can model tumours where two cell communities can co-evolve but reduce their joint fitness such as by increasing drug response when both are present~\cite{millerTumorSubpopulationInteractions1991}.

\subsection{Quantifying key instance attribution}
The simple setting of MIL lends itself  to quantifying the interpretability of importance distributions over bags of instances such as those provided by attention. For standard MIL this is often called key-instance attribution~\cite{liuKeyInstanceDetection2012a}, which amounts to identifying positive instances inside positive bags. When ground truth instance-level labels are known, this can be formulated as a supervised binary classification problem. In this work, we train models with weak, bag-level labels but want to evaluate the attention scores as a prediction score to identify positive label instances.

Of course, we cannot expect attention scores to be well calibrated and to allow their immediate interpretation as a probability score for being "important". We therefore need to be careful with some of the standard classification metrics based on discretising prediction scores, such as accuracy or $F_{1}$. What we require of our attention scores is that they discriminate well between positive and negative instances for some threshold, which can be verified by inspecting the area under the receiver operating characteristic curve (AUROC or AUC). For the sake of clarity, we will refer to the AUC of importance attribution as IAUC, so as not to confuse metrics for the bag-level classification and those for evaluating attention-based explanations.

The multi-population extensions of MIL, \textit{i.e.} AND and XOR don't have canonically defined importances. We propose to extend the ``key instance'' label by assigning it to populations 1 and 2 for both problems defined in \cref{sec:multi-population-mil}, while classifying population 0 as unimportant, since its presence or absence does not impact the bag labels.


\section{Methods}
\label{sec:methods}
\subsection{Attention-Based Deep MIL}
\begin{figure}[h]
\tikzstyle{dataset} = [rectangle, minimum width=0.5cm, minimum height=1.5cm, text centered, draw=black, fill=blue!40]

\tikzstyle{data_h} = [rectangle, minimum width=0.5cm, minimum height=1.cm, text centered, draw=black, fill=blue!40]

\tikzstyle{trap} = [trapezium, minimum width=0.2cm, minimum height=.8cm, text centered, draw=black, fill=green!10]

\tikzstyle{square} = [rectangle, minimum width=0.9cm, minimum height=0.9cm, text centered, draw=black]

\tikzstyle{round} = [circle, text centered, draw=black, fill=orange!20]

\tikzstyle{tria} = [dart, text centered, draw=black, fill=gray!10]

\tikzstyle{arrow} = [thick,->,>=stealth]

\centering
\resizebox{\columnwidth}{!}{
\begin{tikzpicture}[node distance=.1cm]

    \node (data1) [dataset] {};
    \node (data2) [dataset, right of=data1 , yshift=-.2cm] {};
    \node (data3) [dataset, right of=data2 , yshift=-.2cm] {};
    \node (data4) [dataset, right of=data3 , yshift=-.2cm] {};
    \node (x1) [below of=data1, xshift=-0.5cm, yshift=-0.6cm] {\tiny $x_1$};
    \node (x2) [below of=x1, xshift=0.1cm, yshift=-0.1cm] {\tiny $x_2$};
    \node (xm) [below of=x2, xshift=0.1cm, yshift=-0.1cm] {\tiny ...};
    \node (xM) [below of=xm, xshift=0.1cm, yshift=-0.1cm] {\tiny $x_M$};
    
    \node (trap1) [trap,  right of=x2, xshift=2.1cm, yshift=.7cm, rotate=-90] {\rotatebox{90}{$\phi$}};
    \node (trap1_desc) [below of=trap1, xshift=0cm, yshift=-1.5cm,] {Featurizer};
    \node (set) [left of=trap1_desc, xshift=-1.4cm, yshift=0cm] {Set};
    \node (set2) [below of=set, xshift=0cm, yshift=-0.3cm] {\footnotesize input space};
    \draw [arrow] (0.55cm,-0.2cm) to  (trap1);

    \node (h1) [data_h, right of=trap1, xshift=1.6cm, yshift=0.2cm] {};
    \node (h2) [data_h, right of=h1 , yshift=-.2cm] {};
    \node (h3) [data_h, right of=h2 , yshift=-.2cm] {};
    \node (h4) [data_h, right of=h3 , yshift=-.2cm] {};
    \node (phix1) [below of=h1, xshift=-0.65cm, yshift=-0.3cm] {\tiny $\phi(x_1)$};
    \node (phix2) [below of=phix1, xshift=0.05cm, yshift=-0.1cm] {\tiny $\phi(x_2)$};
    \node (phixm) [below of=phix2, xshift=0.05cm, yshift=-0.1cm] {\tiny ...};
    \node (phixM) [below of=phixm, xshift=0.05cm, yshift=-0.1cm] {\tiny $\phi(x_M)$};
    \node (set3) [right of=trap1_desc, xshift=1.9cm, yshift=0cm] {Set};
    \node (set4) [below of=set3, xshift=0cm, yshift=-0.3cm] {\footnotesize embedded space};
    \draw [arrow] (trap1) to  (3.25cm,-0.2cm);
    
    \node (circle1) [round, right of=h2, xshift=1.5cm, yshift=0cm] {$\alpha$};
    \node (circle1_desc) [right of=set3, xshift=1.3cm, yshift=0cm] {Aggregator};
    \draw [arrow]  (4.05cm,-0.2cm) to  (circle1);
    
    \node (bag_rep) [data_h, right of=circle1, xshift=1.3cm, yshift=0cm] {};
    \node (bag_rep_desc) [right of=circle1_desc, xshift=1.3cm, yshift=0cm] {Set};
    \node (bag_rep_desc2) [below of=bag_rep_desc, xshift=0.cm, yshift=-0.3cm] {\footnotesize representation};

    \draw [arrow]  (circle1) to  (bag_rep);
    
    \node (triangle1) [tria, right of=bag_rep, xshift=1.3cm, yshift=0cm] {$\rho$};
    \node (triangle1_desc) [right of=bag_rep_desc, xshift=1.4cm, yshift=0cm] {Classifier};
    \draw [arrow] (bag_rep) to (triangle1);

\end{tikzpicture}
}
\caption{Deep-Sets-like permutation invariant networks map bags of instances $x_i$ to bags of latent representations $\phi(x_i)$ which are then aggregated as a set representation. This set representation is processed by a classifier to obtain a prediction  $\rho$. In all experiments in this paper, the aggregator $\alpha$ is the attention mechanism described in~\cref{eq:2}.
    \label{fig:deepsets_architecture}}
\end{figure}
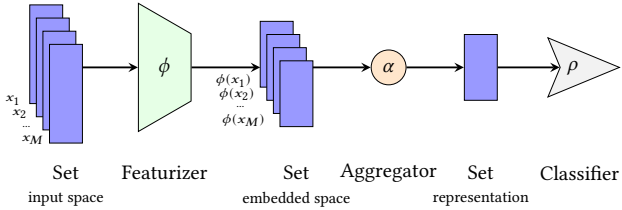

Permutation-invariant models are best-suited to handle MIL tasks as they introduce an inductive bias tailored to sets of instances where order does not matter.
To this end, the Deep Sets architecture~\cite{zaheerDeepSets2018} was designed to produce an independent latent representation of each instance, which are then aggregated with a permutation invariant function such as the mean. The aggregated latent representation is further processed to produce a bag label, as shown in~\cref{fig:deepsets_architecture}.

Attention-based aggregation is another permutation-invariant operation that dynamically performs weighted averages using the attention scores. This was shown to improve performance and  provide insights into the data through the assigned weights~\cite{ilseAttentionbasedDeepMultiple2018}.
With attention-based aggregation, a data point $X =
\left\{
 x_{1},\dots,x_{M}
\right\} $ is mapped to a prediction $y$ as follows:
\begin{gather}
  \label{eq:2}
  z_{i} = \phi(x_{i}), \quad
  Z = \sum\limits_{m=1}^{M} a_{m} z_{m},\nonumber\\
  y =\rho(Z),
\end{gather}
where $\phi$ and $\rho$ are approximated by neural networks and $a_m$ is the attention scores of instance  $x_m$, defined as:
\begin{align}
    a_m & = \frac{\exp \{ \mathbf{w}^\top \tanh [ \mathbf{V} \phi(x_m)^\top ] \} }{\sum\limits_{j=1}^M \exp \{ \mathbf{w}^\top \tanh [ \mathbf{V} \phi(x_j)^\top ] \}},
    \label{eq_attention}
\end{align}
and, $\mathbf{V} \in {\rm I\!R}^{L \times K}$ and $\mathbf{w} \in {\rm I\!R}^{L \times 1}$ are trainable parameters. Notice that as $\sum\limits_{j=1}^M a_m = 1$,  Eq.~\ref{eq_attention} defines normalized discrete weights over the instances.

\subsection{Synthetic Datasets}

We generate synthetic datasets with well-defined ground truth instance labels using three data modalities.
These instance labels were kept hidden from the model at all times and only used to evaluate the performance of the attention attribution.

The first type of datasets, referred to as \textit{Gaussian MIL}, \textit{Gaussian AND} or \textit{Gaussian XOR}, was built by sampling instances from normal distributions, $\mathcal{N}(\mu, \sigma=1)$ with $\mu \in \mathbb{R}^4$. Populations 0,1 and 2 correspond to three choices of $\mu$: $\mu_0 = (0, 0, 0, 0)^\top$, $\mu_1 = (1, 1, 1, 1)^\top$ and $\mu_2 = (-1, 1, 1, 1)^\top$.

The second type of datasets trades 4-dimensional vectors for $28\times28$ pixels images of MNIST handwritten digits and are referred to as \textit{MNIST MIL}, \textit{MNIST AND} or \textit{MNIST XOR}.
The bags defined by first specifying the list of digits allowed in each population and then randomly sampling images of the specified digits from the original MNIST dataset~\cite{lecunMnistDatabaseHandwritten2005}.
Images of the digit "3" are given the instance label 1 in every problem while images of the digit "9" have instance label 2 in the XOR and AND cases. Any other digits are considered unimportant (label 0).

The last data modality mimics data produced by single-cell proteomics experiments. We used experimental single-cell mass-cytometry (CyTOF) measurements from breast cancer tumours~\cite{wagnerSingleCellAtlasTumor2019} to produce pseudo-samples by randomly selecting  epithelial cells. Each cell is characterised by 27 protein abundance measurements  from a panel of markers chosen for cell phenotyping. The work that collected and  published these data~\cite{wagnerSingleCellAtlasTumor2019} grouped cells into 9 super-clusters of functionally and phenotypically distinct cells, including 
7 clusters of luminal cells and two clusters of basal cells (B1 and B2). Basal cells  are indicative of more dangerous tumours, in particular,  super-cluster B2  was found to be strongly associated with triple-negative tumours~\cite{eliasTripleNegativeBreastCancer2010}. We therefore define the \textit{CyTOF MIL}, \textit{CyTOF AND} or \textit{CyTOF XOR} with populations 0, 1, and 2 respectively corresponding to luminal cells, B2 cells and B1 cells.

In all settings, we generate bags of 250 instances, which are drawn from bi- or trinomial distributions of populations 0, 1 and~2. In the MIL setting, we use a binomial distribution with equi-probable outcomes while in the multi-population settings we use a trinomial distribution where population 0 has probability $0.4$ and populations 1 and 2 have probability $0.3$. While not described in this paper, we have confirmed that our results are quite robust to changes in these parameters except for extreme cases (extremely low fractions of some population or very small bags).

\section{Results}
\label{sec:results}
The basis of our analysis is a hyperparameter search for each task and data modality. We perform a grid search through possible configurations for our models and train each configuration with five random initialisations. Models are then ranked and selected on the basis of their performance on a validation set, and evaluated on a separate test set. More details on the hyperparameter search are provided in~\cref{sec:hyperp-search}.

\subsection{Models with high accuracy can have poorly behaved attention}
\label{sec:models-with-high}

\begin{figure}
\centering
\begin{subfigure}{0.23\textwidth}
\centering
\includegraphics[width=\textwidth]{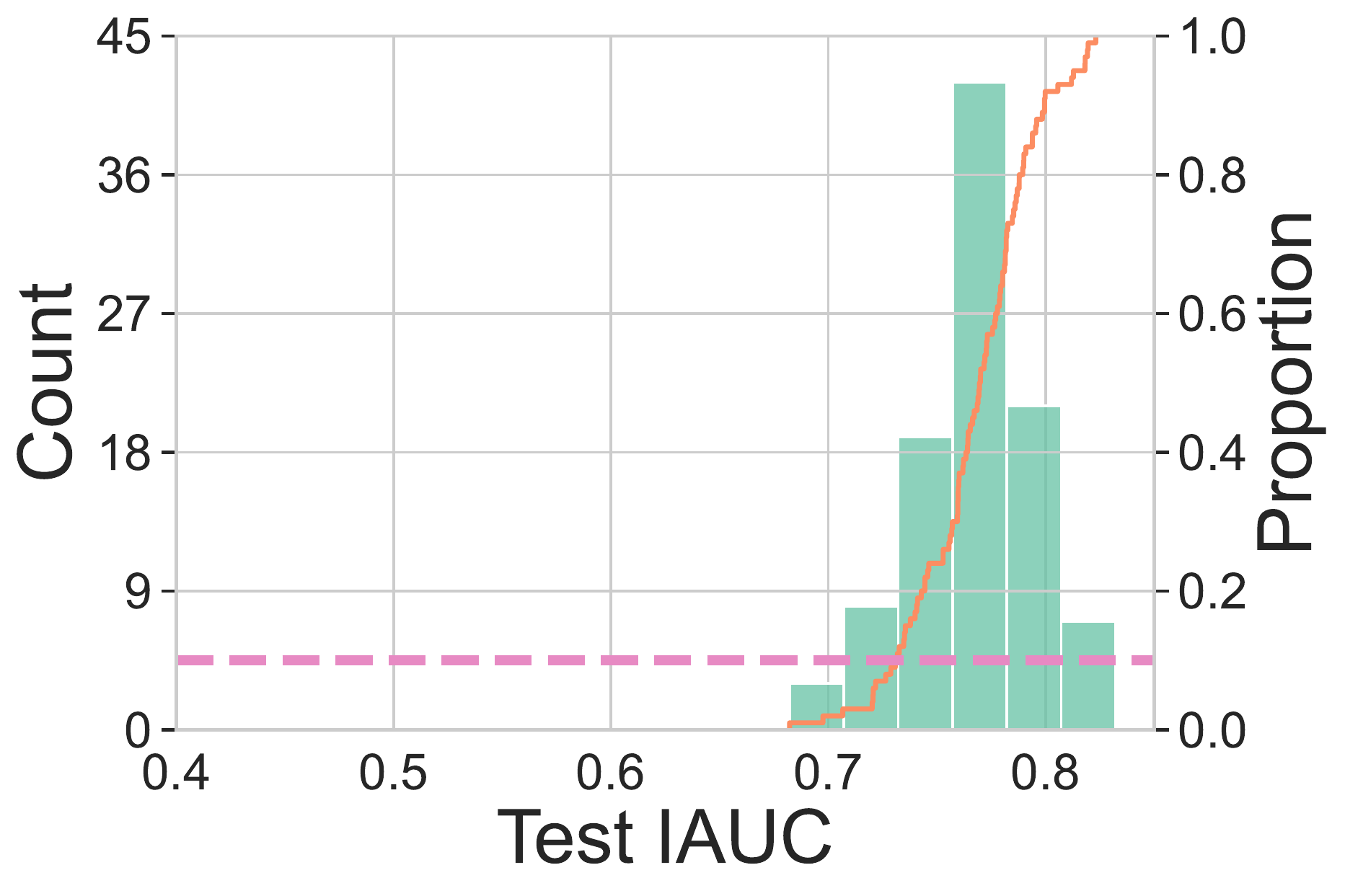}
\caption{}
\end{subfigure}
\begin{subfigure}{0.23\textwidth}
\centering
\includegraphics[width=\textwidth]{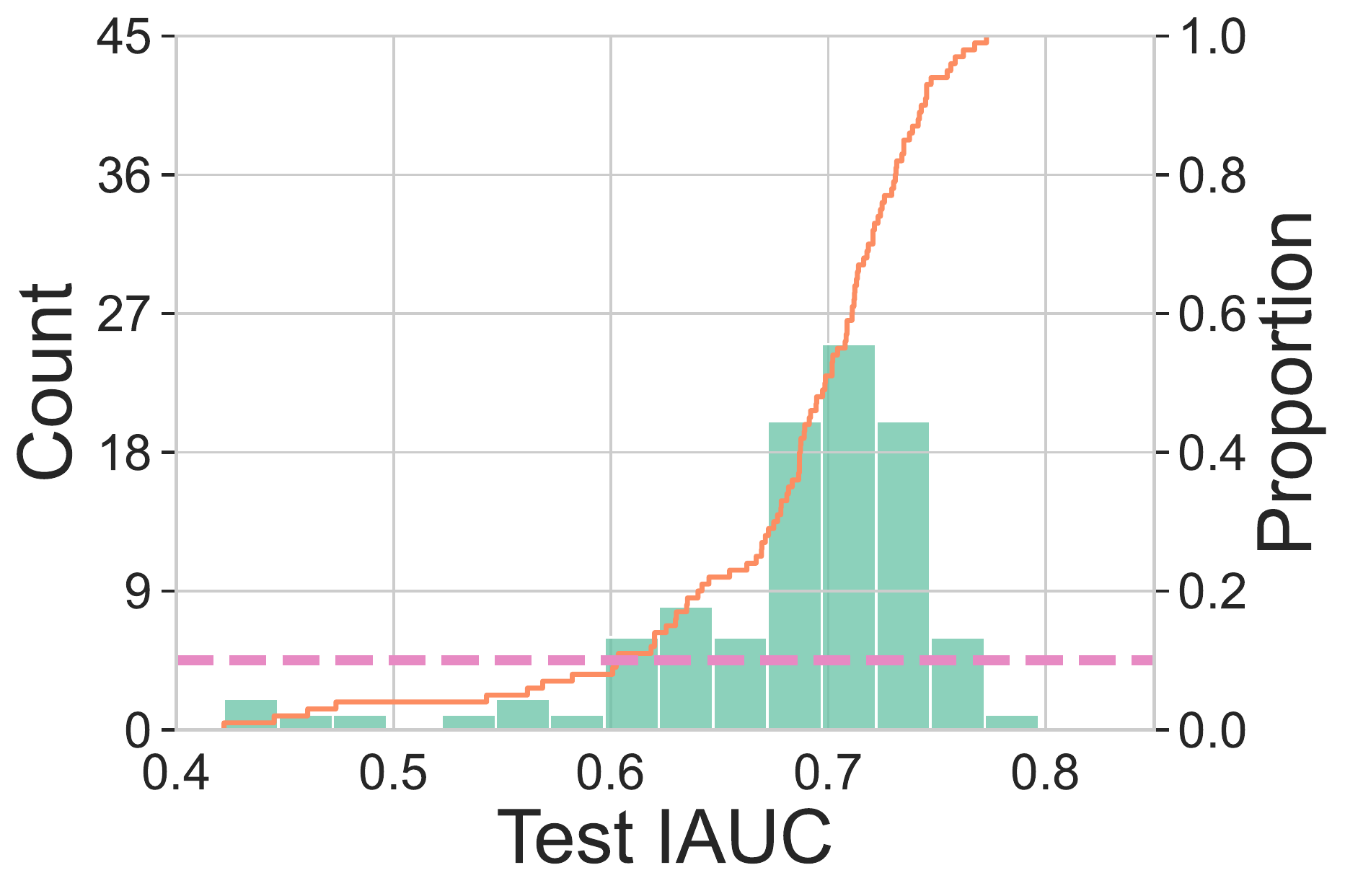}
\caption{}
\end{subfigure}
\caption[]{\textbf{(a)} Configuration with stable IAUC. \textbf{(b)} Configuration with significant fraction of low IAUC. Both configurations were trained on the Gaussian MIL setting. The left Y-axis refers to the histogram (in green), while the right Y-axis refers to the  cumulative frequency plot (orange line). The magenta line is a guide for the eye showing the 10\% threshold used to define bad configurations in~\cref{tab:auroc_outliers}.  \label{fig:mil_auroc_dist}  }
\end{figure}

To reproduce the process of selecting models in a setting where instance-level importances are unknown, we select five candidate model configurations from our hyperparameter search based on their validation accuracy and evaluate the interpretability of their attention distributions.
We train 100 repetitions of each of those top five configurations with different random seeds and evaluate how well the attention scores separate negative from positive instances in bags with a positive label. As we show in \cref{fig:mil_auroc_dist}, some configurations have narrow distributions of IAUC centred around a reasonable value (0.75), meaning that all model realisations provide meaningful interpretations through their attention distributions while others have a non-negligible fraction of outliers with a very poor identification of important instances (IAUC around 0.5).

This pattern repeats over all problems and data modalities we evaluated. We summarise the results of our analysis in \cref{tab:auroc_outliers}, where we report the mean test IAUC across all configurations and the number of ``bad'' configurations, defined as those  having  10\% or higher fraction of realisations with an IAUC less than 0.65. Detailed results with all IAUC distributions are available in \cref{sec:auroc-distr-top}.

\begin{table}
  \centering
\begin{tabularx}{\linewidth}{lZZZ}
\toprule
\thead{Data} & \thead{Problem} & \thead{Mean IAUC} & \thead{\# bad config.} \\
\midrule
{} & MIL & 0.75 & 1/5 \\
\rowcolor{Gray}\cellcolor{white} Gaussian & AND & 0.59 & 5/5 \\
{} & XOR & 0.72 & 5/5 \\
\cmidrule(lr){2-4}
{} & MIL & 0.80 & 0/5 \\
\rowcolor{Gray}\cellcolor{white} MNIST & AND & 0.69 & 2/5 \\
{} & XOR & 0.84 & 2/5 \\
\cmidrule(lr){2-4}
{} & MIL & 0.76 & 1/5 \\
\rowcolor{Gray}\cellcolor{white} CyTOF & AND & 0.75 & 3/5 \\
{} & XOR & 0.77 & 3/5 \\

\bottomrule
\end{tabularx}
\caption{Evaluation of attention explanations performances. Multi-population problems tend to have more bad configurations than MIL, which can still have poor explanations. In general, AND problems also have an overal lower IAUC.\label{tab:auroc_outliers}}
\end{table}

\begin{figure}
  \centering
  Gaussian Data (PCA)\\
  \centering
  \includegraphics[width=0.45\textwidth]{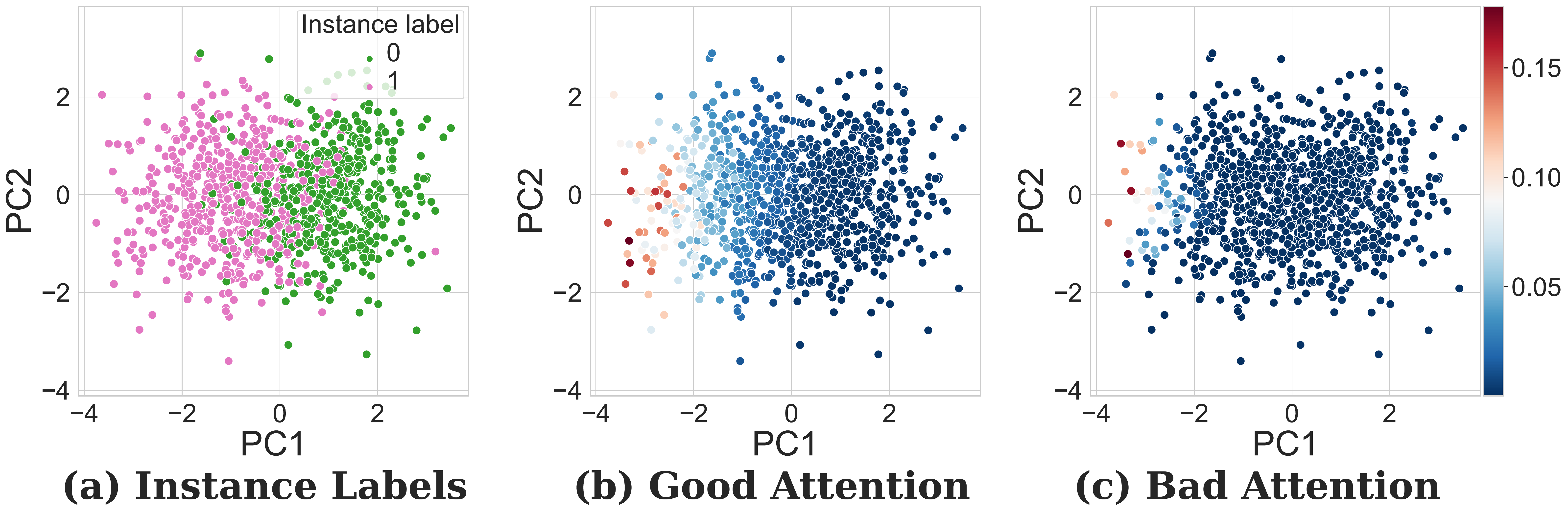}\\
  \vspace{1em}
  CyTOF Data (TSNE)\\
  \centering
  \includegraphics[width=0.45\textwidth]{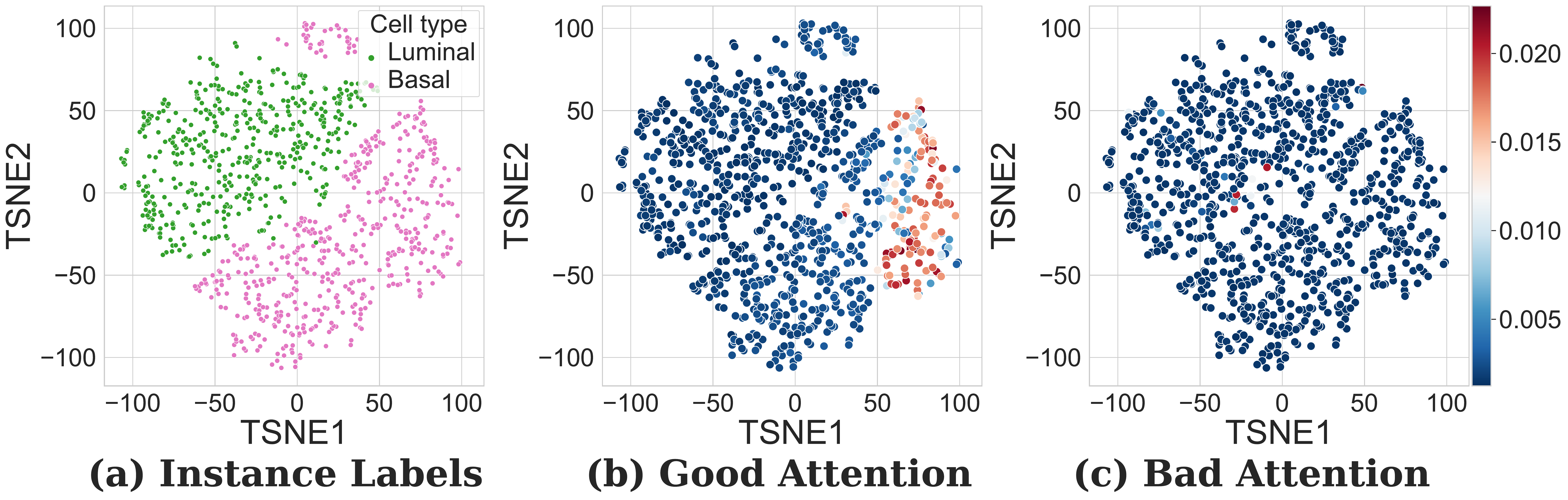}\\
  \caption{Low-dimensional projections of MIL data with showing attention scores for an example of a "good" and a "bad" model, as well as the instance labels shown for reference.}
  \label{fig:instance_attention}
\end{figure}

To further illustrate the difference in behavior between ``good'' and ``bad'' models, we show low-dimensional representations of both of our numerical datasets (Gaussian and CyTOF) in~\cref{fig:instance_attention}, where the attention distributions are visible. ``Good'' models have an essentially constant attention over unimportant instances and show a sharp gradient on positive instances moving away from the class boundary, while ``bad'' models essentially have uniform attention over much of the dataset, with the exception of a small minority of the data, which is not necessarily a subset of the positive instances.

\subsection{Repetitions of the same model have little correlation between performance and interpretability}
\label{sec:repet-same-model}
The stochasticity of training multiple neural networks with the same hyperparameters leads to the variability in the quality of the explanations provided by their attention maps. Of course, this stochasticity also leads to variability in the validation and test performance of these models. It is therefore natural to investigate whether, for a fixed configuration, there is a correlation between the classification performance at the bag level and the quality of the attention-based explanations. This analysis might provide a way to weed out problematic models at the validation stage.

\begin{figure}
  \centering
  \includegraphics[width=0.4\textwidth]{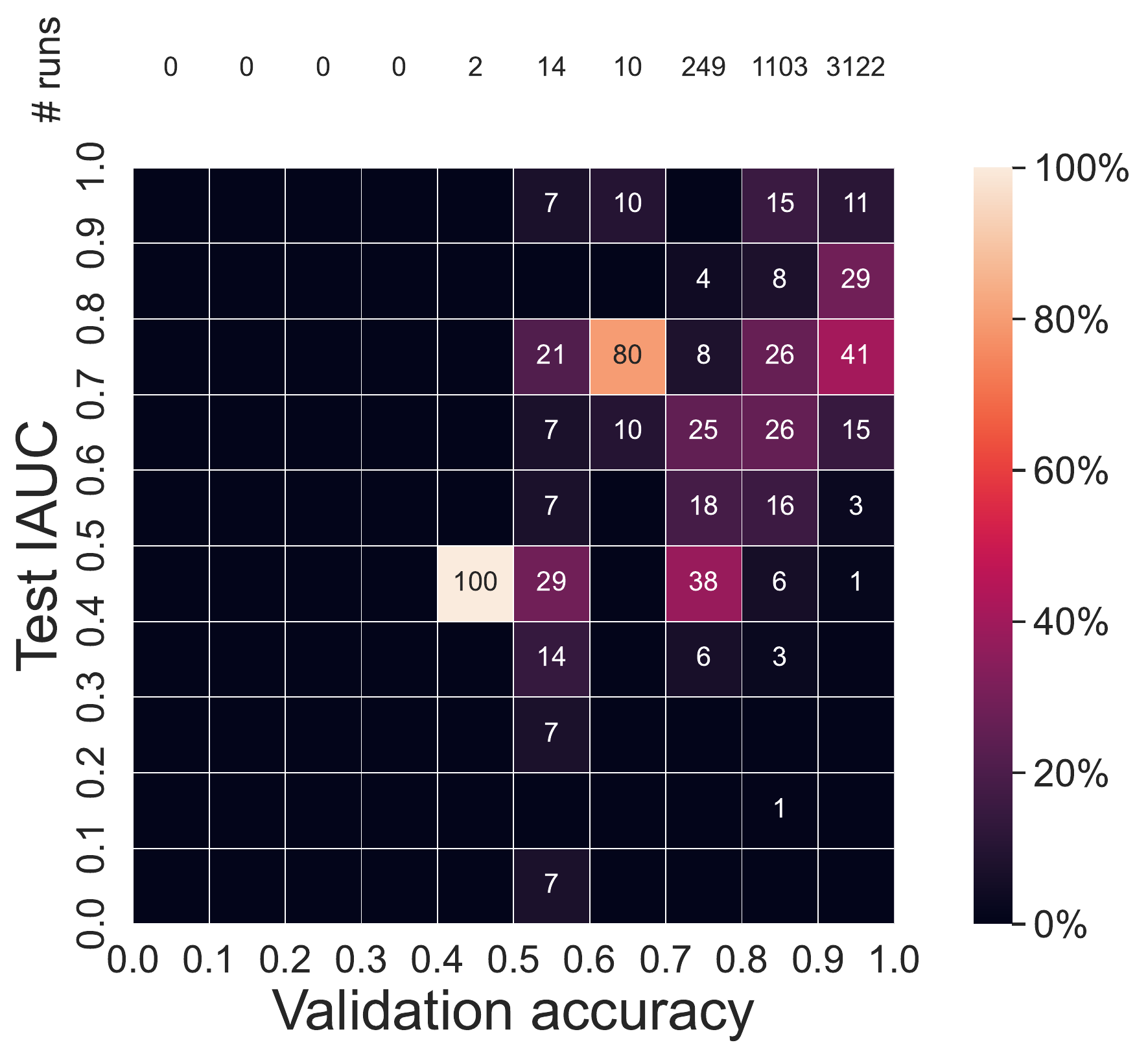}
  \caption{Relationship between validation accuracy and test IAUC for top configurations.
    Models are binned by validation accuracy and IAUC and each bin displays the fraction of total models \textit{per column} (\textit{i.e.} per accuracy bin). The total number of models in each column is reported at the top.
    \label{fig:acc_vs_auroc_top_config}}
\end{figure}

For each problem and data modality, we use the top 5 configurations defined in \cref{sec:models-with-high} to evaluate how well validation-time classification performance discriminates between models with low and high-quality explanations.
As we show in~\cref{fig:acc_vs_auroc_top_config}, high performance is not a good indicator of good explanations, and the correlation between accuracy and IAUC exists but is rather mild. A more detailed picture separated by problem and data modality is available in~\cref{sec:corr-iauc-acc}.
In the case of MIL problems on Gaussian data~(\cref{fig:acc_vs_auroc_mil}), all models with the top configurations reach a validation accuracy  of 100\% while having varying IAUC values.
On more complex problems, not all realisations reach perfect accuracies, and a limited amount of correlation can be observed. Indeed, as shown in~\cref{fig:acc_vs_auroc_mnist_and}, it is often the case that only the models with top validation accuracies reach the top values for the IAUC.
Nevertheless, there is still significant variability among the models with top validation accuracies so that filtering out models with a poorer validation performance is not enough to avoid models with poor explanations.

We measure the Spearman correlation $\rho$ between the validation-time accuracy of the 100 repetitions of each top configurations for all our classification tasks and the IAUC score and report them in~\cref{tab:acc_auroc_corr}. For each problem, we further report the configurations with the highest and lowest spreads of IUAC values ($\Delta \text{IAUC}$) between individual top-performing realisations. Namely, to compute $\Delta\text{IAUC}$, we select the models in the highest decile of validation accuracy for each configuration and measure the spread between their maximum and minimum IAUC values. This provides a way of observing how specific configurations have a large variability of IAUC even when filtering for models with high classification performance.

%

\begin{table}
\centering%
\begin{tabularx}{\linewidth}{lsbbb}
\toprule
\thead{Data} & \thead{Problem} & \thead{{\textcolor{white}.}Spearman $\rho$} & \thead{High $\Delta$IAUC} & \thead{Low $\Delta$IAUC} \\
\midrule

{} & MIL & $\phantom{-}0$ & 0.74 & 0.13 \\
\rowcolor{Gray} \cellcolor{white} Gaussian & AND & $\phantom{-}0.50 \pm 0.18$ & 0.69 & 0.12 \\
{} & XOR & $\phantom{-}0.65 \pm 0.16$ & 0.28 & 0.11 \\
\cmidrule(lr){2-5}
{} & MIL & $-0.03 \pm 0.06$ & 0.41 & 0.13 \\
\rowcolor{Gray} \cellcolor{white} MNIST & AND & $-0.01 \pm 0.12$ & 0.42 & 0.10  \\
{} & XOR & $\phantom{-}0.12 \pm 0.24$ & 0.83 & 0.06 \\
\cmidrule(lr){2-5}
{} & MIL & $\phantom{-}0$ & 0.62 & 0.23 \\
\rowcolor{Gray} \cellcolor{white} CyTOF & AND & $\phantom{-}0.07 \pm 0.12$ & 0.81 & 0.26 \\
{} & XOR & $\phantom{-}0.04 \pm 0.06$ & 0.63 & 0.27 \\
\bottomrule
\end{tabularx}
\caption{Predictivity of classification performance for informative explanations. We report the Spearman correlation between the validation accuracy and the IAUC as well as the highest and lowest $\Delta\text{IAUC}$ found among the models. When all trained models have accuracy 1, we report the Spearman correlation as 0 since accuracy cannot provide any information about IAUC.
\label{tab:acc_auroc_corr}}
\end{table}

\subsection{Ensembling improves explanation robustness}
\label{sec:ensembl-impr-expl}
While the risk of poor explanations is real, most trained models with good performance  achieve satisfying interpretation-based explanation quality.
We therefore propose to use ensembling to reduce the risk of encountering poorly-performing single models.
Two strategies are possible:
\begin{itemize}
  \item Single-configuration ensembling, where a fixed hyperparameter set is chosen based on validation performance and multiple realisations are trained with different random seeds.
  \item Multi-configuration ensembling, where we chose a number of high-performing models and ensemble realisations of each hyperparameter choice.
\end{itemize}
For both approaches, the ensembling is performed with the goal of obtaining \textit{more robust attention-based explanations}. More concretely, for each bag of instances, each model produces an attention distribution over the instances and we compute the average attention scores across models. This yields a valid attention distribution for the ensemble in the sense that the averaged distribution also sums to 1.

As we show in~\cref{tab:ensemble_bad}, ensembling does improve the fraction of models with bad explanations (as defined in~\cref{sec:models-with-high}), and multi-configuration ensembling provides the best option for most cases. The results we report for single-configuration ensembling are the average of the results obtained for the top five configurations found for each problem through hyperparameter search. As we show in more details in~\cref{sec:ensembles-bad-prop}, this average hides the fact that single-configuration ensembling fails badly for some configurations, while multi-configuration ensembling does not present this failure mode.

\begin{table}
\centering%
\begin{tabularx}{1.\linewidth}{lsbbb}
\toprule
  \thead{Data} & \thead{Problem} &  \multicolumn{3}{c}{\thead{\% bad configs.}}\\
  \cmidrule(lr){3-5}
   & &\thead{N=1}& \thead{N=20 (single)} & \thead{N=20 (mult.)} \\
\midrule
{} & MIL & 5.3 & 5.3 & 0.0 \\
\rowcolor{Gray} \cellcolor{white} Gaussian & AND & 69.3 & 32.0 & 15.3\\
{} & XOR & 42.3 & 0.0 & 0.0\\
\cmidrule(lr){2-5}
{} & MIL & 1.3 & 0.0 & 0.0 \\
\rowcolor{Gray} \cellcolor{white} MNIST & AND & 12.3 & 5.3 & 6.0\\
{} & XOR & 11.3 & 0.0 & 0.0\\
\cmidrule(lr){2-5}
{} & MIL & 6.0 & 0.0 & 0.0\\
\rowcolor{Gray} \cellcolor{white} CyTOF & AND & 10.0 & 0.0 & 0.0 \\
{} & XOR & 10.7  & 0.0 & 0.0\\
\bottomrule
\end{tabularx}
\caption{Impact of ensembling on the fraction of models with bad explanations. We compare three situations: no ensembling (N=1), and ensembling 20 models with either single configuration ensembling (N=20, single) or multi-configuration ensembling (N=20, mult.).\label{tab:ensemble_bad}}
\end{table}






\section{Discussion}
\label{sec:discussion}
Our experiments confirm that, most of the time, attention mechanisms provide meaningful information about the relative importance of instances in set classification problems like MIL.
Nevertheless, silent failure modes exist where individual models can have good performance at the main weakly supervised task but produce attention maps that are not aligned with the amount of signal carried by data sub-components.
This finding is somewhat worrying: with a bit of bad luck, a researcher could train a good model with poor interpretability and generate new hypotheses based on nonsensical explanations, which could lead to resource waste if they are the basis for experimental studies.
However, attention-based explanations should not altogether be discarded, but be considered with care.
As our ensembling experiments show, sporadically appearing bad-behaving models can be mitigated, but not altogether avoided in a multi-model setup as silent failures seem to fall in the minority.
In some settings, however, ensembling by averaging attention scores does not improve the failure rate.
We suspect that this is due to poor agreement between the attention assignment of different models, leading to poor ensemble performance, which could be improved by switching to majority voting.
If this is the case, we could avoid false positive labelling of important instances by requiring a clear consensus between different models, which we hope to explore in future work.
In any case, some responsible downstream analysis and validation of patterns highlighted by attention mechanisms is warranted when trying to discover new features in data, keeping in mind that there is a small but non-zero probability that the patterns might be misleading.

\section{Conclusion}
\label{sec:conclusion}
We showed across a variety of set-classification tasks and data modalities that silent failure modes exist for attention-based key instance attributions, where attention does not correlate with instance importance. While ensembling multiple random initialisations of the same model and multiple model architecture mitigates the issue, there often remains a probability that explanations based on attention could be misleading, which can range from problematic for scientific discovery to dangerous when using explanations to verify predictions in application settings.
This should not be a reason to abandon attention as a tool for identifying important sub-components of data for a given model, but shows that downstream verification of potential patterns is necessary.
We have hinted at the fact that a more fine-grained approach to ensembling could help filter false positives and this is definitely an interesting avenue for further research. Other important directions which we plan to pursue is the identification of the features of tasks where silent failure is less common, as well as understanding which aspects of model architecture impact the quality of importance attribution.


\begin{acks}
  We thank the Systems Biology group at IBM Research Europe  for useful discussions, as well as Mattia Rigotti and Janis Born. This project was support by SNF grant No. 192128 and the H2020 grant "iPC" (No. 826121).
\end{acks}

\bibliographystyle{ACM-Reference-Format}
\bibliography{biblio, custom_biblio}

\newpage
\appendix

\section{Hyperparameter Searches}
\label{sec:hyperp-search}
\subsection{Gaussian Data}
\label{sec:gaussian-data}

\begin{table}[H]
\centering
\begin{tabularx}{\linewidth}{XY}
  \toprule
  \thead{Parameter} & \thead{Values}\\
  \midrule
  \rowcolor{Gray}
  Batch size & 100\\
  Epoch & 100, 200, 500\\
  \rowcolor{Gray}
  Learning rate & 0.001, 0.005, 0.01, 0.02\\
  Weight decay & 0.0001 \\
  \rowcolor{Gray}
  Loss function & Cross entropy \\
  Optim. algorithm & Adam \\
  \rowcolor{Gray}
  Hidden layer size & 2, 4, 8\\
  Attention size & 1, 2, 4, 8 \\
  \rowcolor{Gray}
  Featurizer depth & 0, 1, 2\\
  Classifier depth & 1, 2, 3 \\
  \bottomrule
\end{tabularx}
\caption{Parameter grid for Gaussian data.\label{tab:gaussian_parameter_grid} }
\end{table}

\subsection{Image Data}
\label{sec:image-data}

\begin{table}[H]
\centering
\begin{tabularx}{\linewidth}{XY}
  \toprule
  \thead{Parameter} & \thead{Values} \\
  \midrule
  \rowcolor{Gray}
  Batch size & 100\\
  Epoch & 500\\
  \rowcolor{Gray}
  Learning rate & 0.001, 0.005, 0.01, 0.02\\
  Weight decay & 0.0001 \\
  \rowcolor{Gray}
  Loss function & Cross entropy \\
  Optim. algorithm & Adam \\
  \rowcolor{Gray}
  Hidden layer size & 8, 16, 32, 64\\
  Attention size & 1, 2, 4, 8, 10 \\
  \rowcolor{Gray}
  Featurizer depth & 1, 2\\
  Classifier depth & 1, 2 \\
  \bottomrule
\end{tabularx}
\caption{Parameter grid for image data.\label{tab:mnist_parameter_grid} }
\end{table}

\subsection{CyTOF Data}
\label{sec:cytof-data}

\begin{table}[H]
\centering
\begin{tabularx}{\linewidth}{XY}
  \toprule
  \thead{Parameter} & \thead{Values} \\
  \midrule
  \rowcolor{Gray}
  Batch size & 100\\
  Epoch & 500\\
  \rowcolor{Gray}
  Learning rate & 0.001, 0.005, 0.01, 0.02\\
  Weight decay & 0.0001 \\
  \rowcolor{Gray}
  Loss function & Cross entropy \\
  Optim. algorithm & Adam \\
  \rowcolor{Gray}
  Hidden layer size & 4, 8, 16\\
  Attention size & 1, 2, 4, 8 \\
  \rowcolor{Gray}
  Featurizer depth & 1, 2, 3\\
  Classifier depth & 1, 2, 3 \\
  \bottomrule
\end{tabularx}
\caption{Parameter grid for CyTOF data.\label{tab:cytof_parameter_grid} }
\end{table}
\clearpage

\onecolumn
\section{IAUC Distributions for Top Models}
\label{sec:auroc-distr-top}
Models marked with an asterisk have a significant proportion of bad runs, i.e. 10\% or more of them achieved an IAUC below 0.65.
\begin{figure}[hb]
\centering
\begin{subfigure}{0.18\textwidth}
\centering

\includegraphics[width=\textwidth]{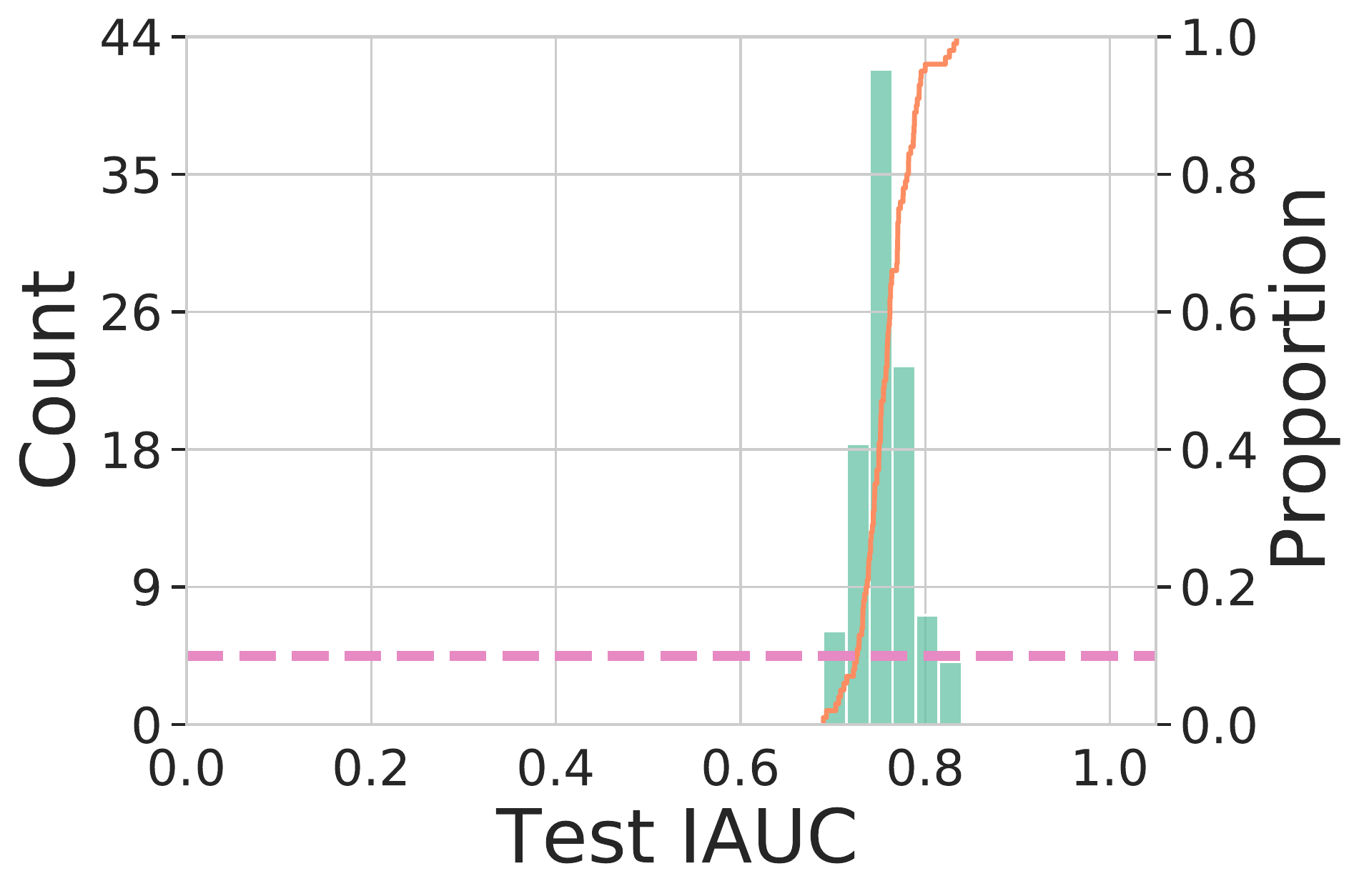}
\caption{Model 1}
\end{subfigure}
\begin{subfigure}{0.18\textwidth}
\centering
 
\includegraphics[width=\textwidth]{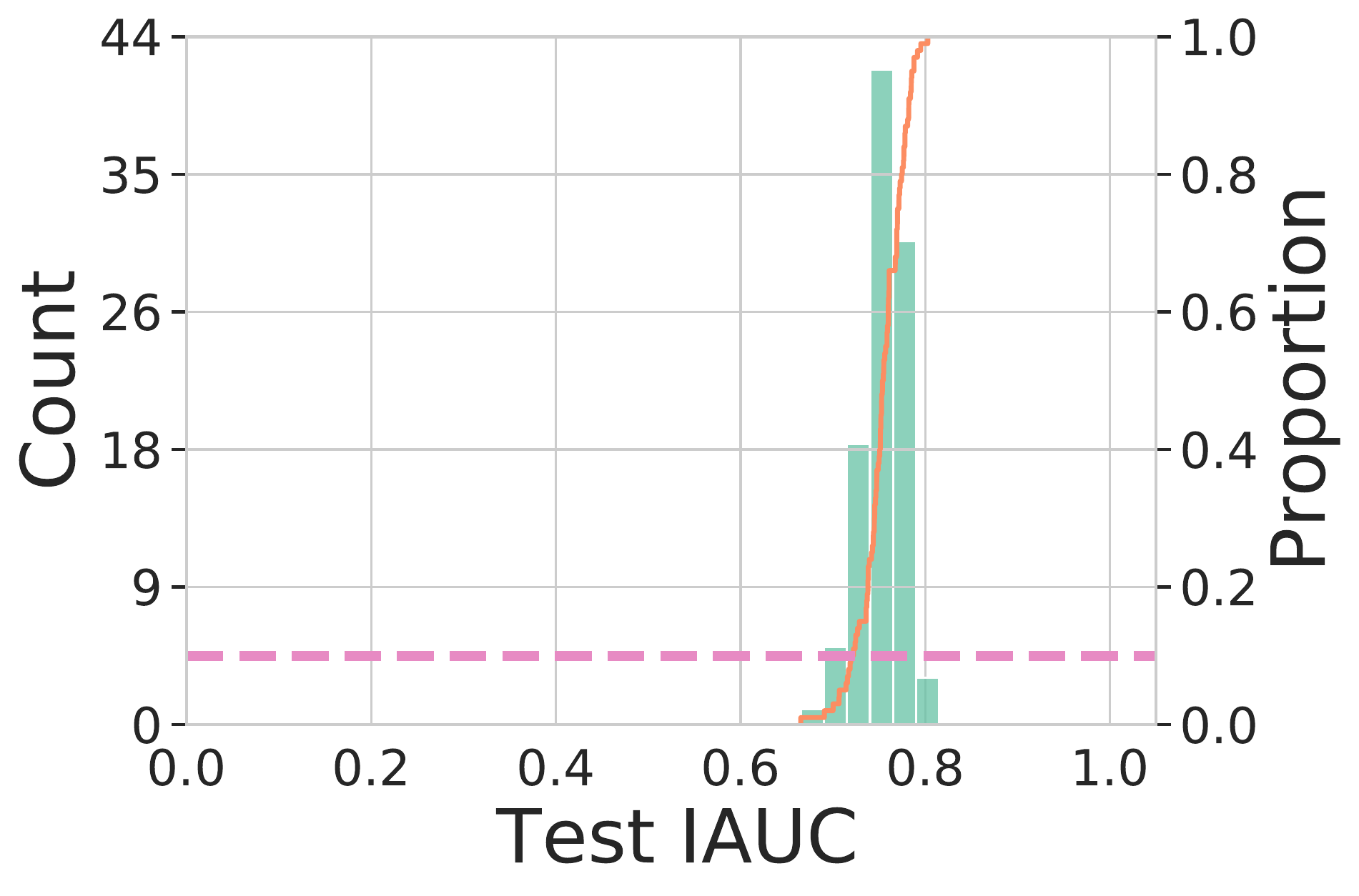}
\caption{Model 2}
\end{subfigure}
\begin{subfigure}{0.18\textwidth}
\centering

\includegraphics[width=\textwidth]{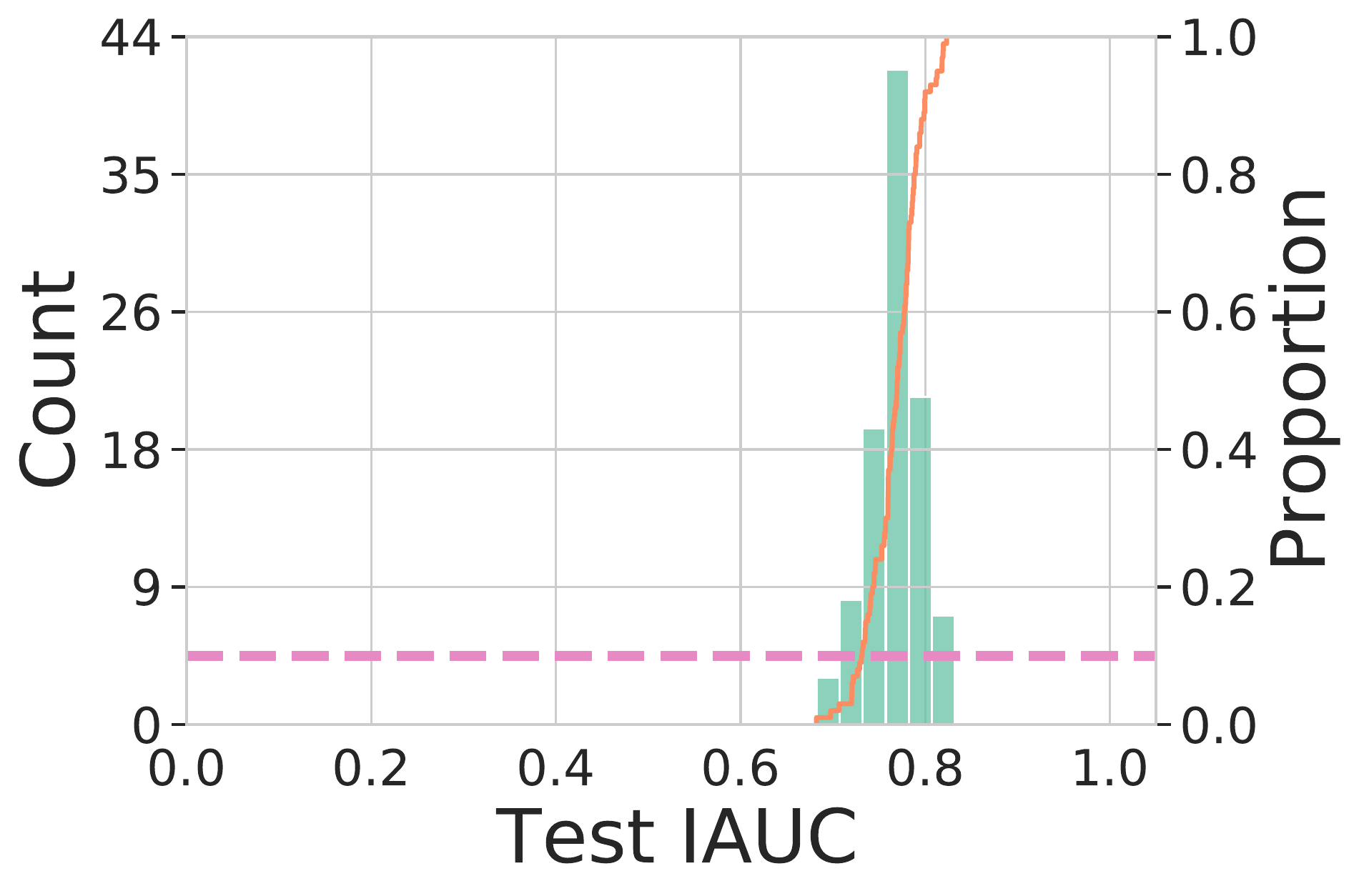}
\caption{Model 3}
\end{subfigure}
\begin{subfigure}{0.18\textwidth}
\centering

\includegraphics[width=\textwidth]{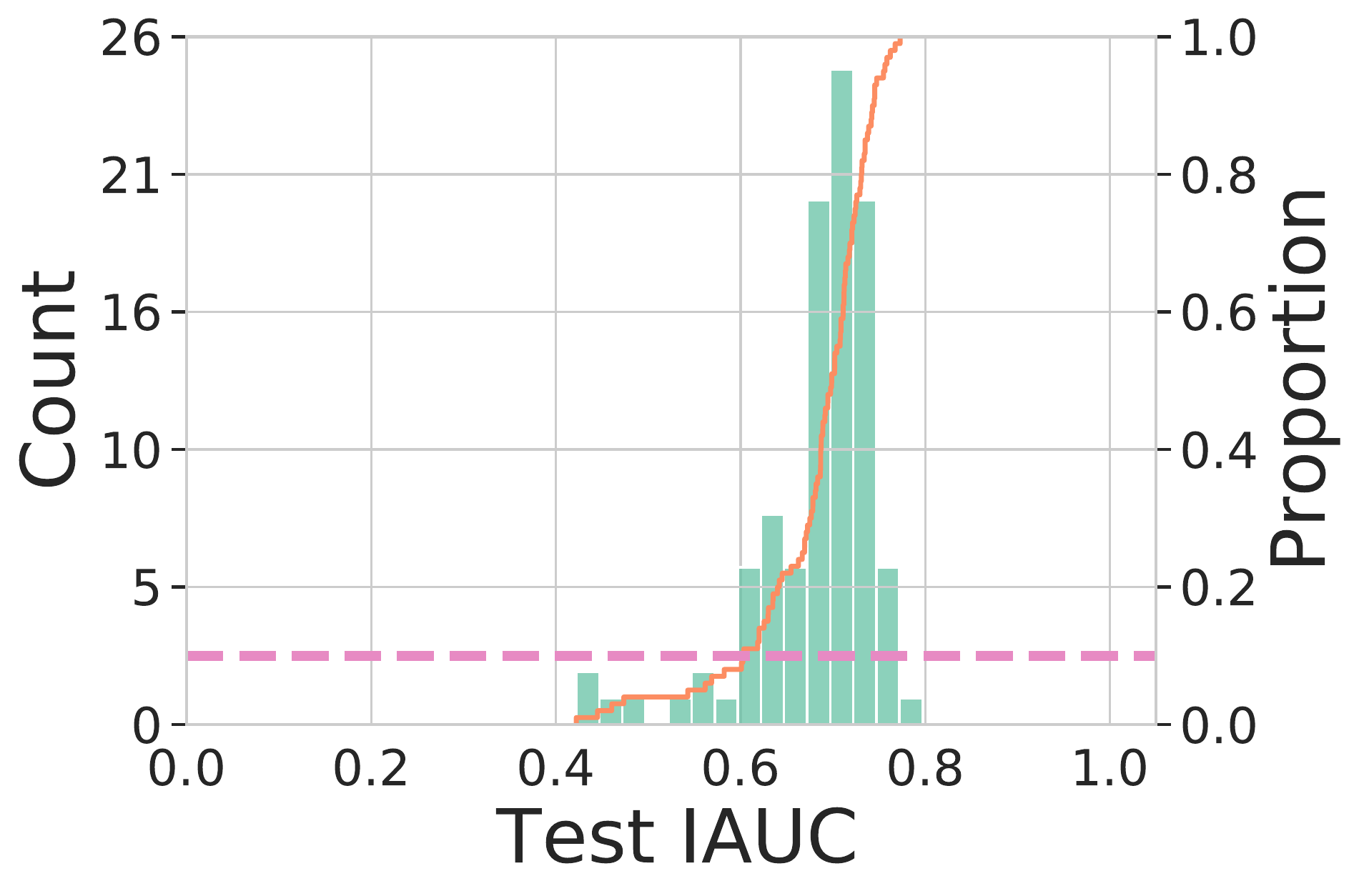}
\caption{Model 4 *}
\end{subfigure}
\begin{subfigure}{0.18\textwidth}
\centering

\includegraphics[width=\textwidth]{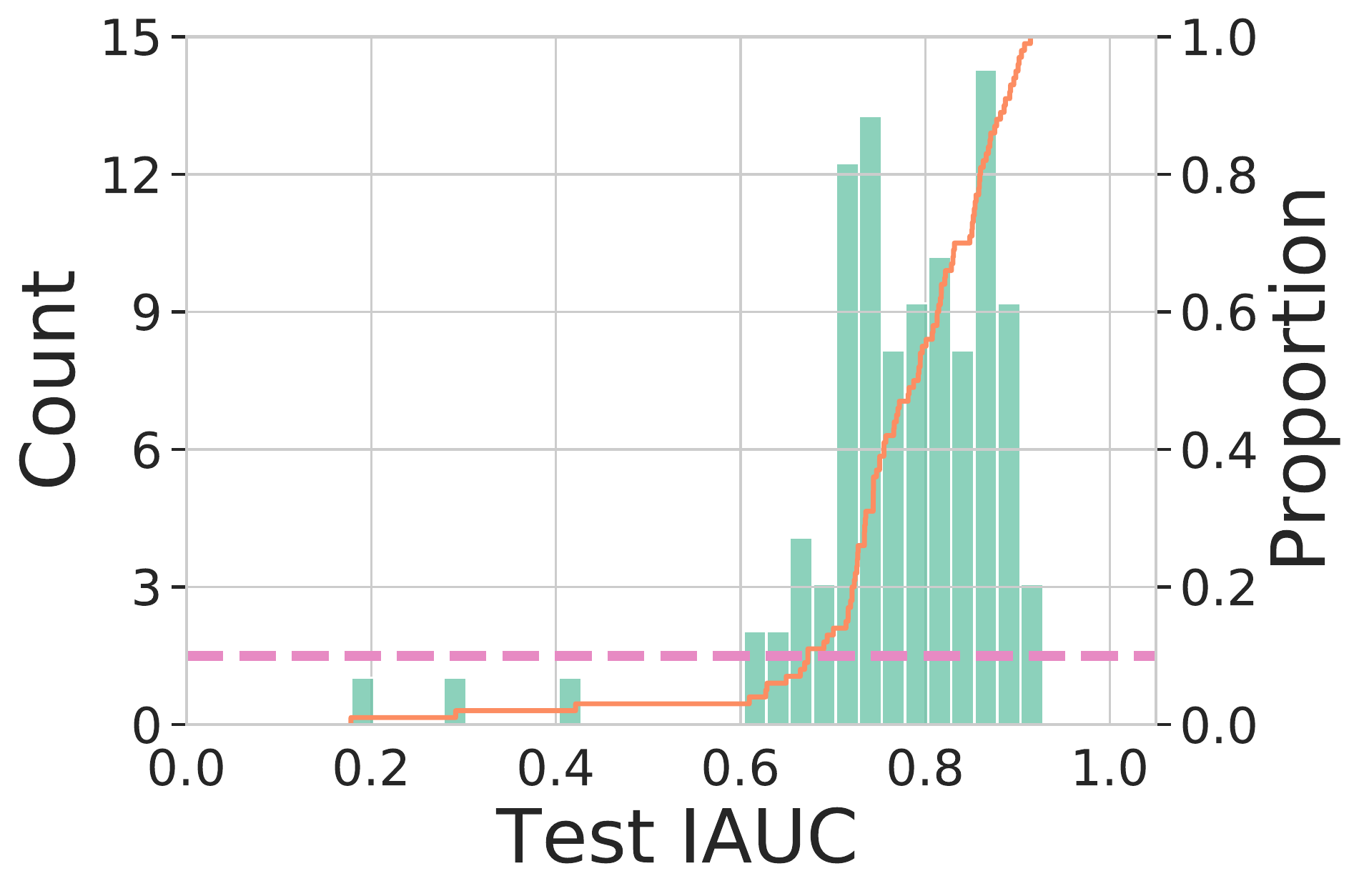}
\caption{Model 5}
\end{subfigure}
\caption[]{Gaussian MIL\label{fig:gaussian_mil_auroc_dist_top_models} }
\end{figure}

\begin{figure}[hb]
\centering
\begin{subfigure}{0.18\textwidth}
\centering

\includegraphics[width=\textwidth]{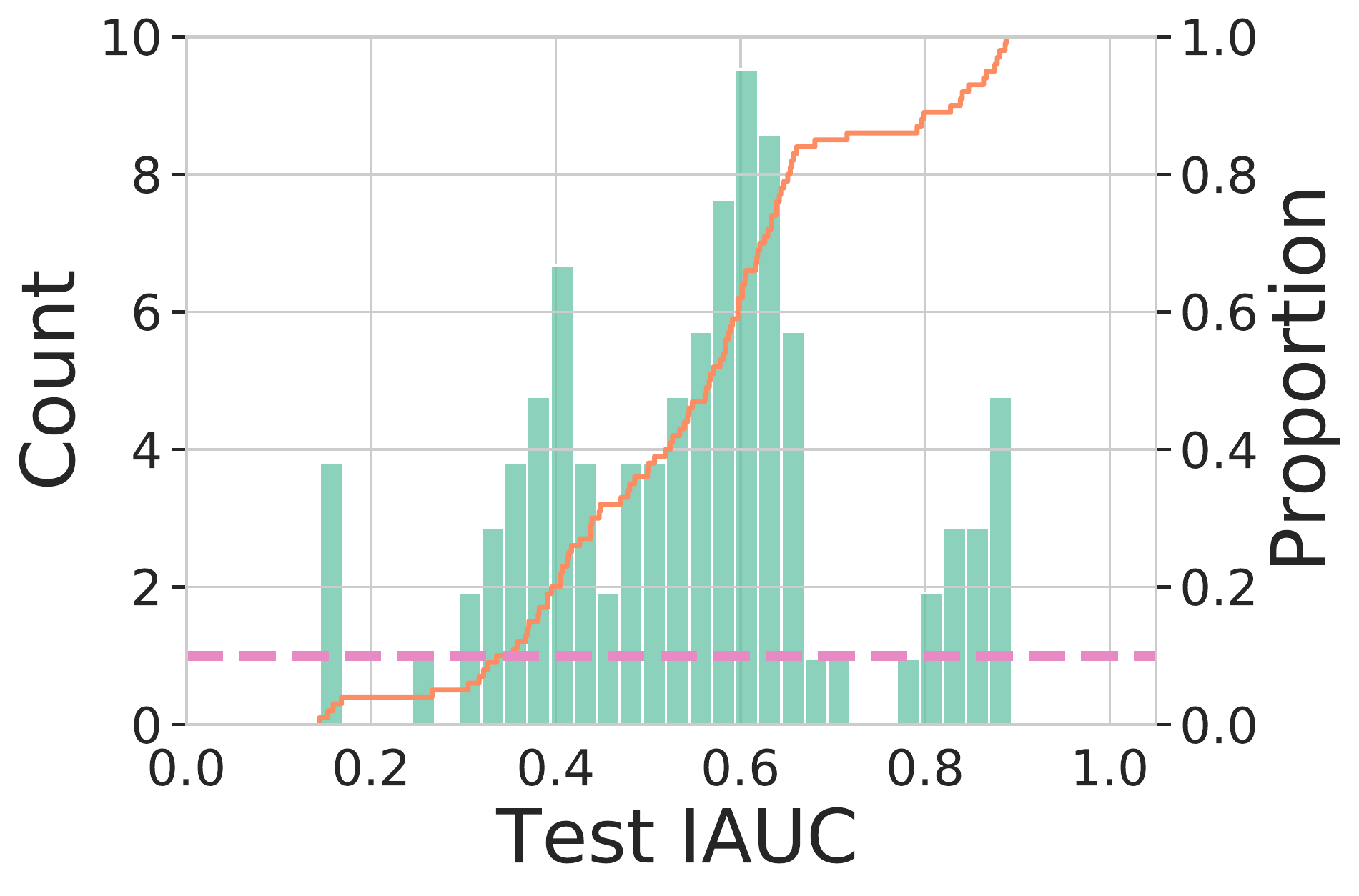}
\caption{Model 1 *}
\end{subfigure}
\begin{subfigure}{0.18\textwidth}
\centering
 
\includegraphics[width=\textwidth]{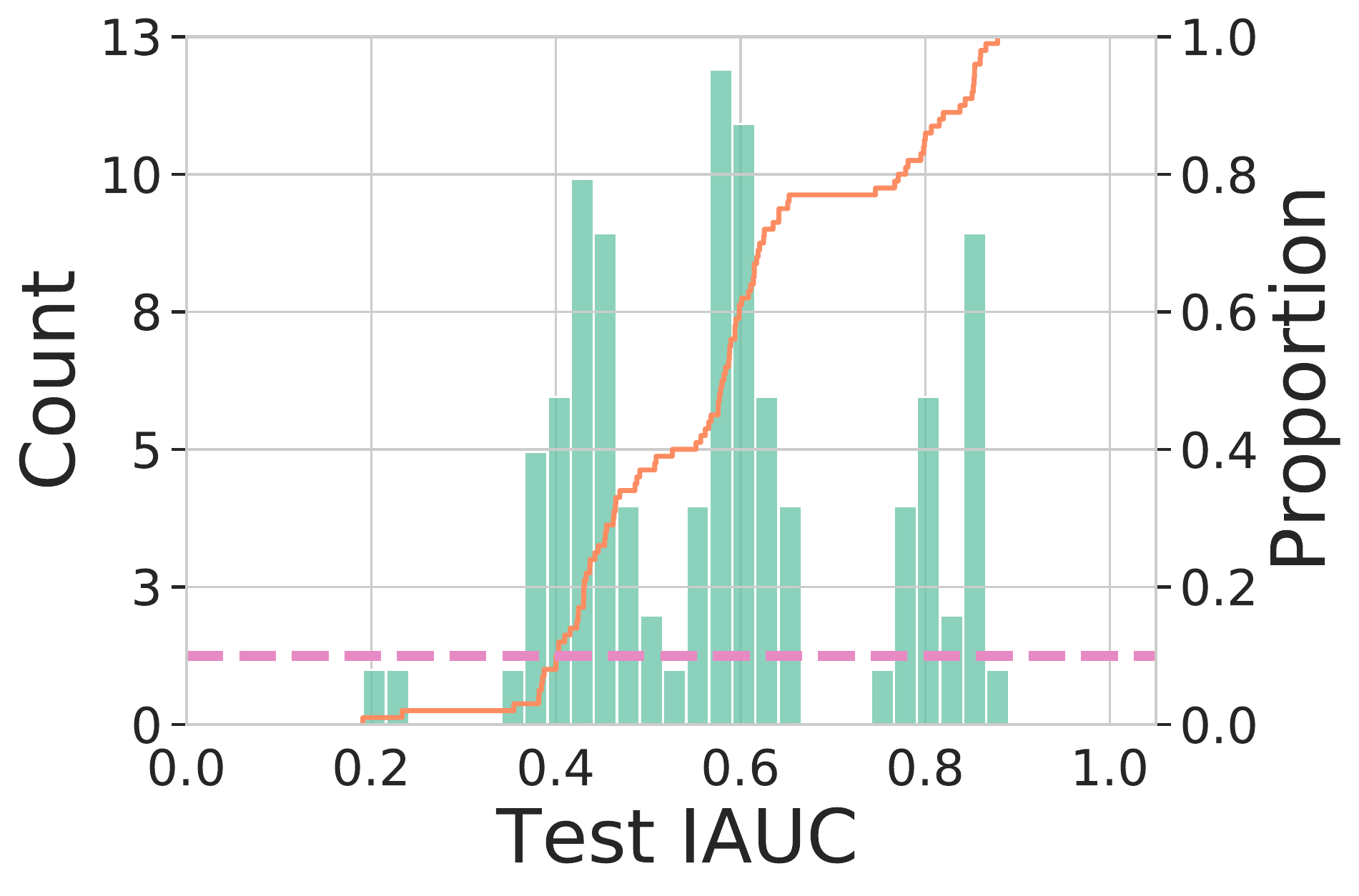}
\caption{Model 2 *}
\end{subfigure}
\begin{subfigure}{0.18\textwidth}
\centering

\includegraphics[width=\textwidth]{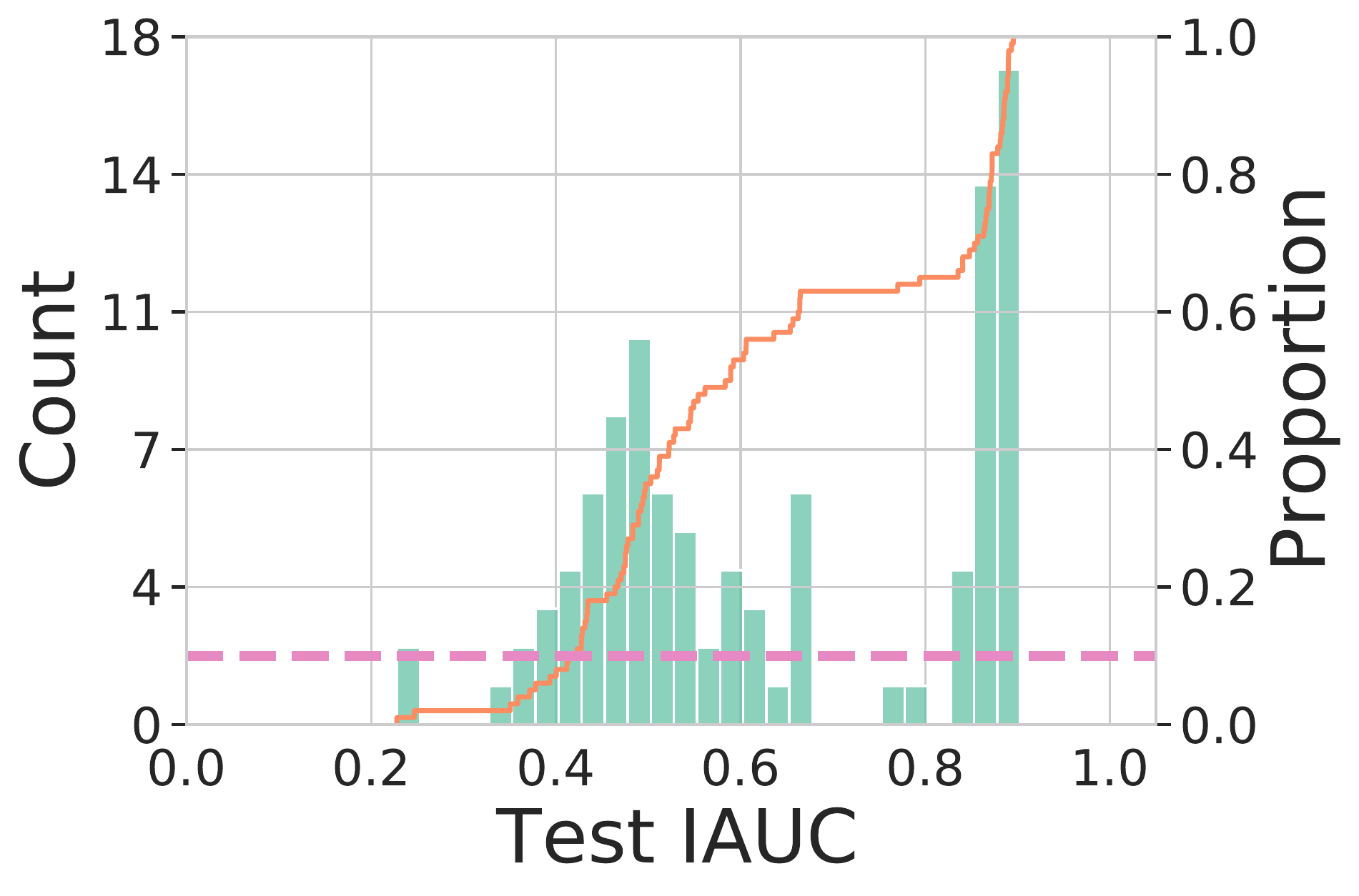}
\caption{Model 3 *}
\end{subfigure}
\begin{subfigure}{0.18\textwidth}
\centering

\includegraphics[width=\textwidth]{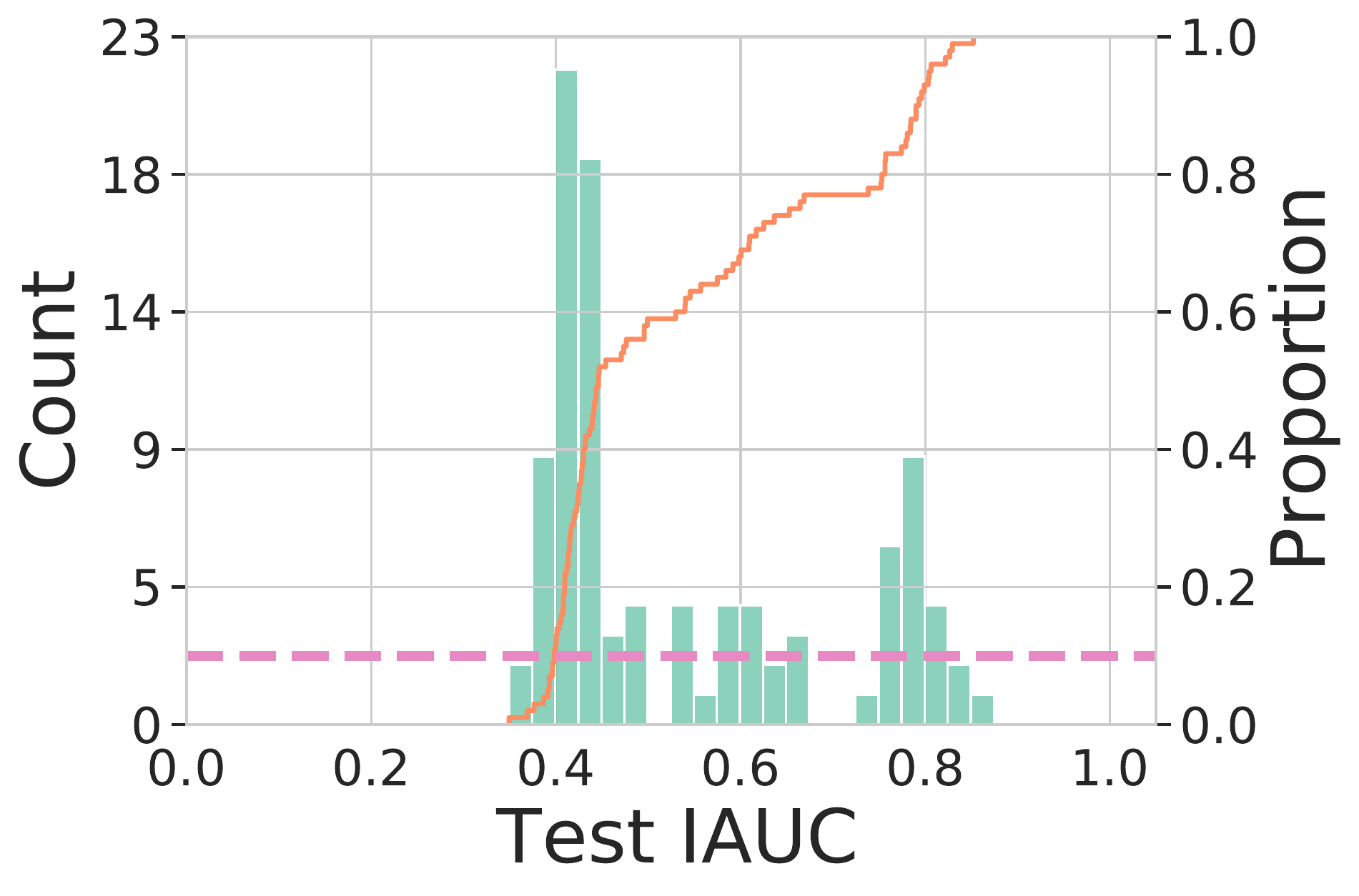}
\caption{Model 4 *}
\end{subfigure}
\begin{subfigure}{0.18\textwidth}
\centering

\includegraphics[width=\textwidth]{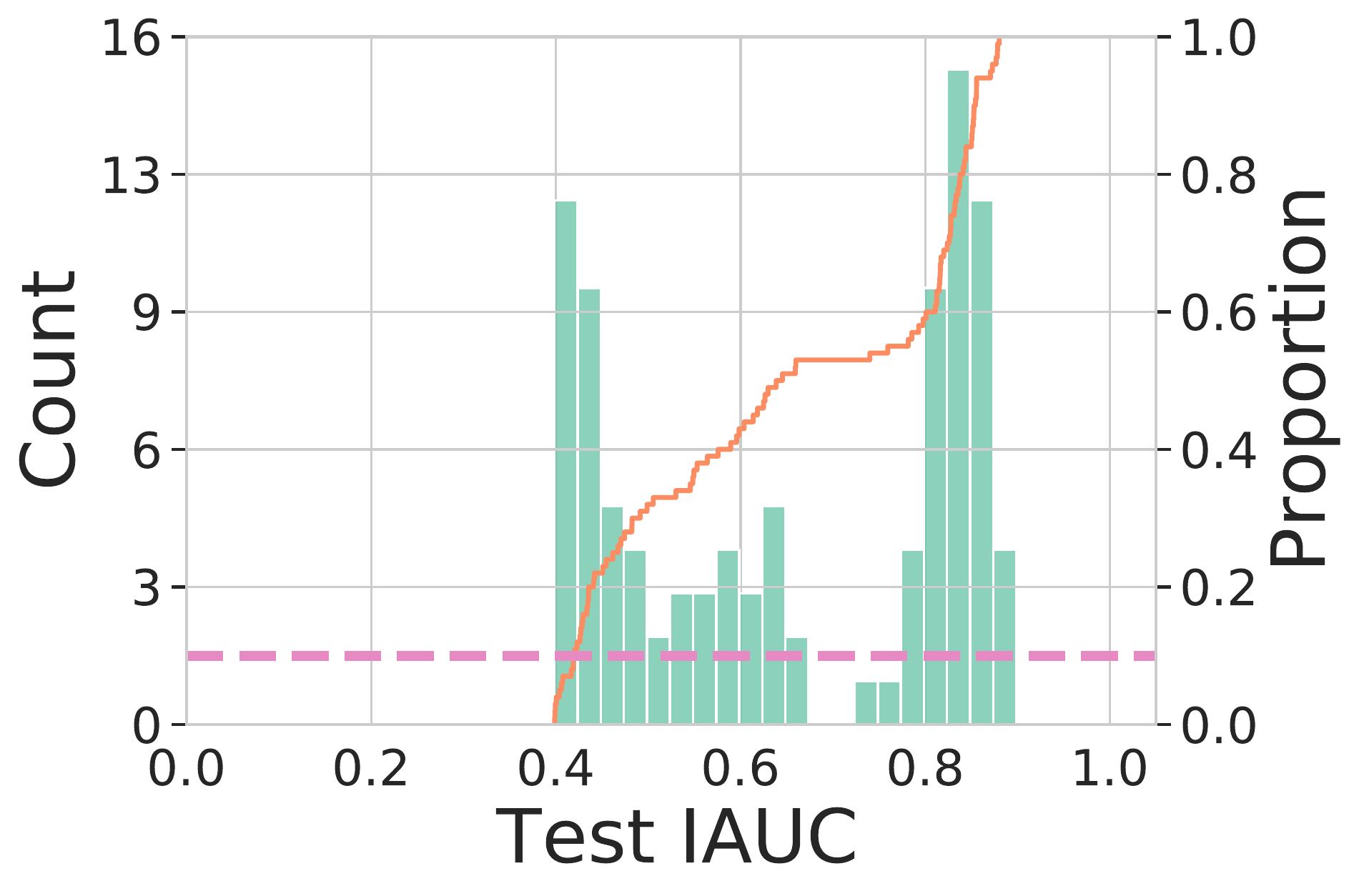}
\caption{Model 5 *}
\end{subfigure}
\caption[]{Gaussian AND\label{fig:gaussian_and_auroc_dist_top_models} }
\end{figure}

\begin{figure}[hb]
\centering
\begin{subfigure}{0.18\textwidth}
\centering

\includegraphics[width=\textwidth]{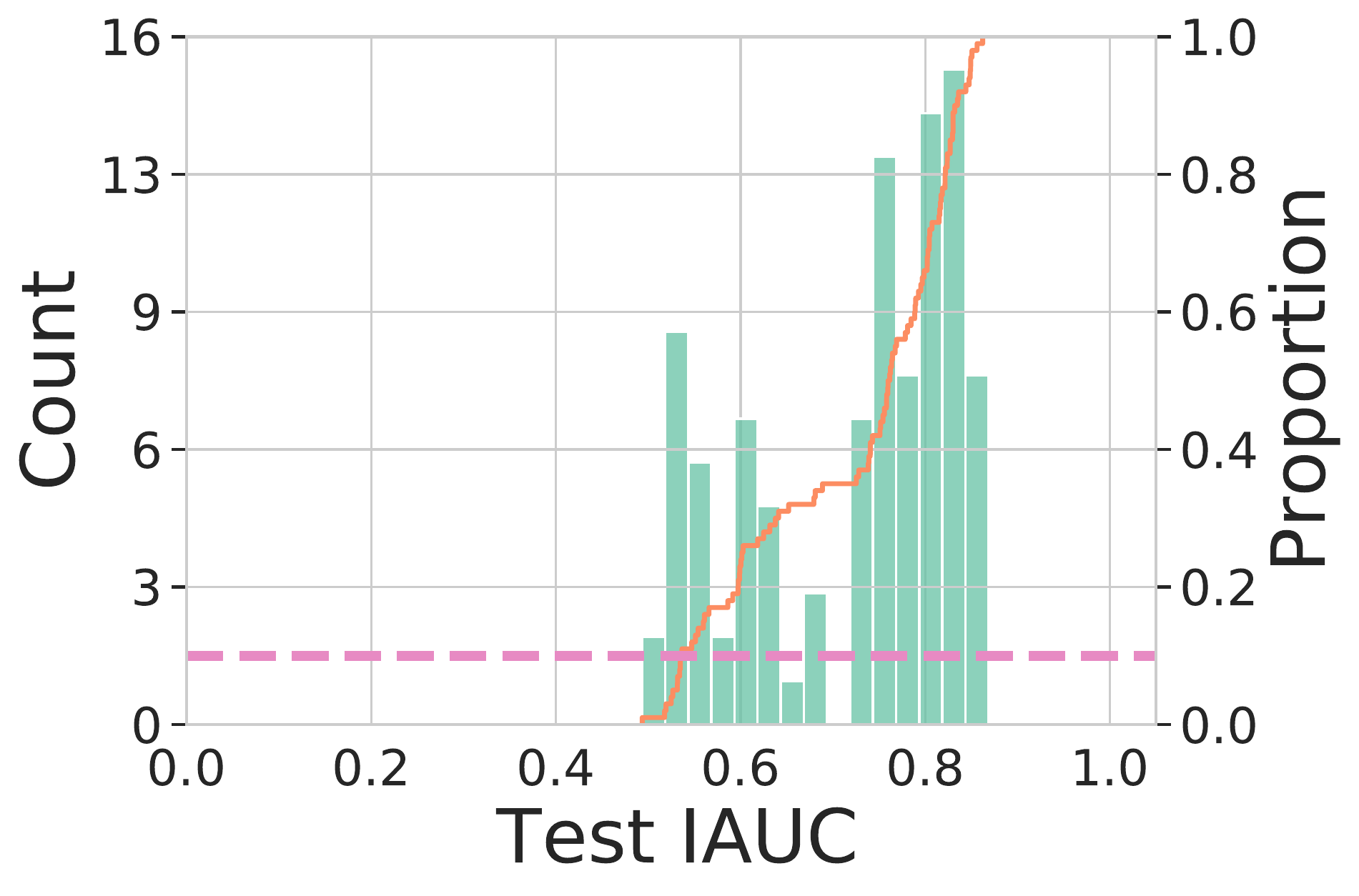}
\caption{Model 1 *}
\end{subfigure}
\begin{subfigure}{0.18\textwidth}
\centering

\includegraphics[width=\textwidth]{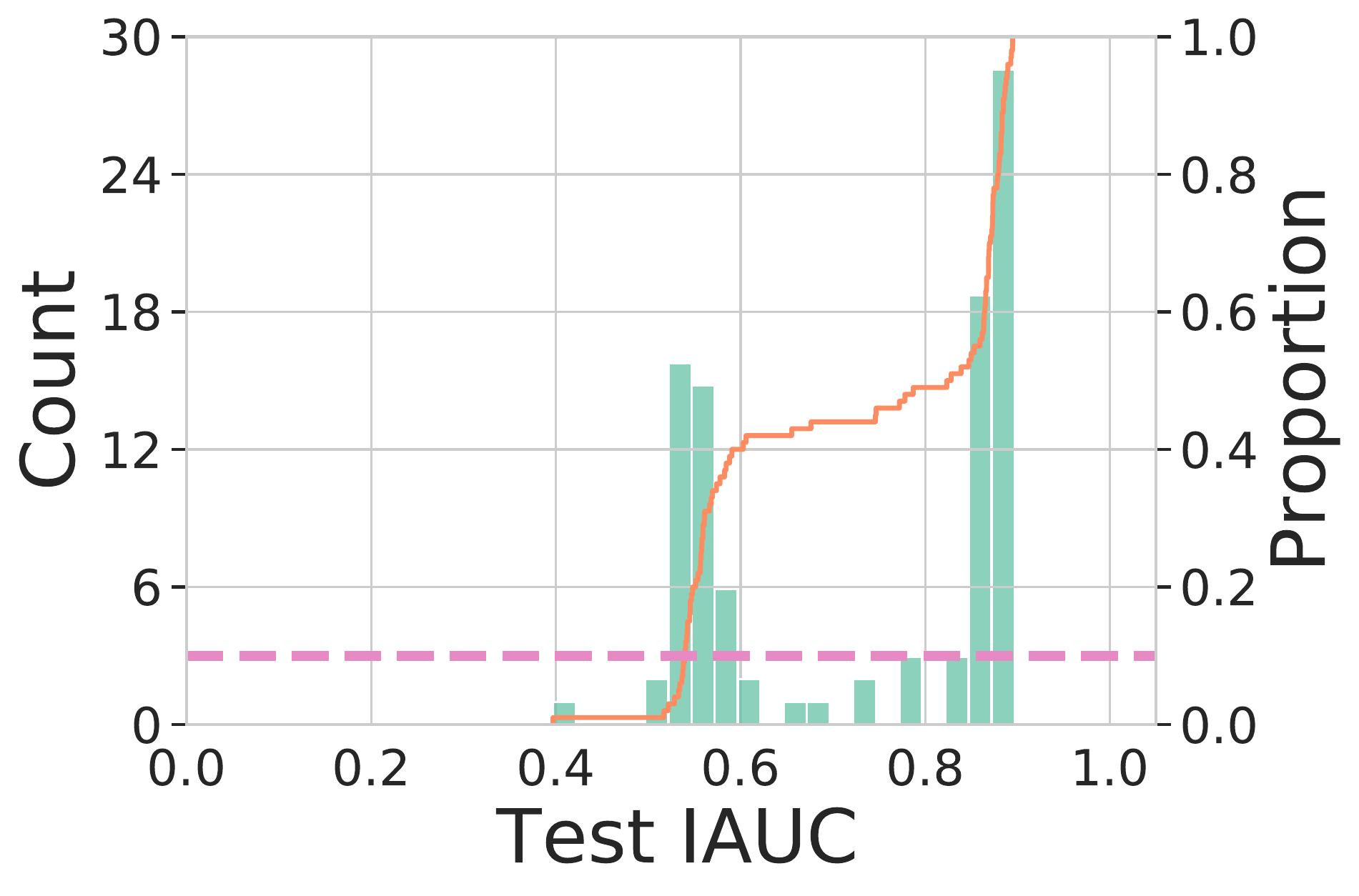}
\caption{Model 2 *}
\end{subfigure}
\begin{subfigure}{0.18\textwidth}
\centering

\includegraphics[width=\textwidth]{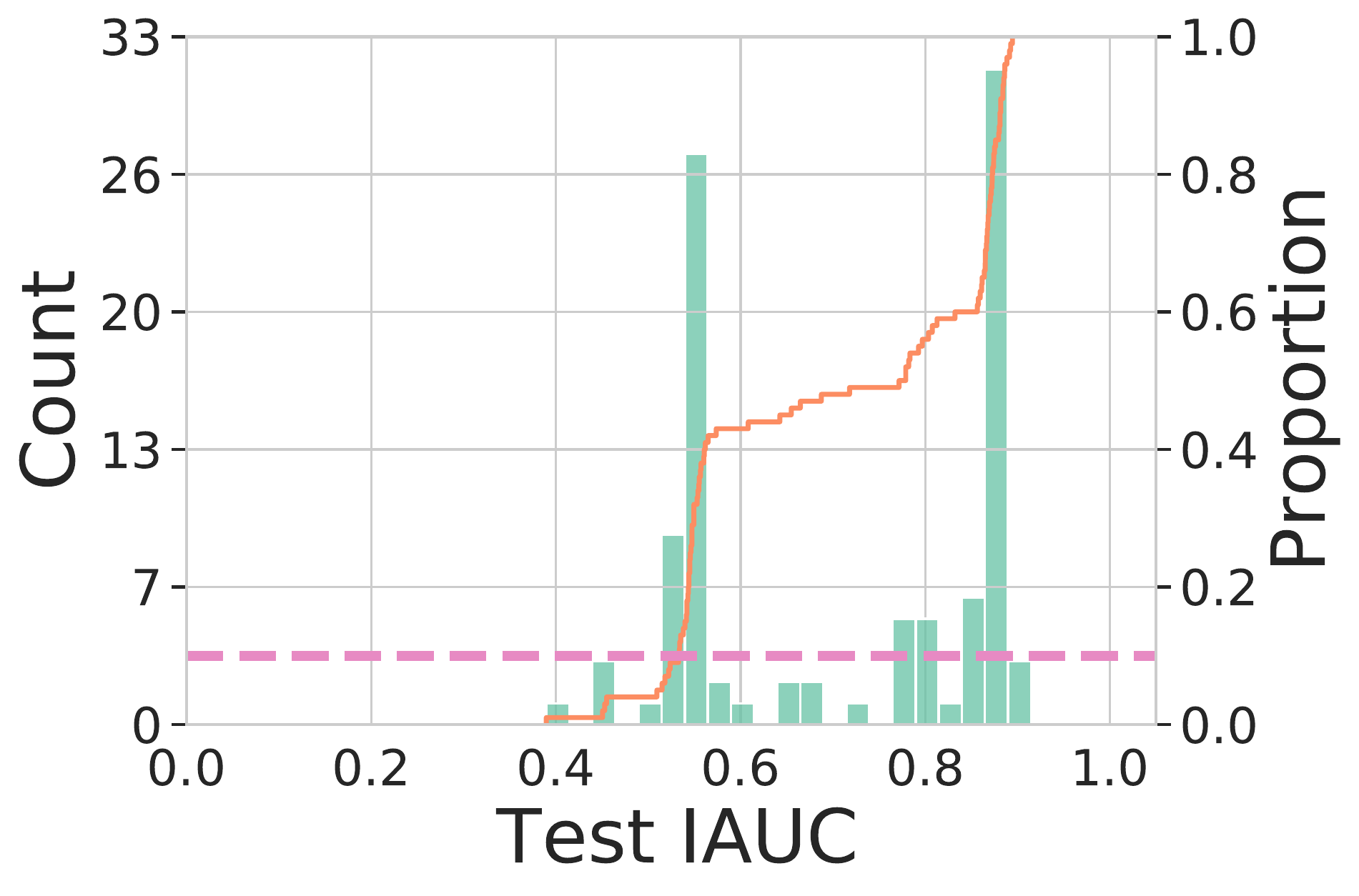}
\caption{Model 3 *}
\end{subfigure}
\begin{subfigure}{0.18\textwidth}
\centering

\includegraphics[width=\textwidth]{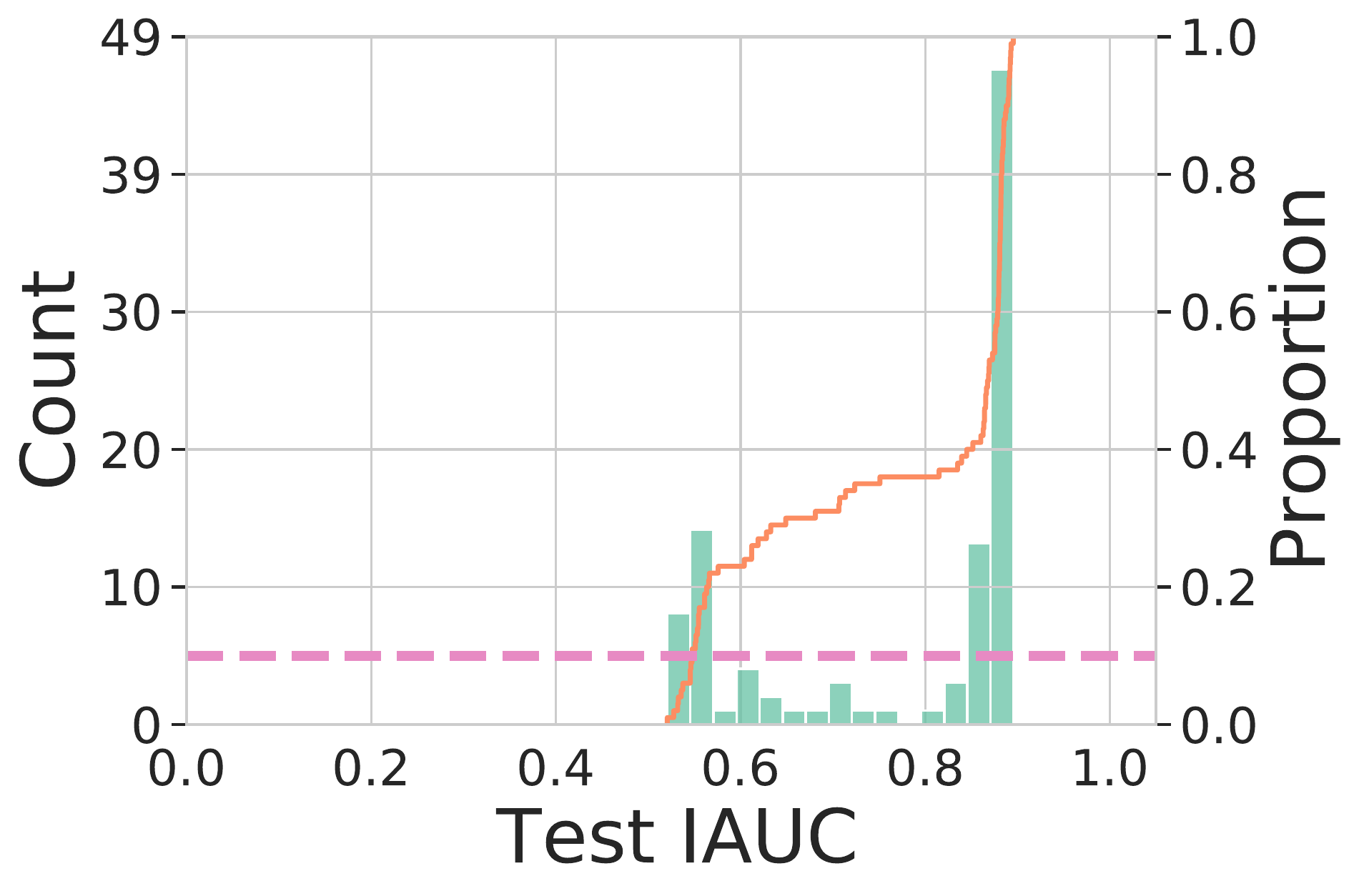}
\caption{Model 4 *}
\end{subfigure}
\begin{subfigure}{0.18\textwidth}
\centering

\includegraphics[width=\textwidth]{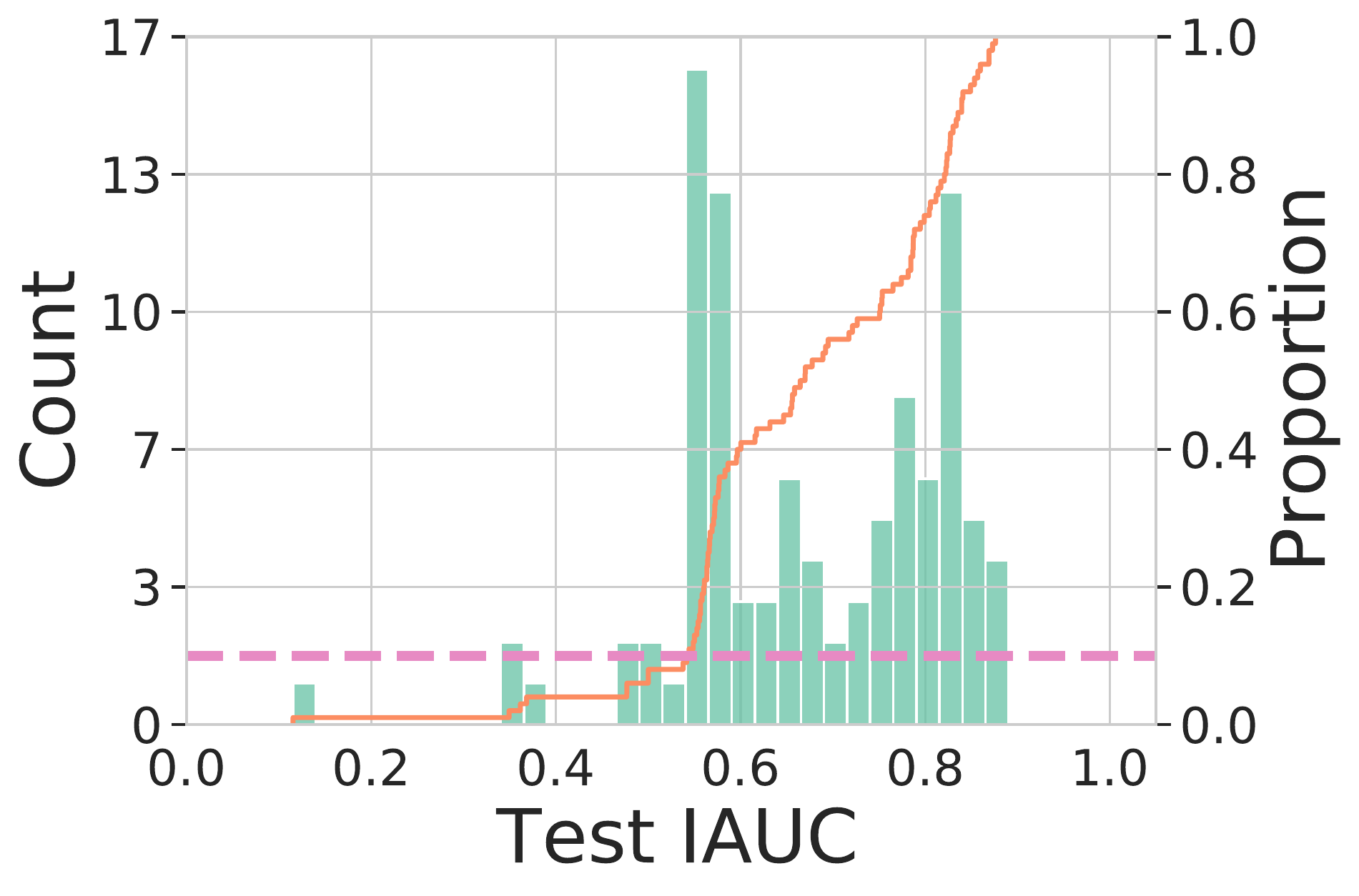}
\caption{Model 5 *}
\end{subfigure}
\caption[]{Gaussian XOR\label{fig:gaussian_xor_auroc_dist_top_models} }
\end{figure}

\begin{figure}[hb]
\centering
\begin{subfigure}{0.18\textwidth}
\centering

\includegraphics[width=\textwidth]{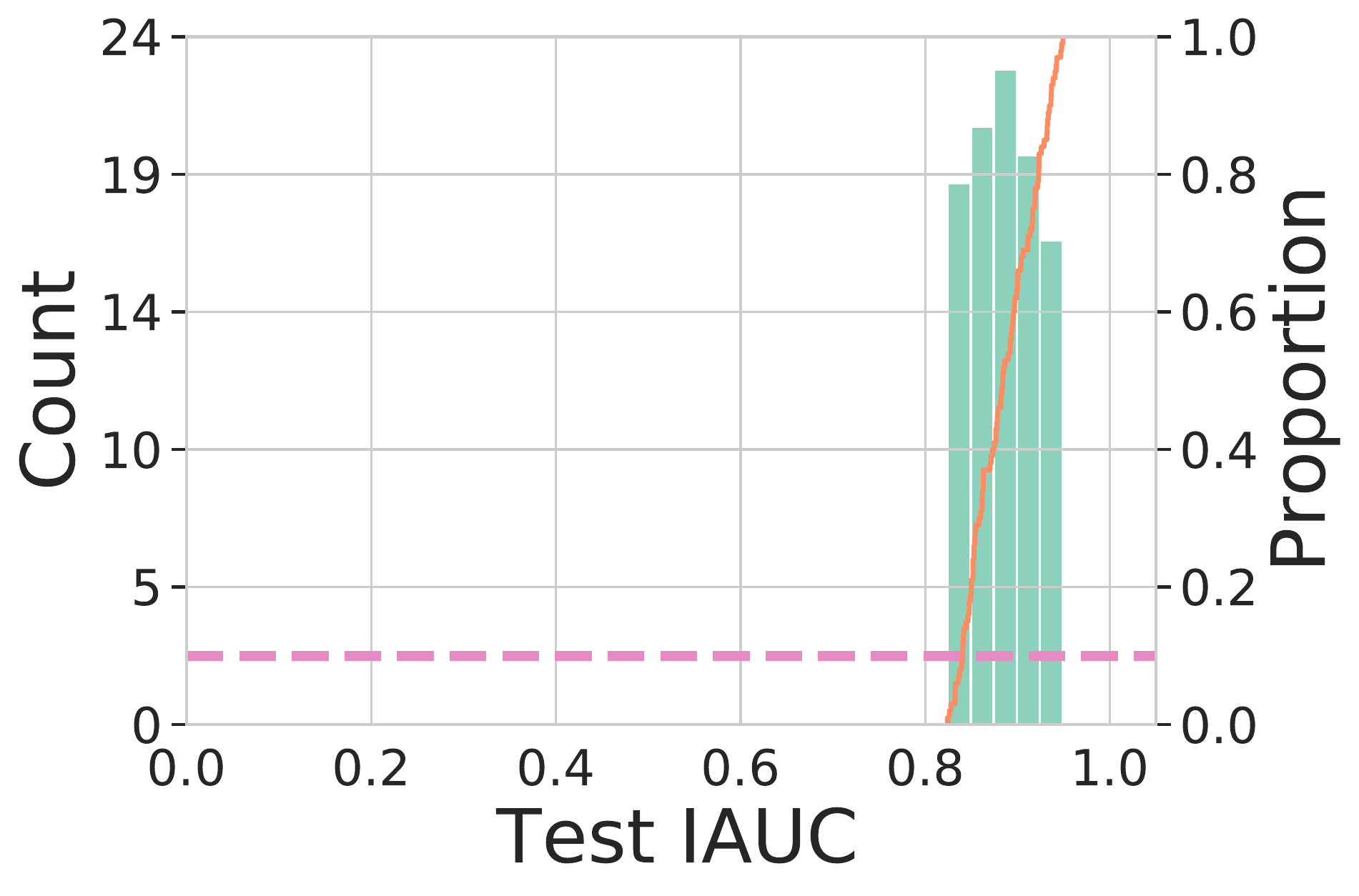}
\caption{Model 1}
\end{subfigure}
\begin{subfigure}{0.18\textwidth}
\centering

\includegraphics[width=\textwidth]{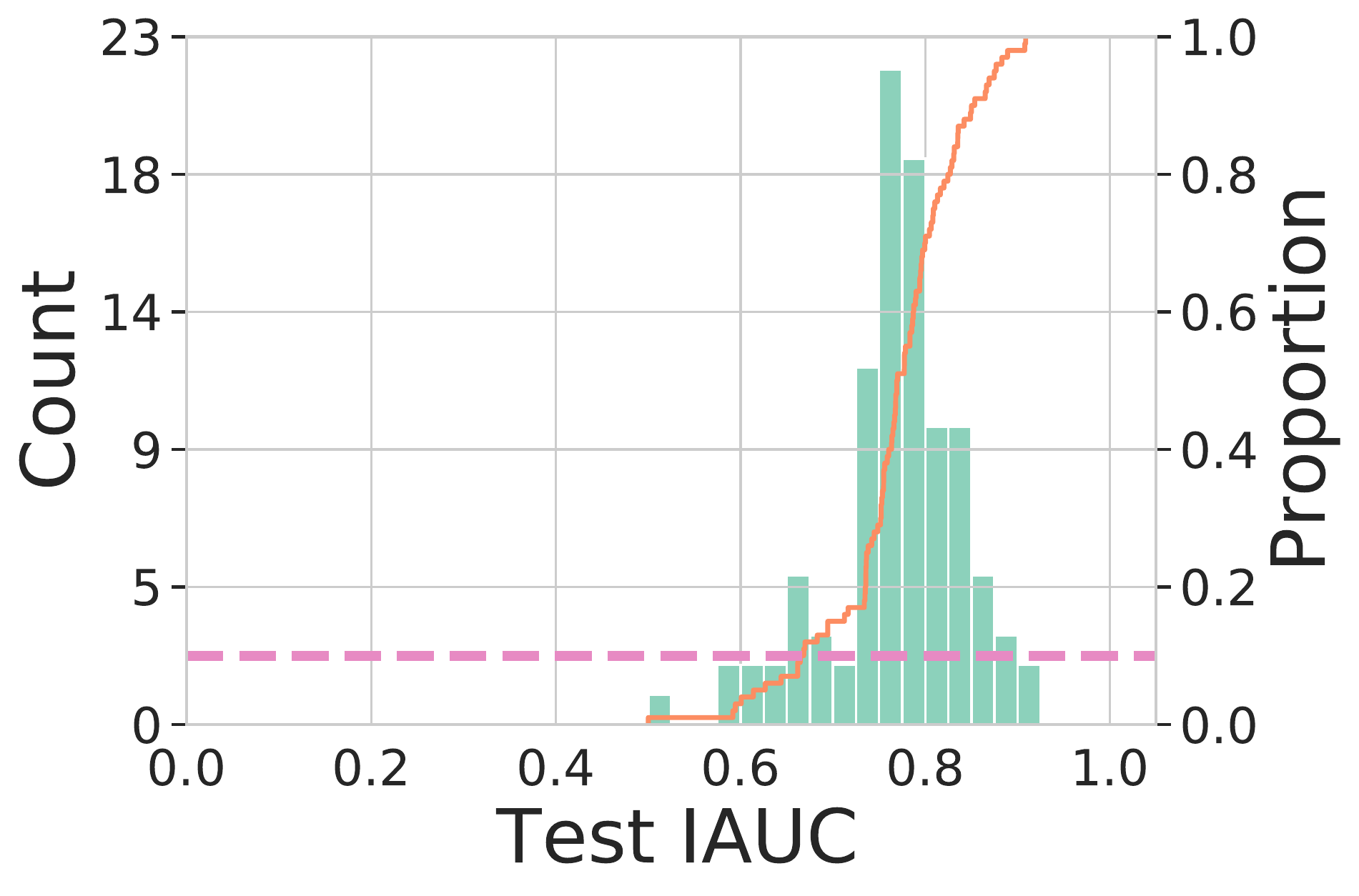}
\caption{Model 2}
\end{subfigure}
\begin{subfigure}{0.18\textwidth}
\centering

\includegraphics[width=\textwidth]{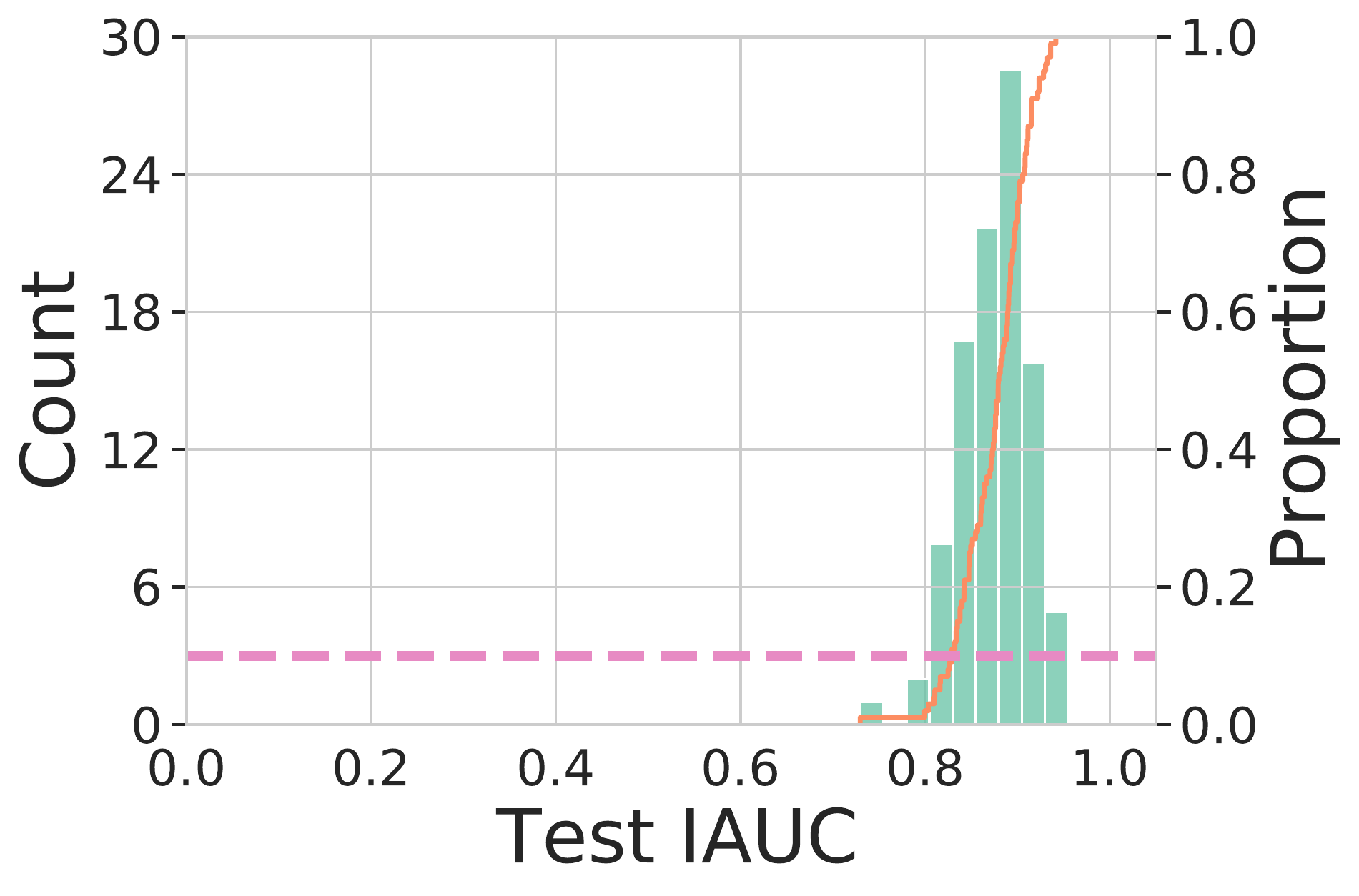}
\caption{Model 3}
\end{subfigure}
\begin{subfigure}{0.18\textwidth}
\centering

\includegraphics[width=\textwidth]{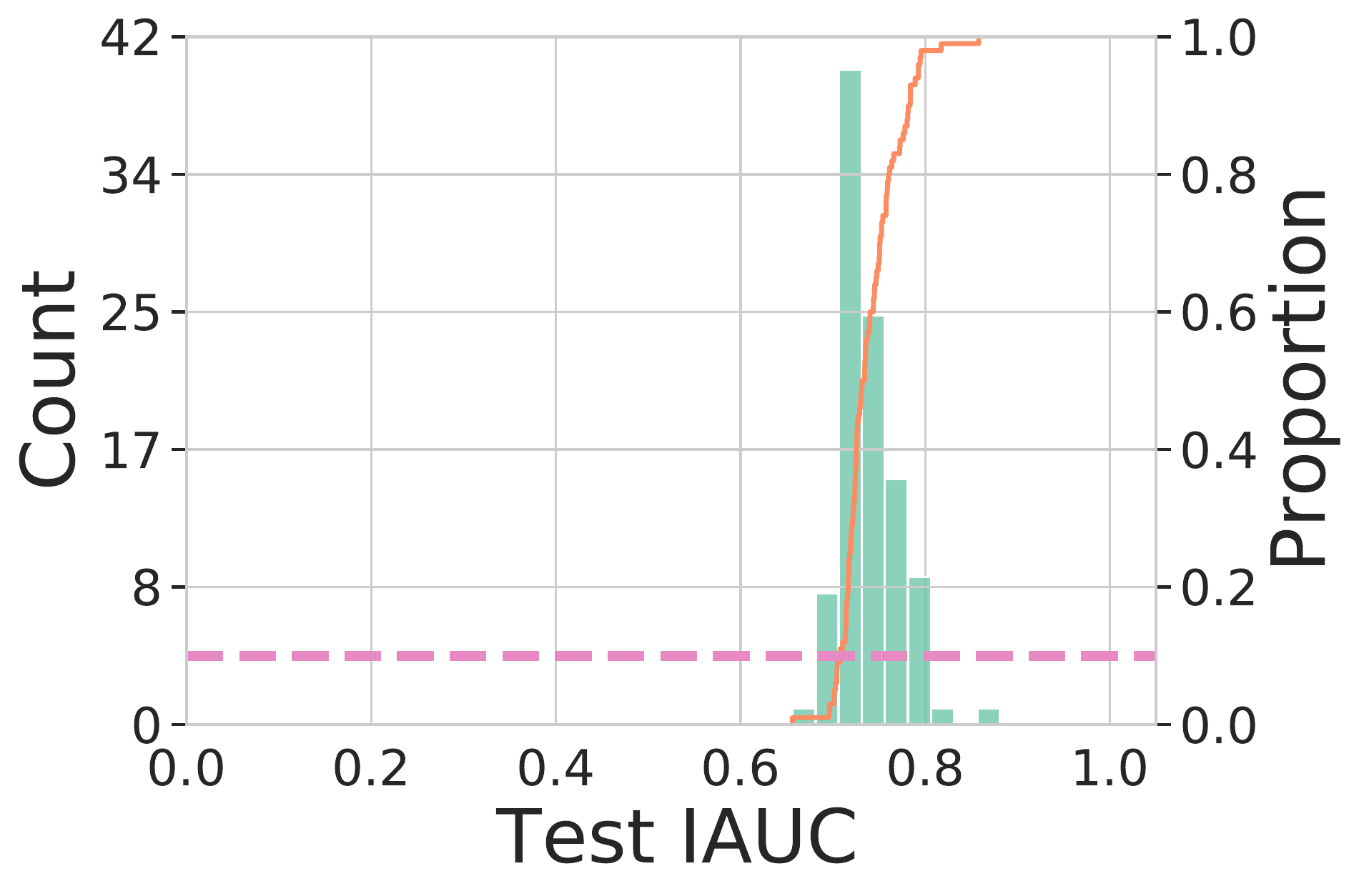}
\caption{Model 4}
\end{subfigure}
\begin{subfigure}{0.18\textwidth}
\centering

\includegraphics[width=\textwidth]{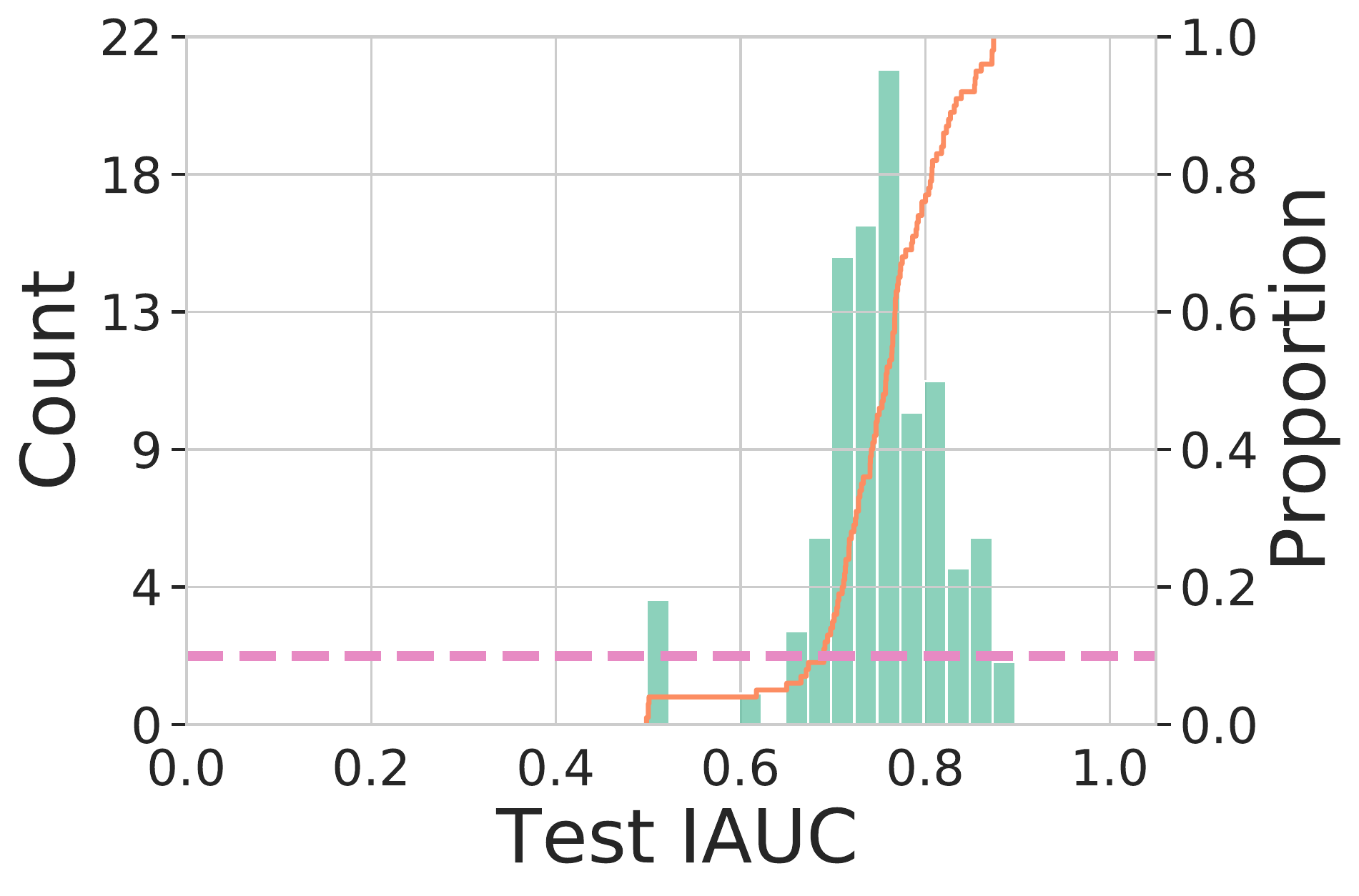}
\caption{Model 5}
\end{subfigure}
\caption[]{MNIST MIL\label{fig:mnist_mil_auroc_dist_top_models} }
\end{figure}

\begin{figure}[hb]
\centering
\begin{subfigure}{0.18\textwidth}
\centering

\includegraphics[width=\textwidth]{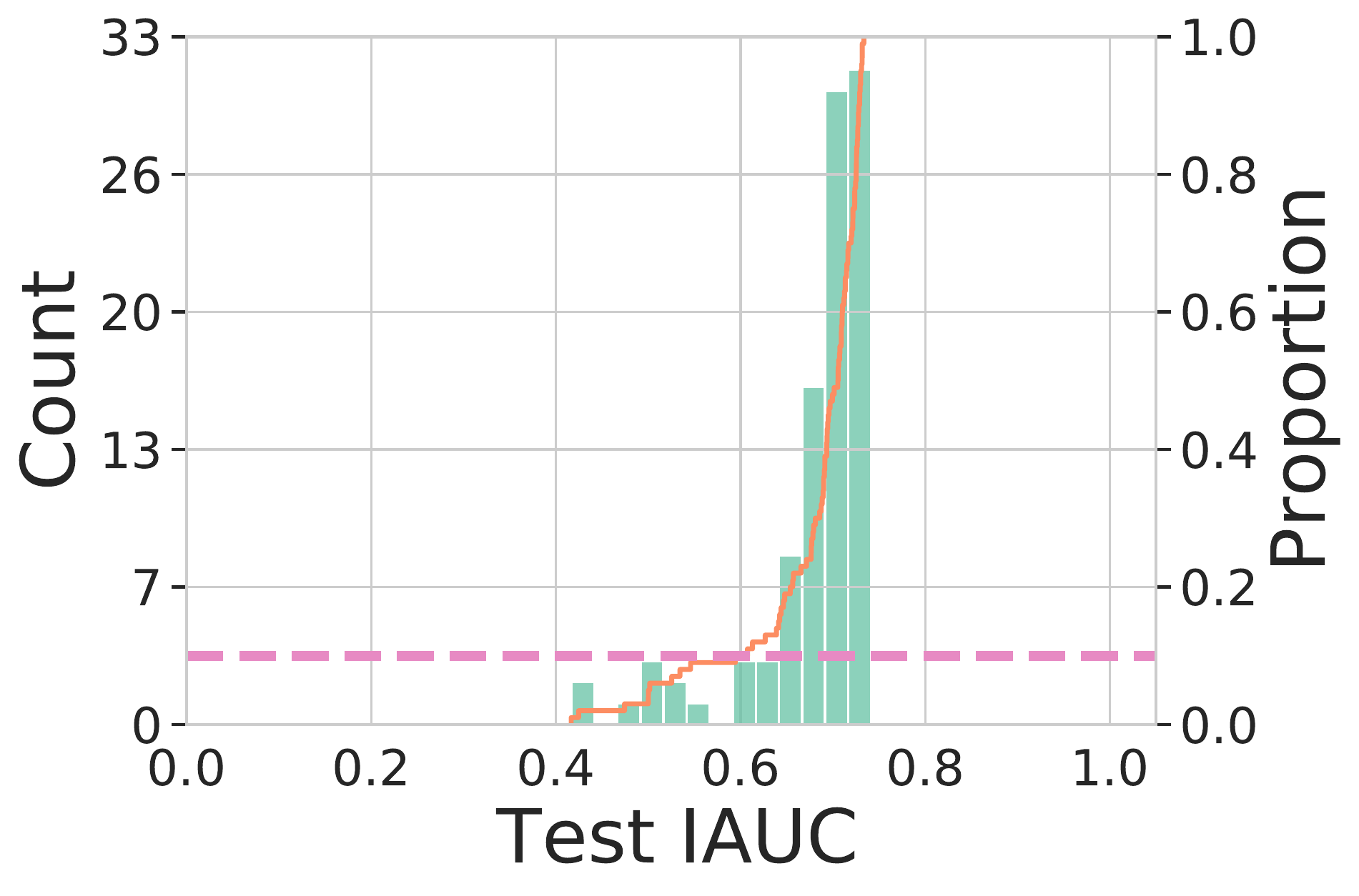}
\caption{Model 1 *}
\end{subfigure}
\begin{subfigure}{0.18\textwidth}
\centering

\includegraphics[width=\textwidth]{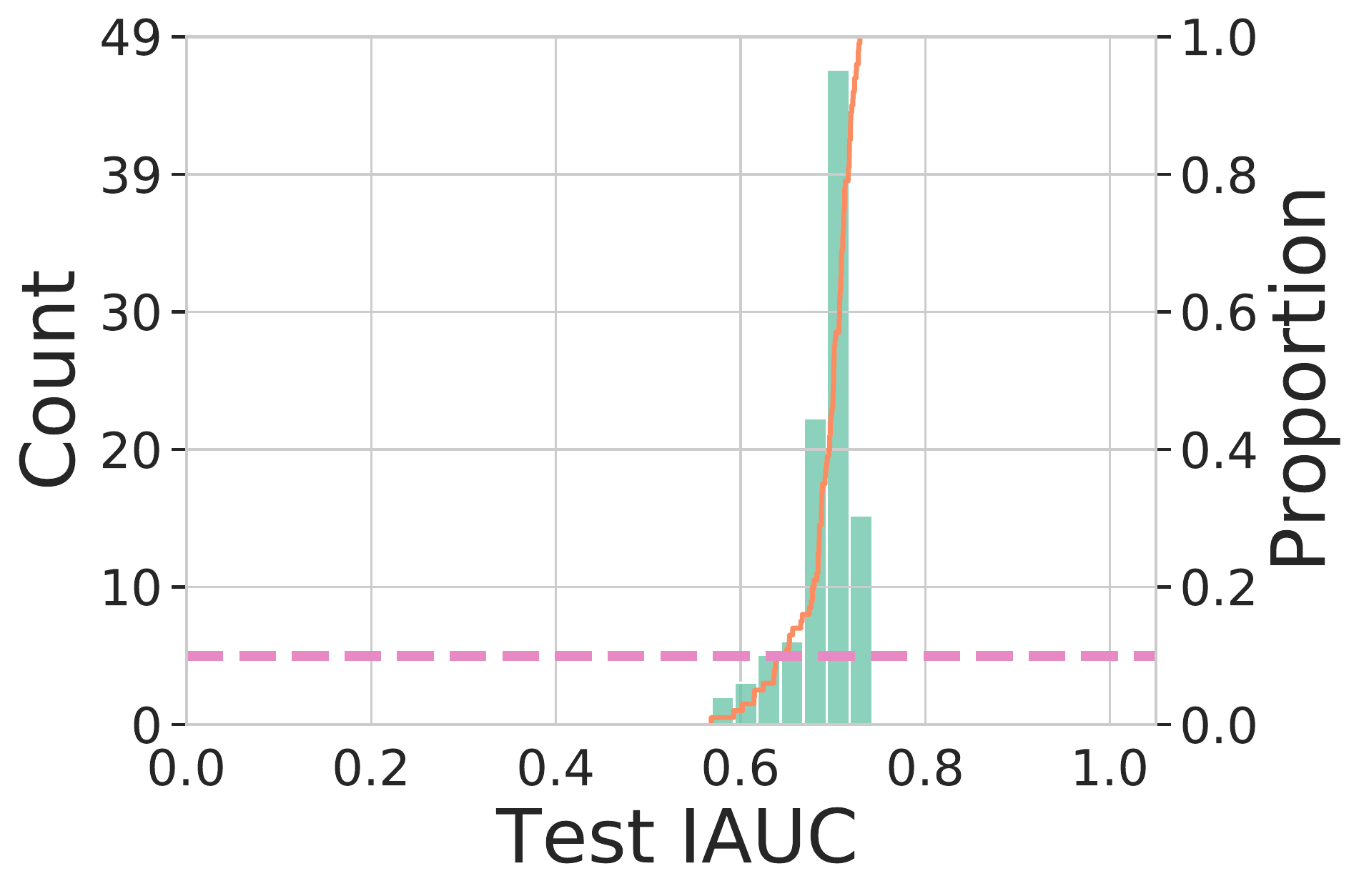}
\caption{Model 2}
\end{subfigure}
\begin{subfigure}{0.18\textwidth}
\centering

\includegraphics[width=\textwidth]{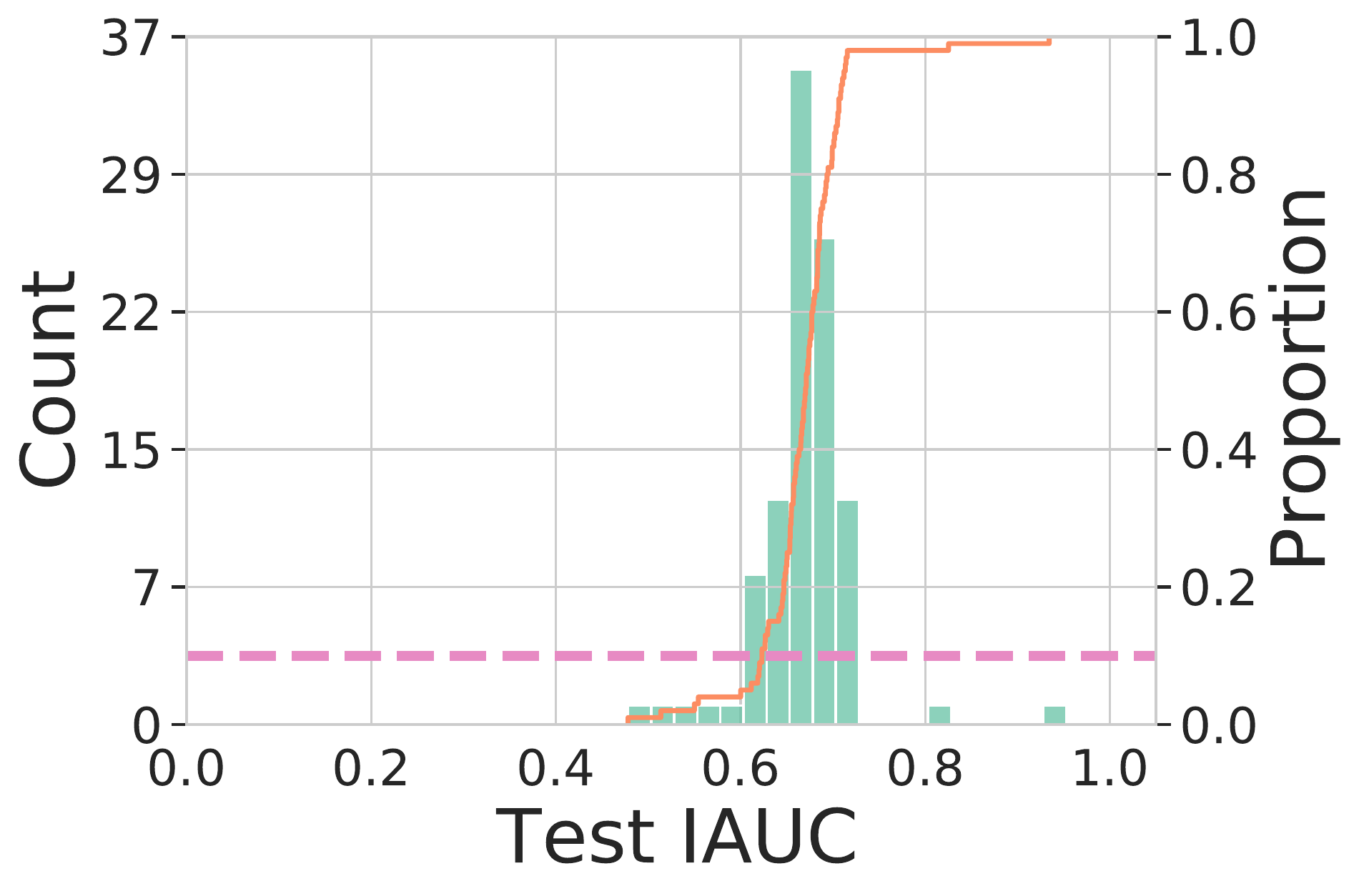}
\caption{Model 3}
\end{subfigure}
\begin{subfigure}{0.18\textwidth}
\centering

\includegraphics[width=\textwidth]{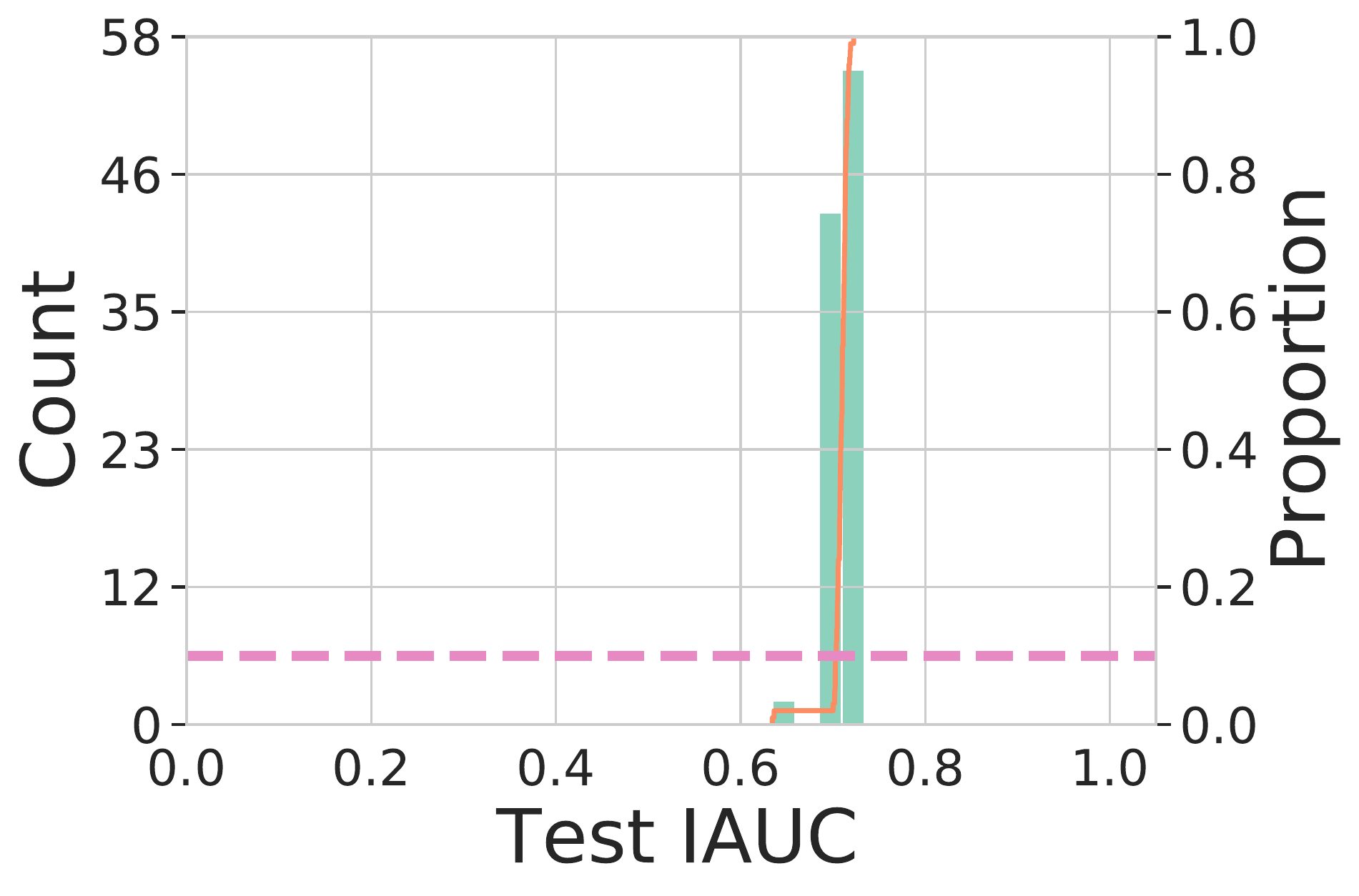}
\caption{Model 4}
\end{subfigure}
\begin{subfigure}{0.18\textwidth}
\centering

\includegraphics[width=\textwidth]{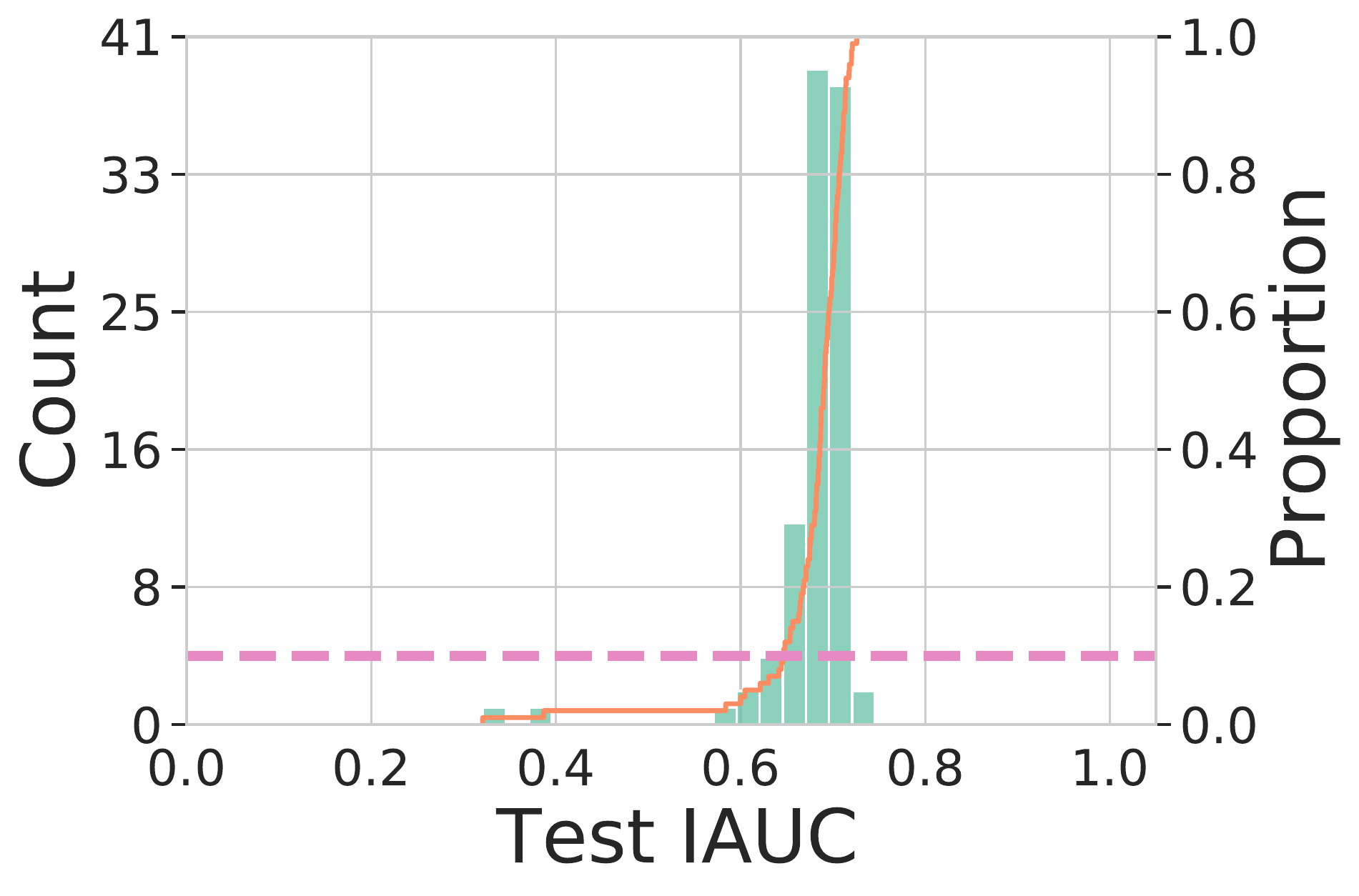}
\caption{Model 5 *}
\end{subfigure}
\caption[]{MNIST AND\label{fig:mnist_and_auroc_dist_top_models} }
\end{figure}


\begin{figure}[hb]
\centering
\begin{subfigure}{0.18\textwidth}
\centering
 
\includegraphics[width=\textwidth]{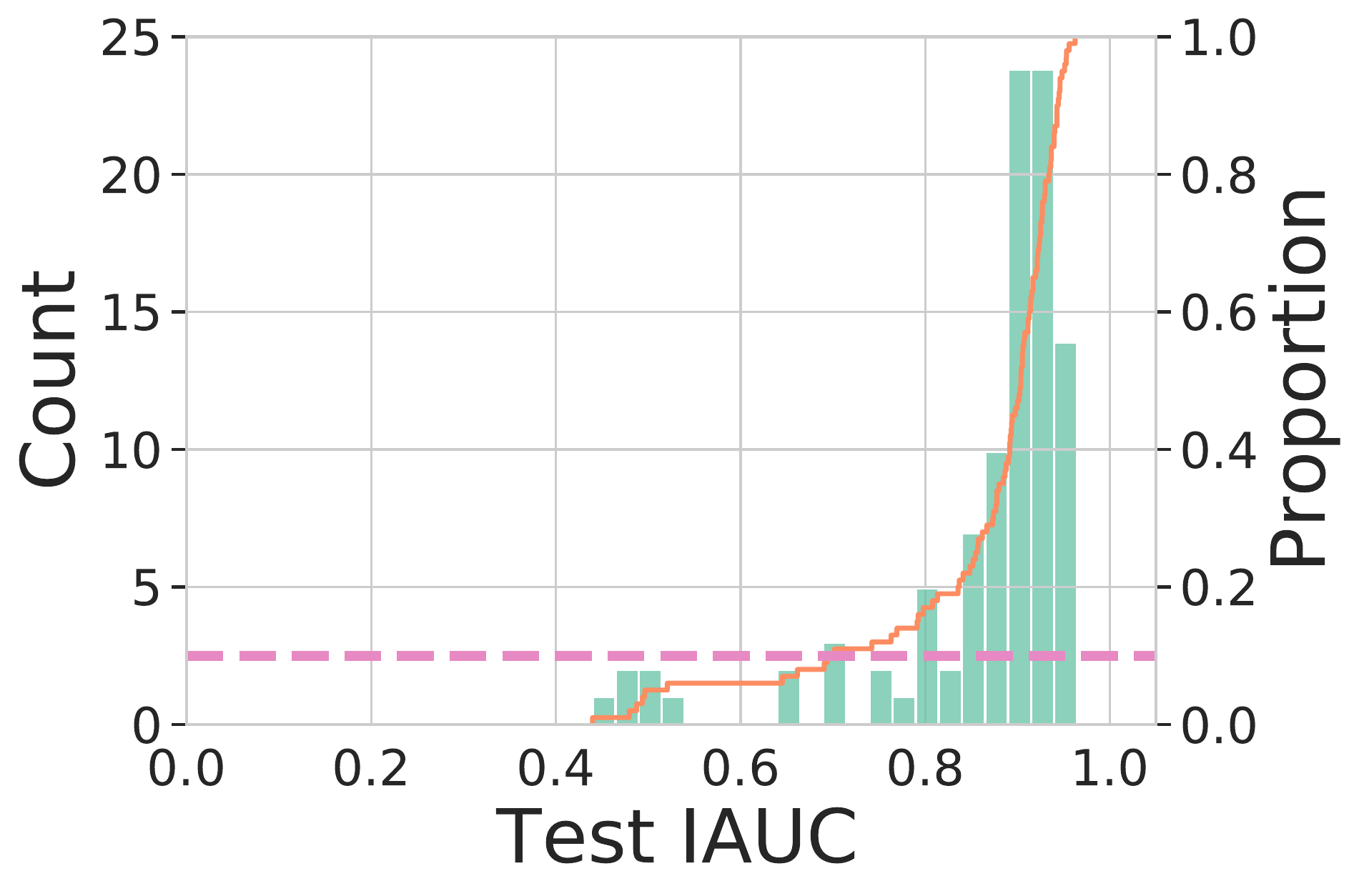}
\caption{Model 1}
\end{subfigure}
\begin{subfigure}{0.18\textwidth}
\centering
 
\includegraphics[width=\textwidth]{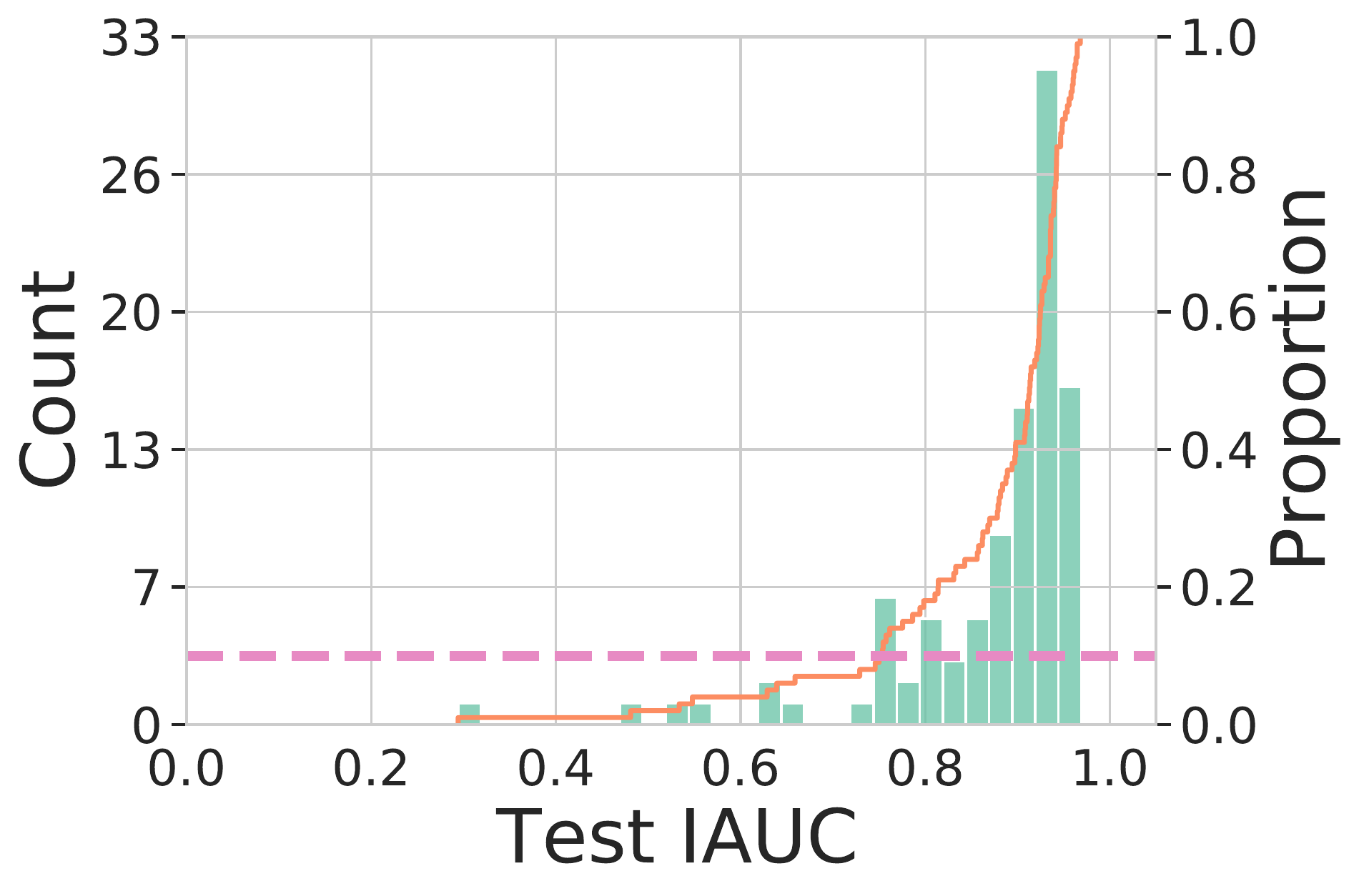}
\caption{Model 2}
\end{subfigure}
\begin{subfigure}{0.18\textwidth}
\centering

\includegraphics[width=\textwidth]{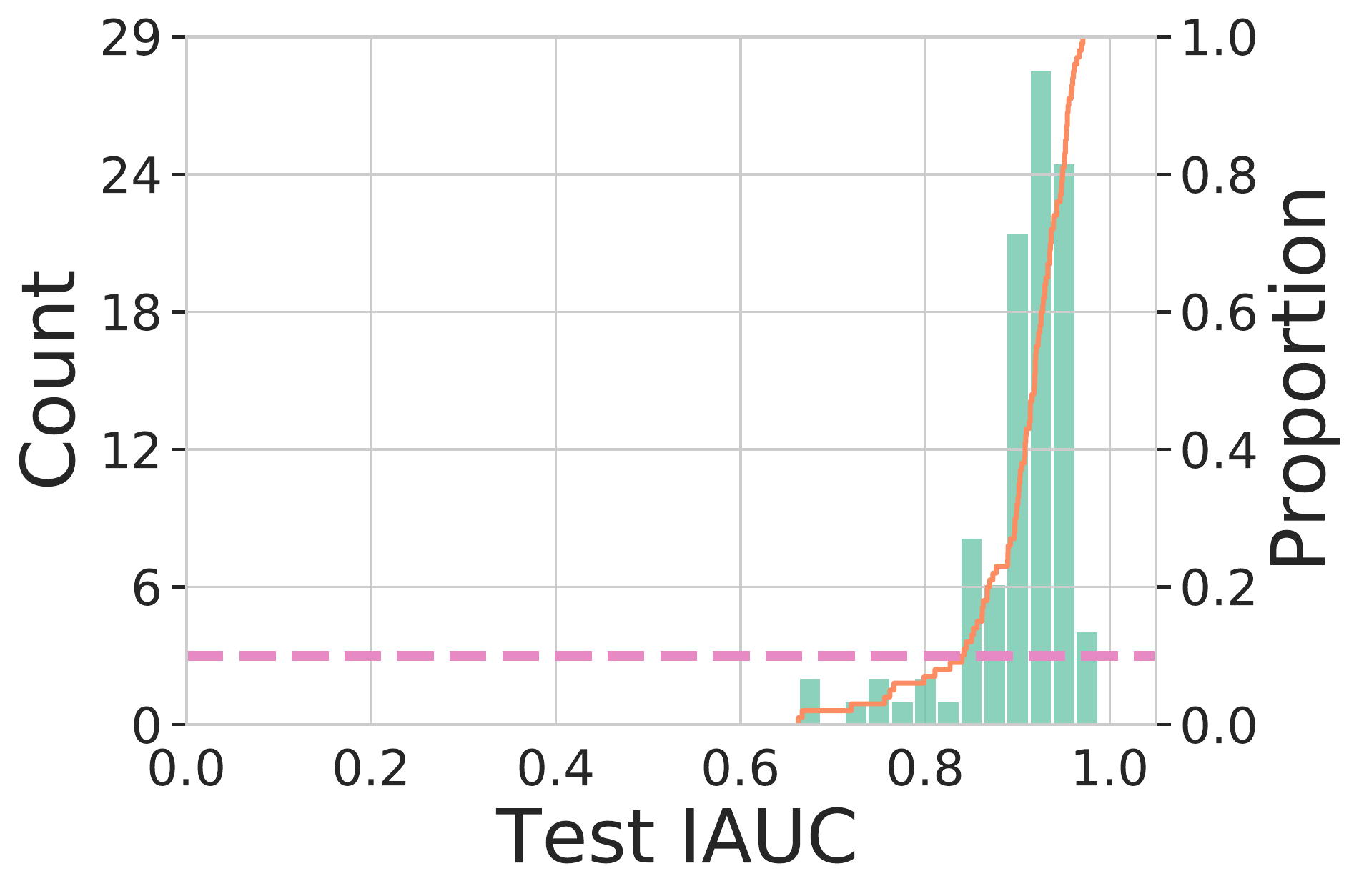}
\caption{Model 3}
\end{subfigure}
\begin{subfigure}{0.18\textwidth}
\centering

\includegraphics[width=\textwidth]{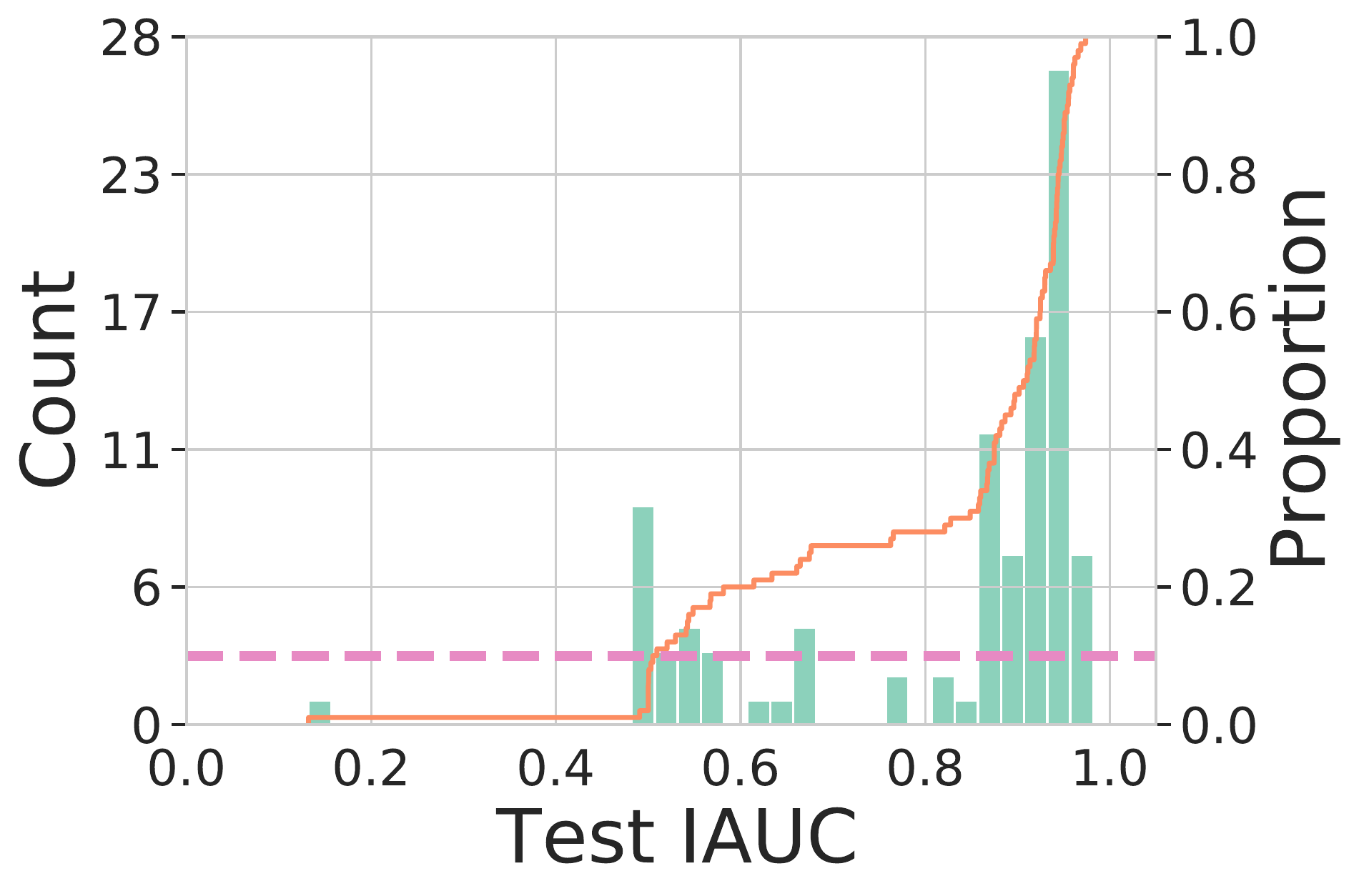}
\caption{Model 4 *}
\end{subfigure}
\begin{subfigure}{0.18\textwidth}
\centering

\includegraphics[width=\textwidth]{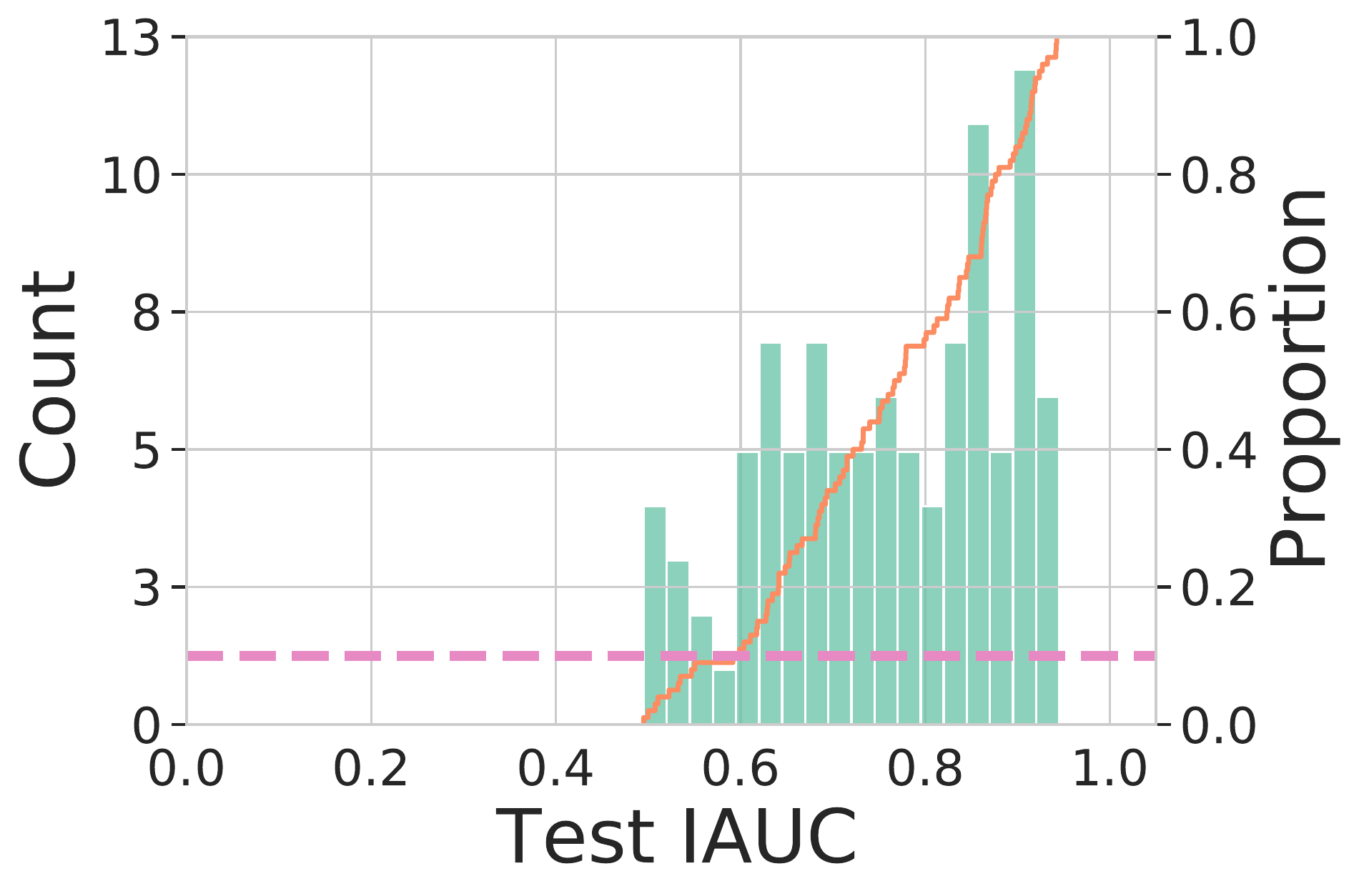}
\caption{Model 5 *}
\end{subfigure}
\caption[]{MNIST XOR\label{fig:mnist_xor_auroc_dist_top_models} }
\end{figure}

\begin{figure}[hb]
\centering
\begin{subfigure}{0.18\textwidth}
\centering

\includegraphics[width=\textwidth]{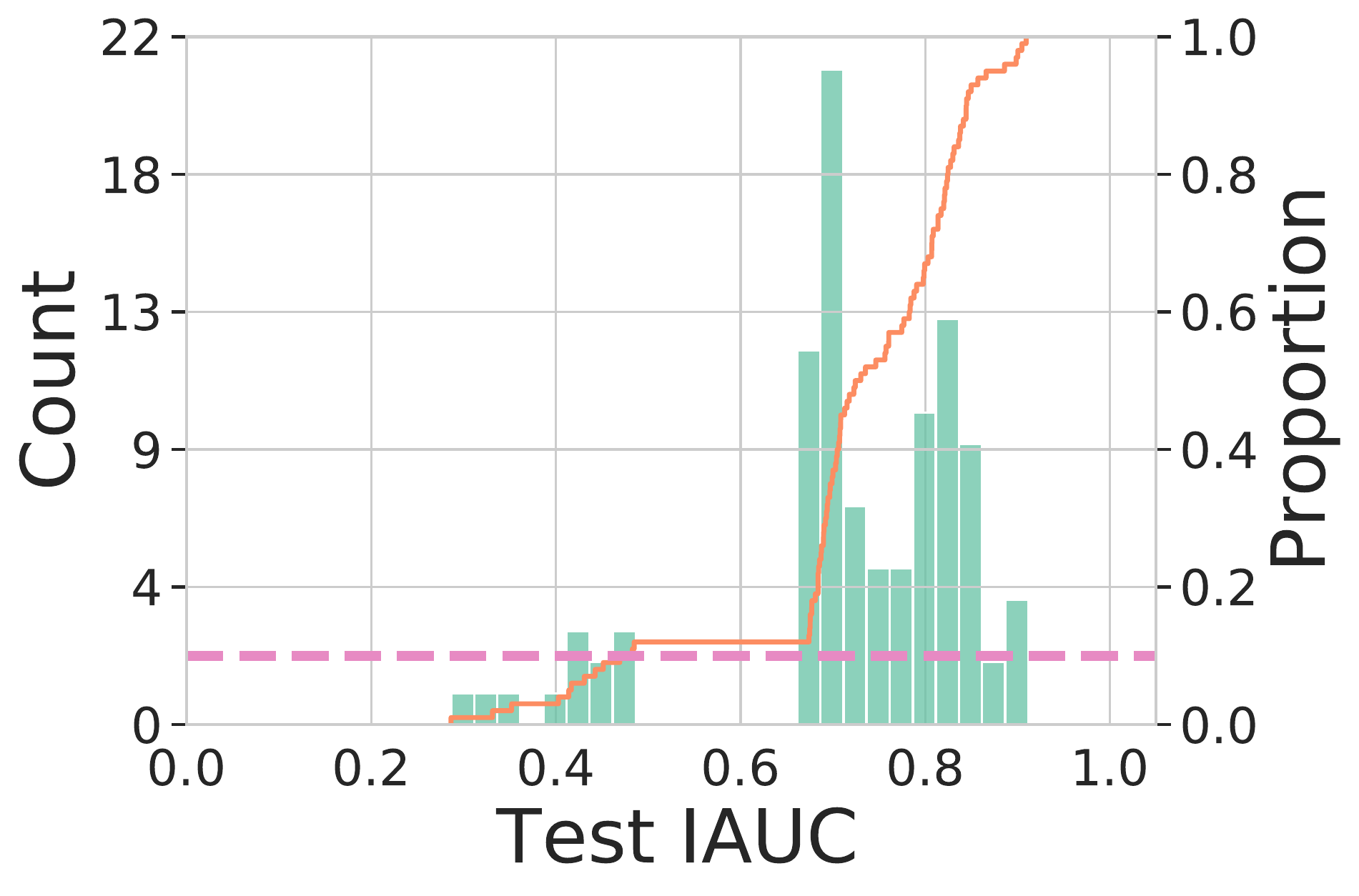}
\caption{Model 1 *}
\end{subfigure}
\begin{subfigure}{0.18\textwidth}
\centering

\includegraphics[width=\textwidth]{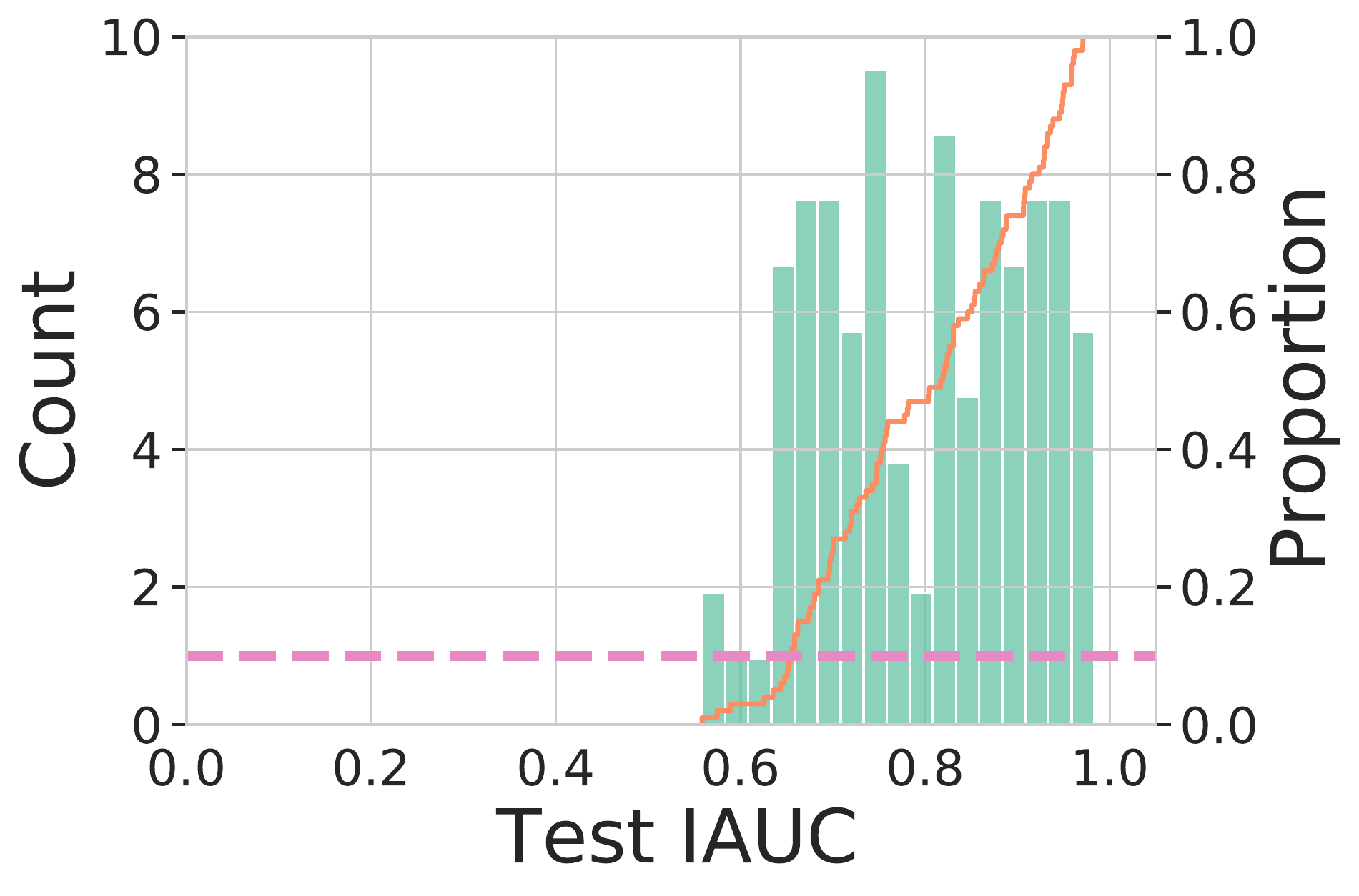}
\caption{Model 2}
\end{subfigure}
\begin{subfigure}{0.18\textwidth}
\centering

\includegraphics[width=\textwidth]{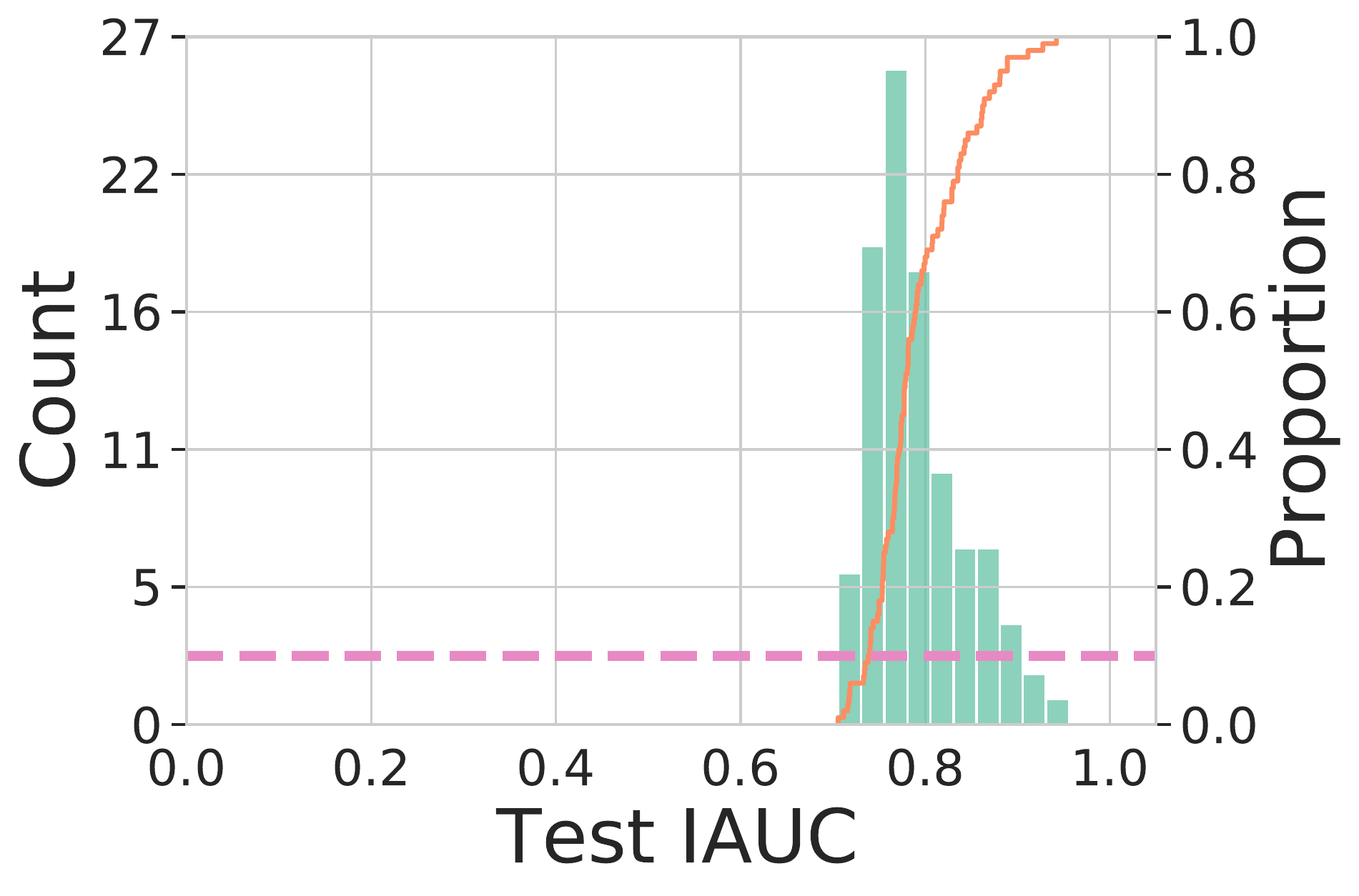}
\caption{Model 3}
\end{subfigure}
\begin{subfigure}{0.18\textwidth}
\centering

\includegraphics[width=\textwidth]{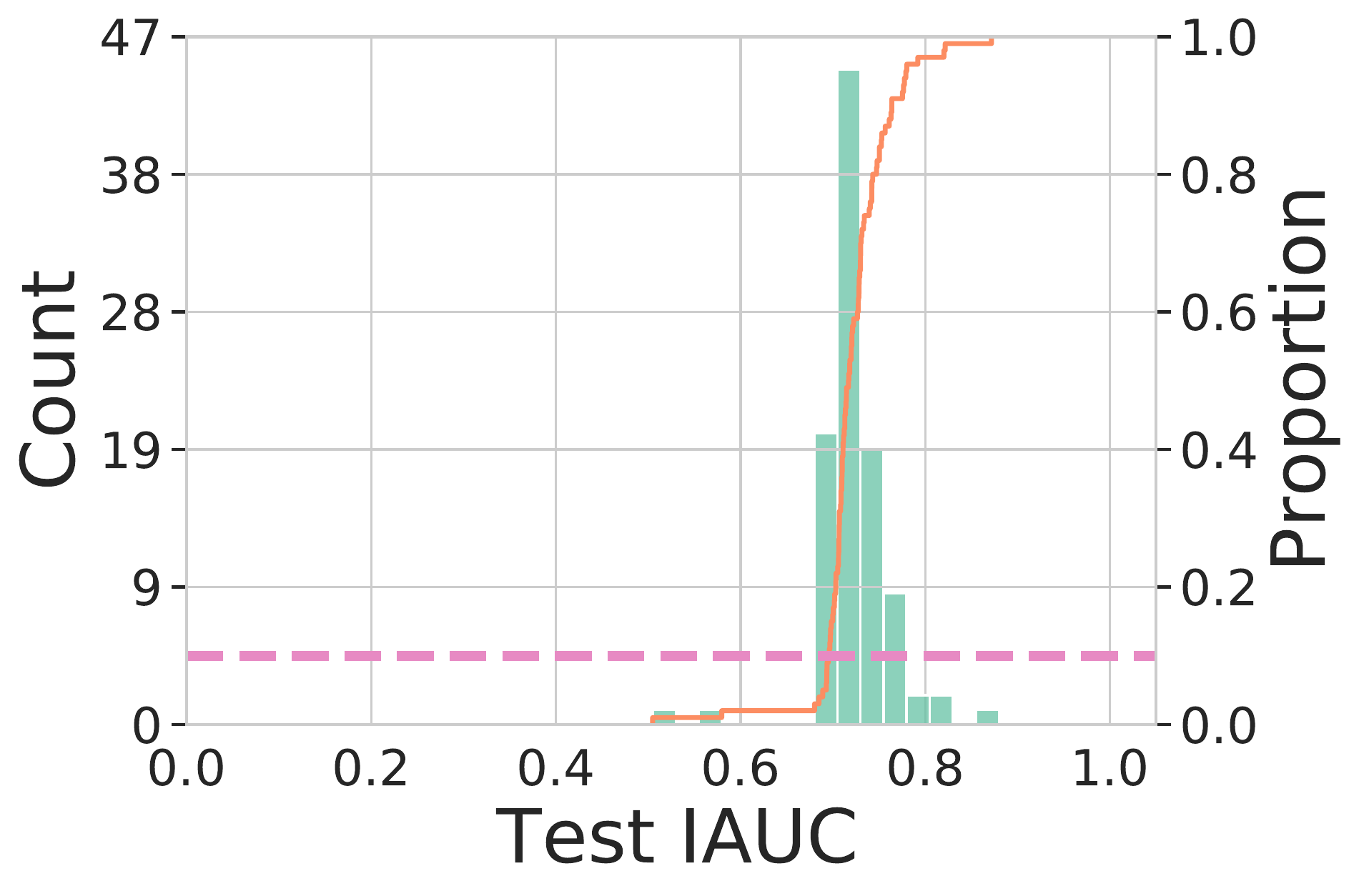}
\caption{Model 4}
\end{subfigure}
\begin{subfigure}{0.18\textwidth}
\centering

\includegraphics[width=\textwidth]{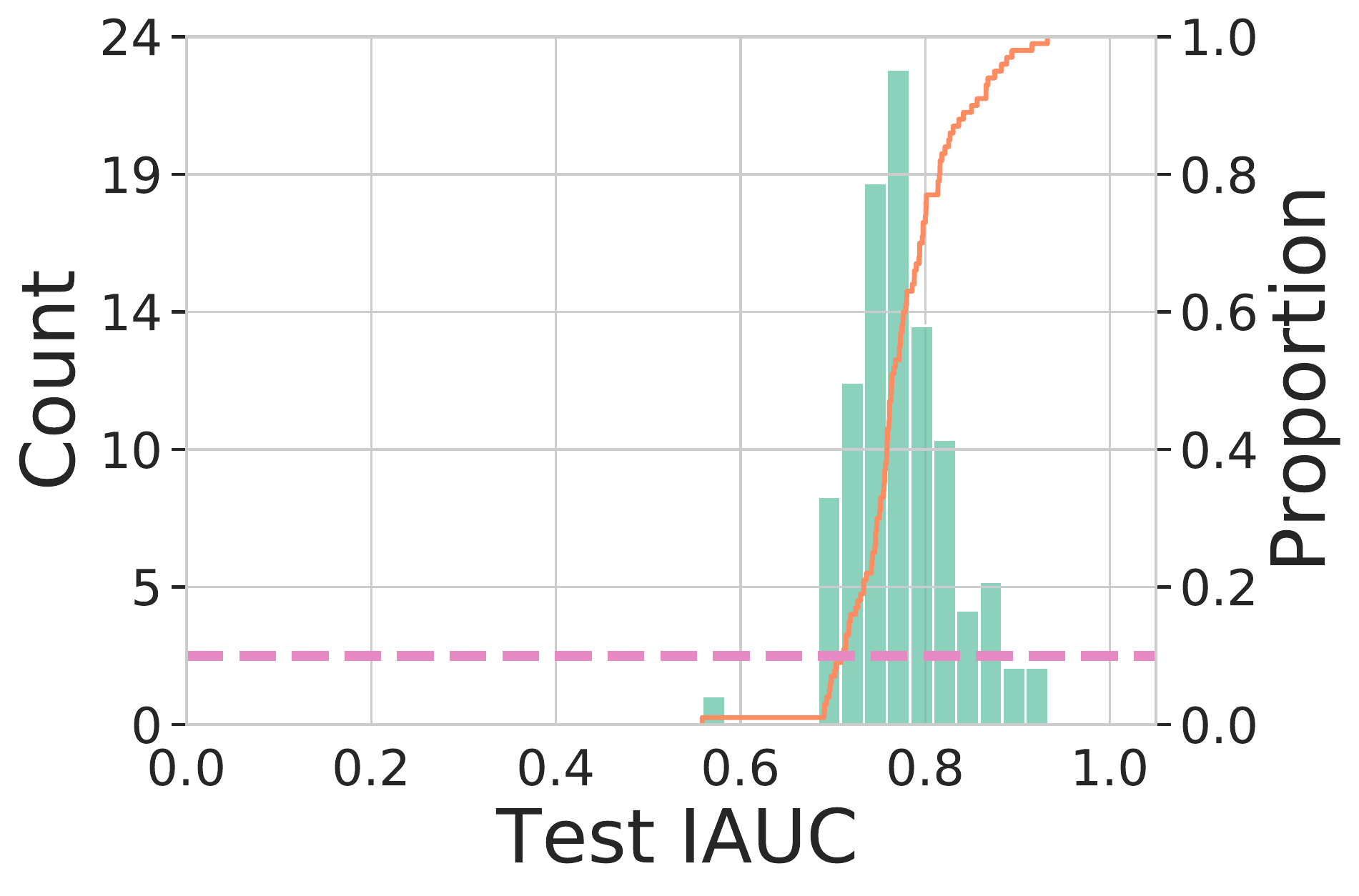}
\caption{Model 5}
\end{subfigure}
\caption[]{CyTOF MIL\label{fig:cytof_mil_auroc_dist_top_models} }
\end{figure}

\begin{figure}[hb]
\centering
\begin{subfigure}{0.18\textwidth}
\centering

\includegraphics[width=\textwidth]{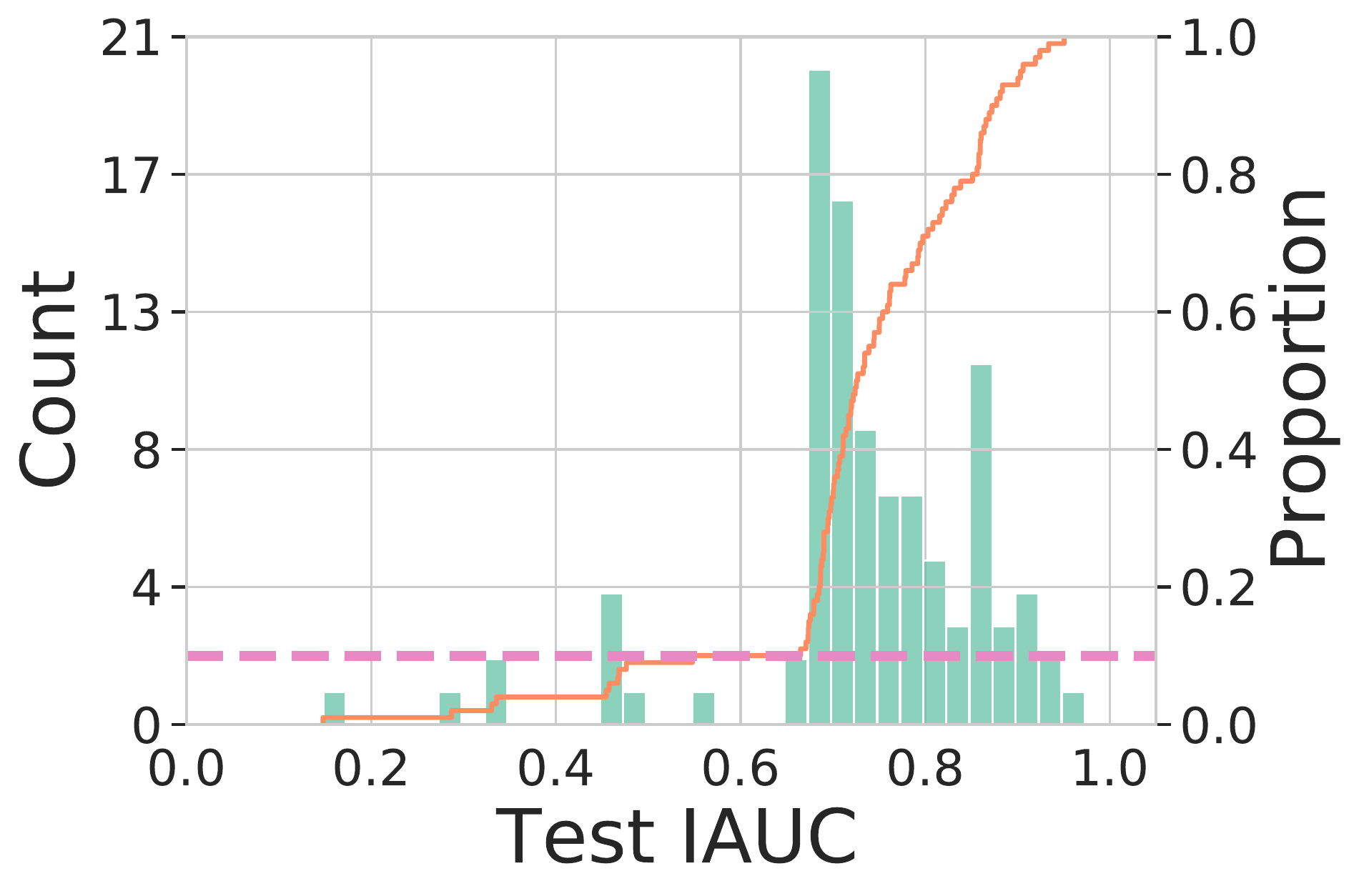}
\caption{Model 1 *}
\end{subfigure}
\begin{subfigure}{0.18\textwidth}
\centering

\includegraphics[width=\textwidth]{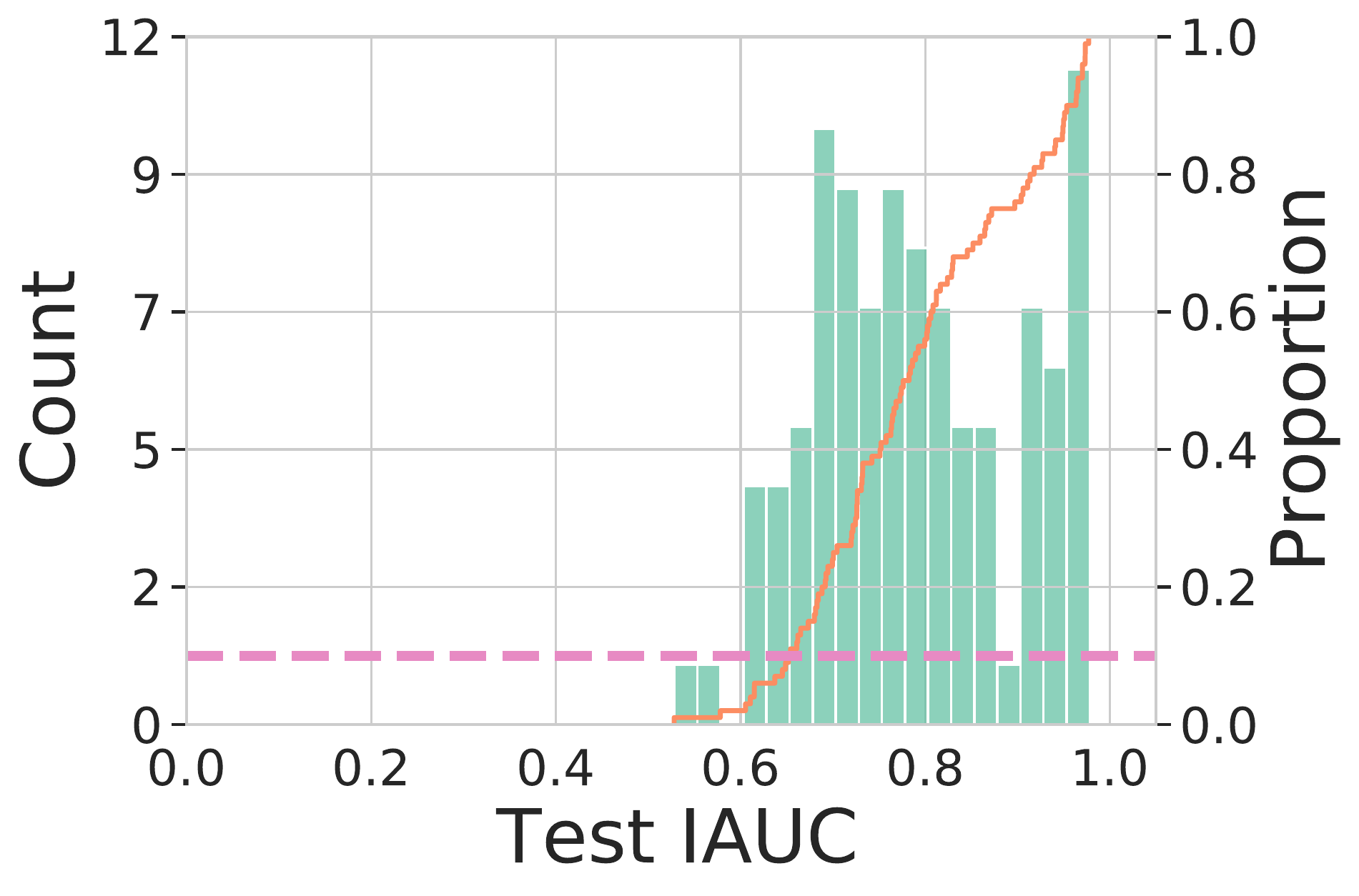}
\caption{Model 2 *}
\end{subfigure}
\begin{subfigure}{0.18\textwidth}
\centering

\includegraphics[width=\textwidth]{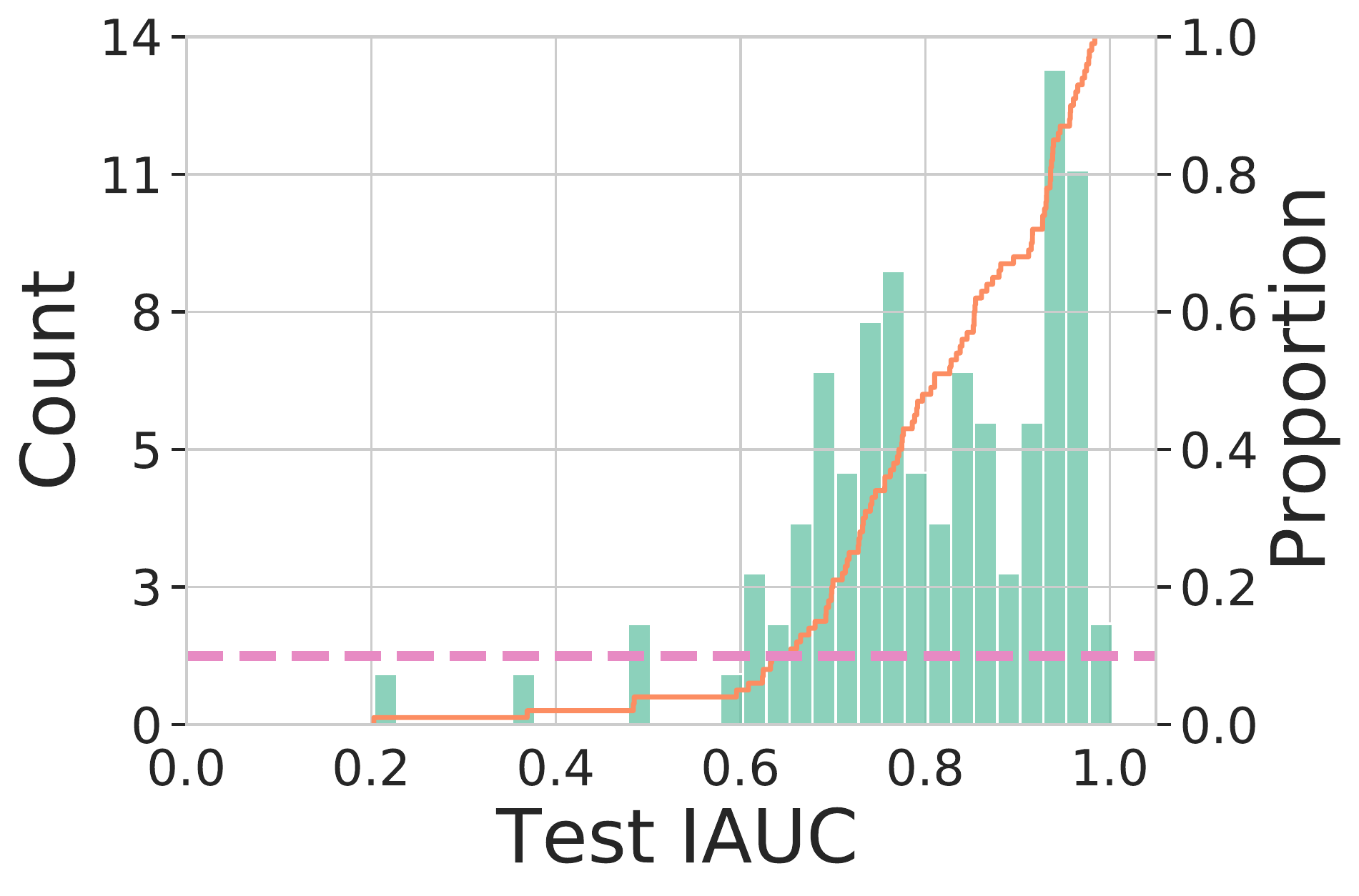}
\caption{Model 3 *}
\end{subfigure}
\begin{subfigure}{0.18\textwidth}
\centering

\includegraphics[width=\textwidth]{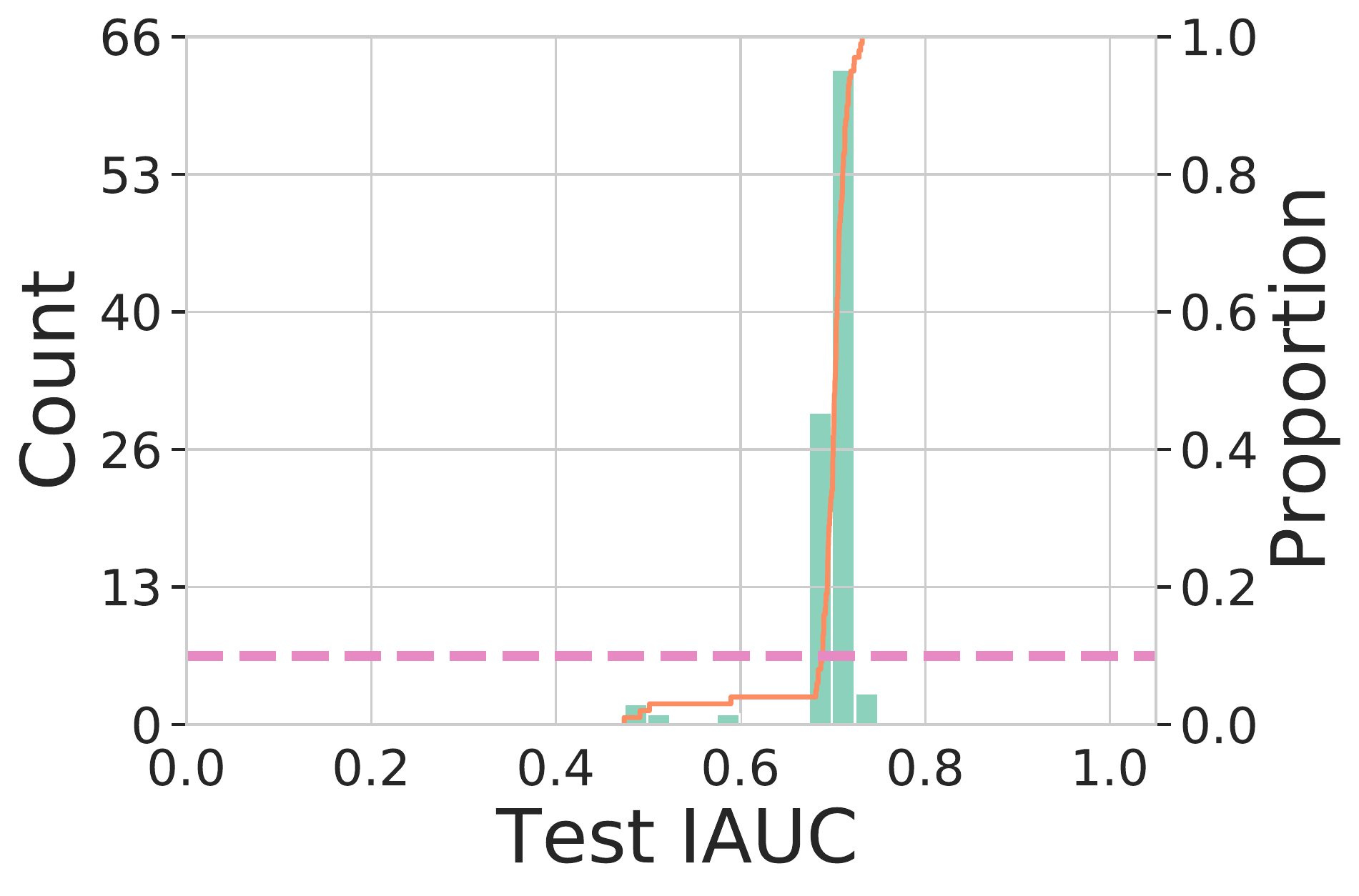}
\caption{Model 4}
\end{subfigure}
\begin{subfigure}{0.18\textwidth}
\centering

\includegraphics[width=\textwidth]{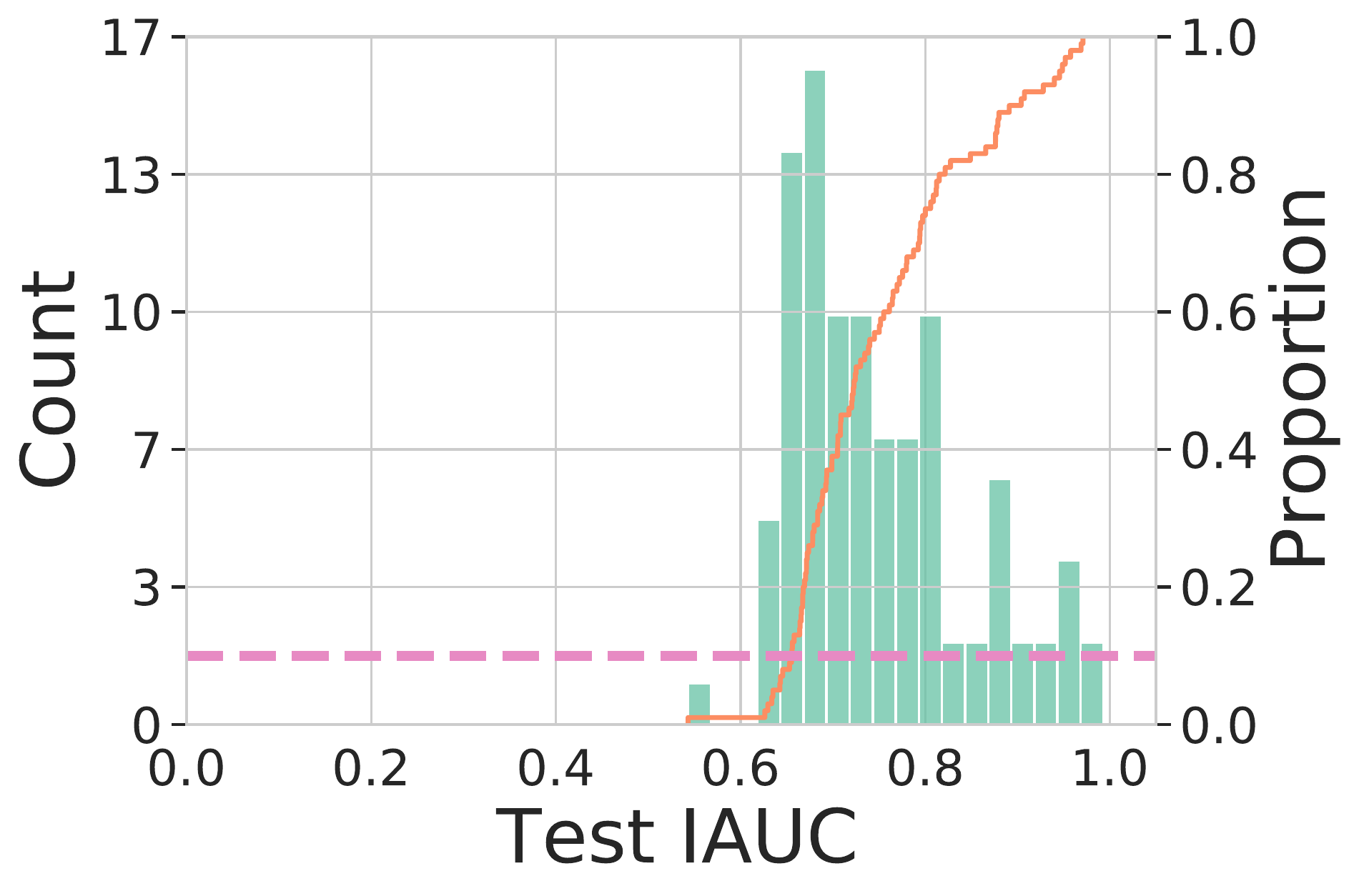}
\caption{Model 5}
\end{subfigure}
\caption[]{CyTOF AND\label{fig:cytof_and_auroc_dist_top_models} }
\end{figure}

\begin{figure}[hb]
\centering
\begin{subfigure}{0.18\textwidth}
\centering

\includegraphics[width=\textwidth]{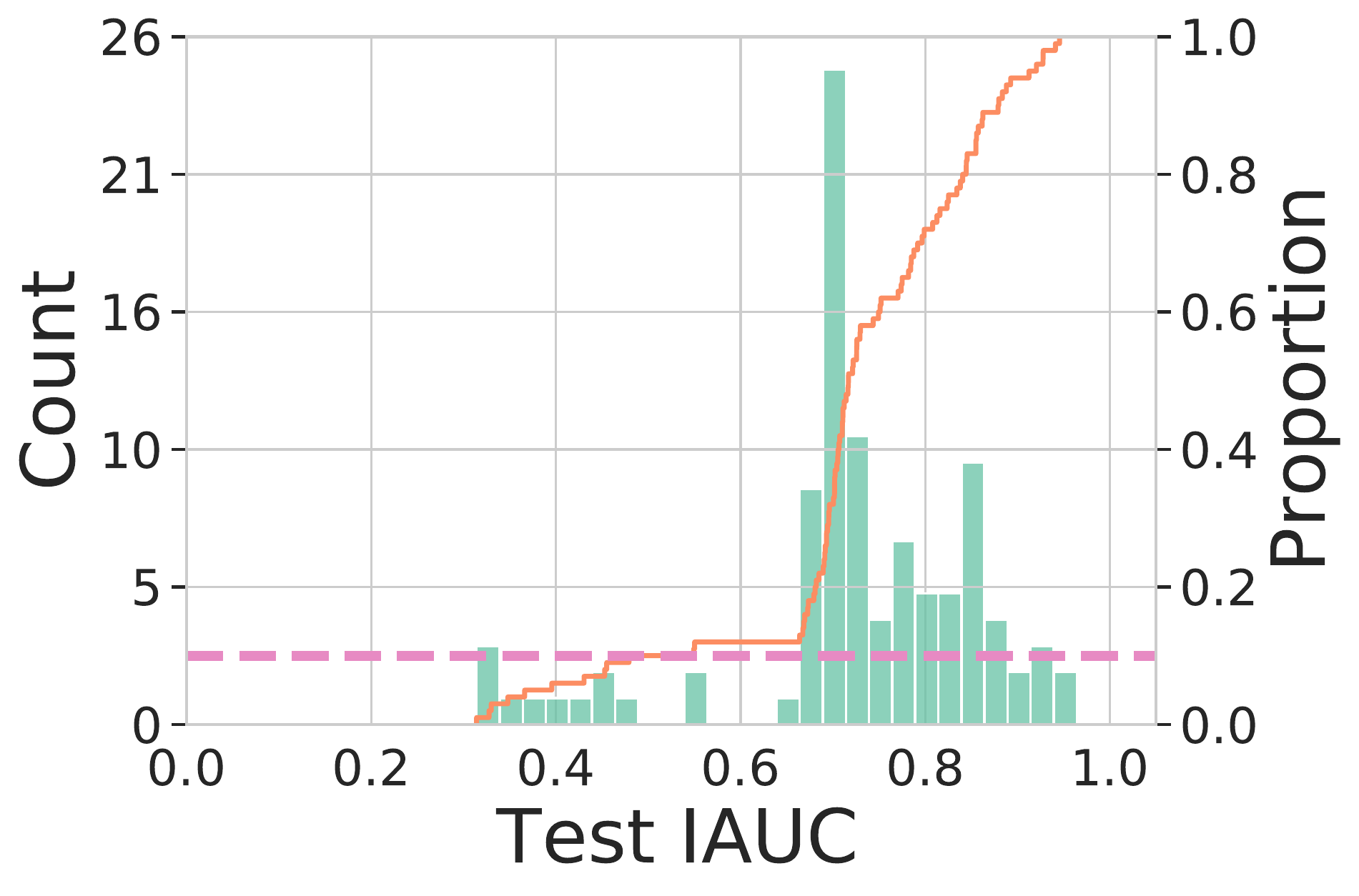}
\caption{Model 1 *}
\end{subfigure}
\begin{subfigure}{0.18\textwidth}
\centering

\includegraphics[width=\textwidth]{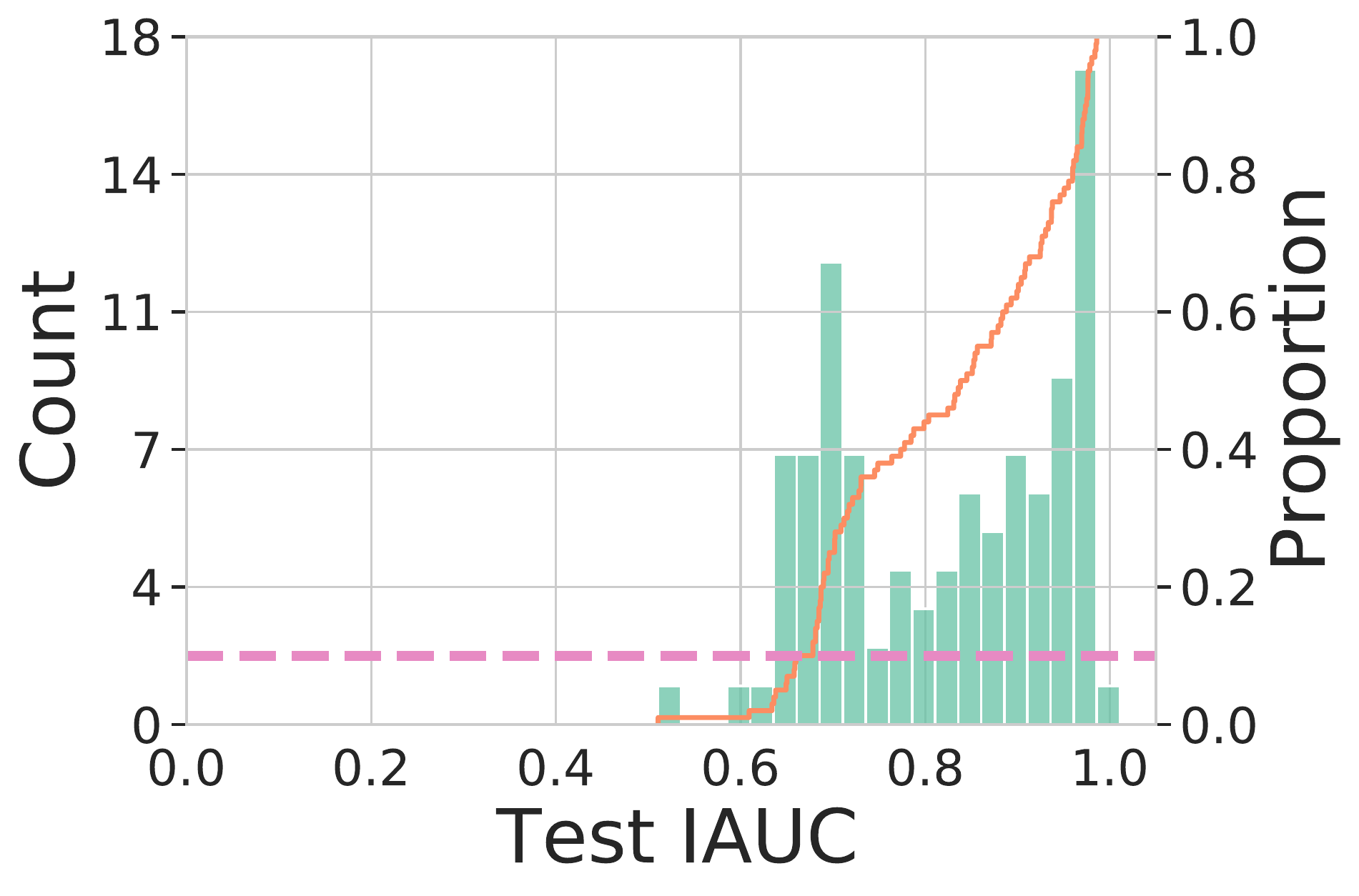}
\caption{Model 2}
\end{subfigure}
\begin{subfigure}{0.18\textwidth}
\centering

\includegraphics[width=\textwidth]{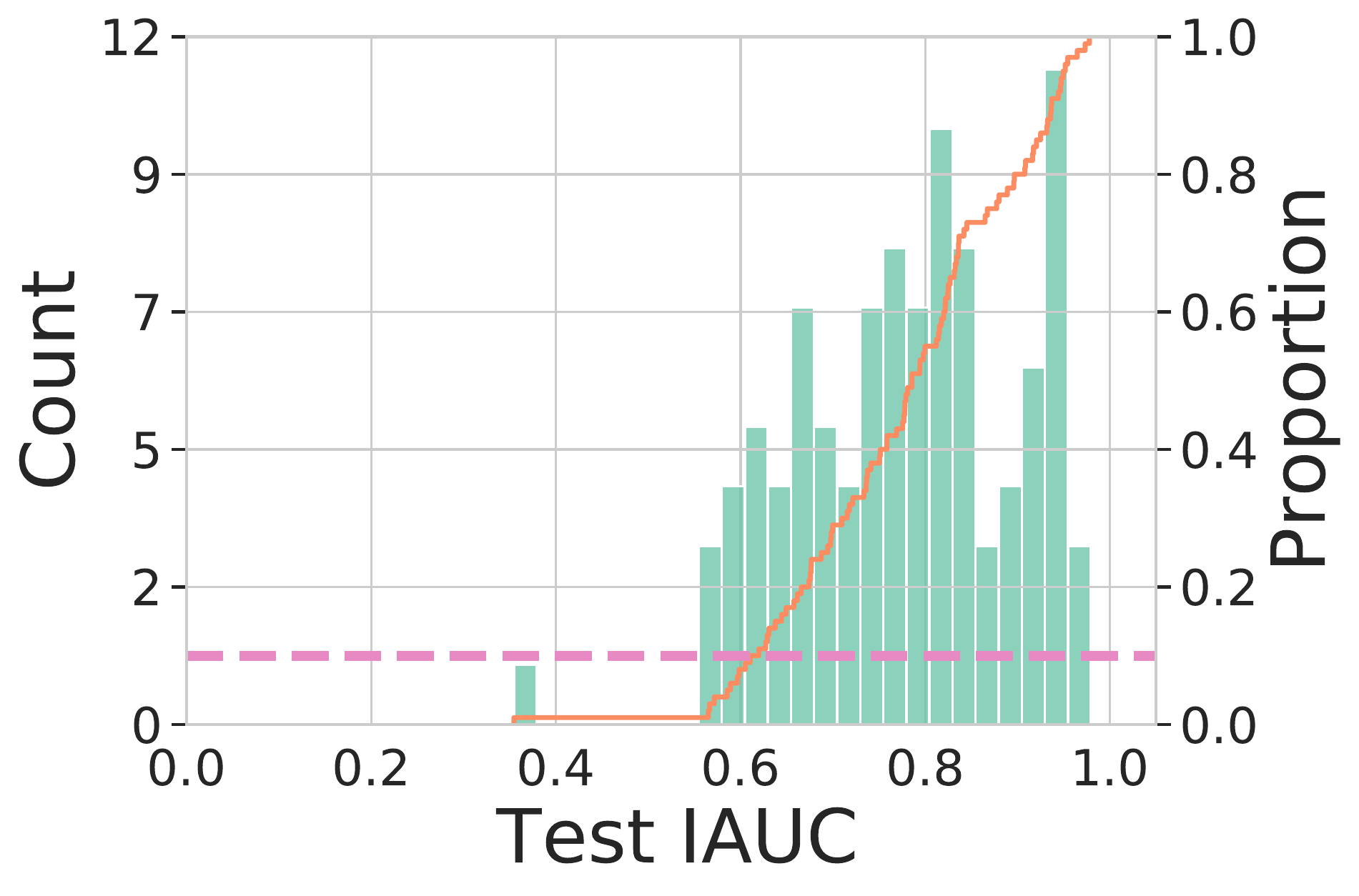}
\caption{Model 3 *}
\end{subfigure}
\begin{subfigure}{0.18\textwidth}
\centering

\includegraphics[width=\textwidth]{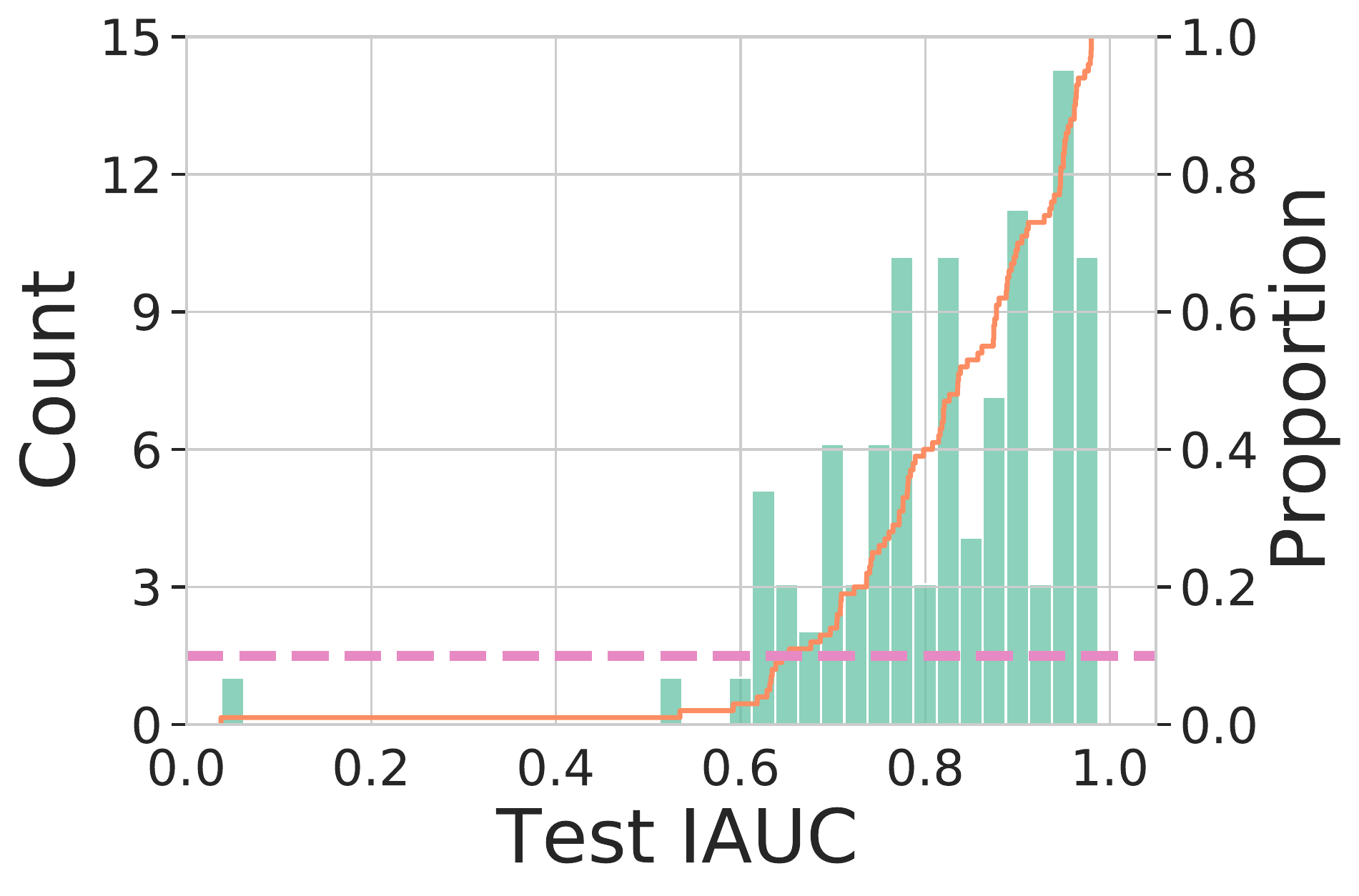}
\caption{Model 4 *}
\end{subfigure}
\begin{subfigure}{0.18\textwidth}
\centering

\includegraphics[width=\textwidth]{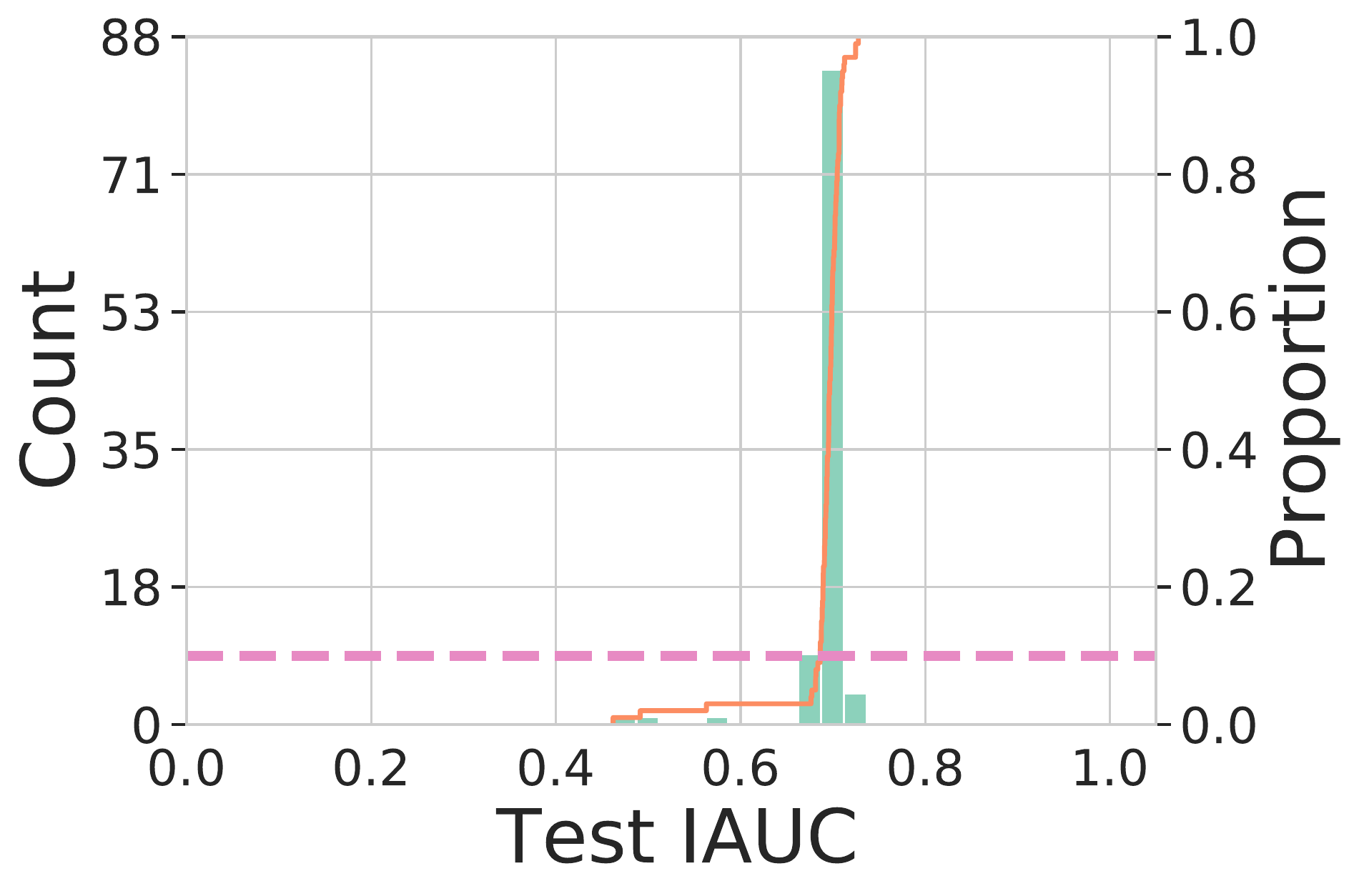}
\caption{Model 5}
\end{subfigure}
\caption[]{CyTOF XOR\label{fig:cytof_and_auroc_dist_top_models} }
\end{figure}


\clearpage
\section{Correlations between IAUC and Accuracy}
\label{sec:corr-iauc-acc}
\begin{figure*}[h]
\begin{multicols}{3}
  \centering
\begin{subfigure}{0.31\textwidth}
\includegraphics[width=\textwidth]{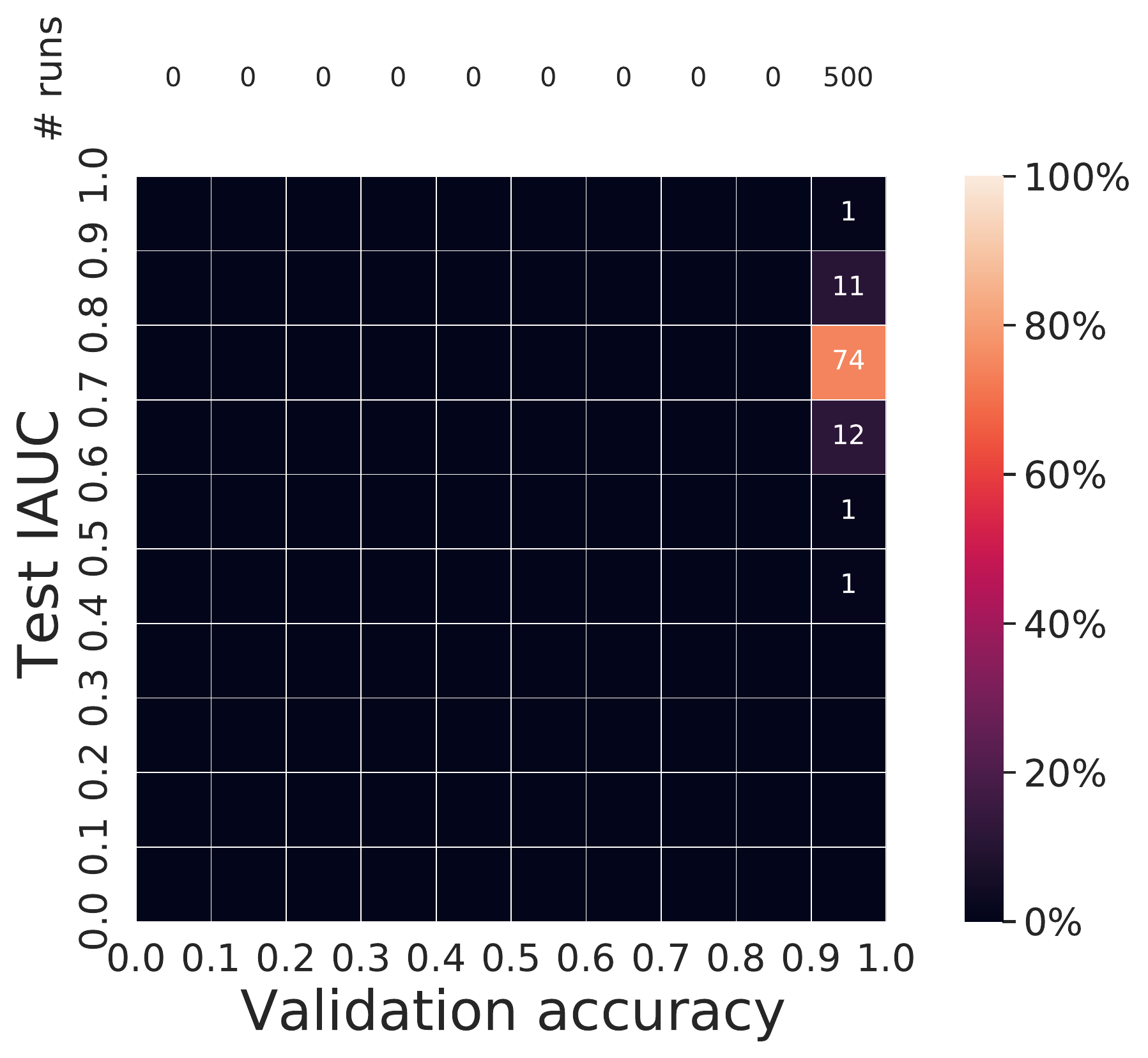}
\caption{Gaussian MIL\label{fig:acc_vs_auroc_mil}}
\end{subfigure}
\begin{subfigure}{0.31\textwidth}
\includegraphics[width=\textwidth]{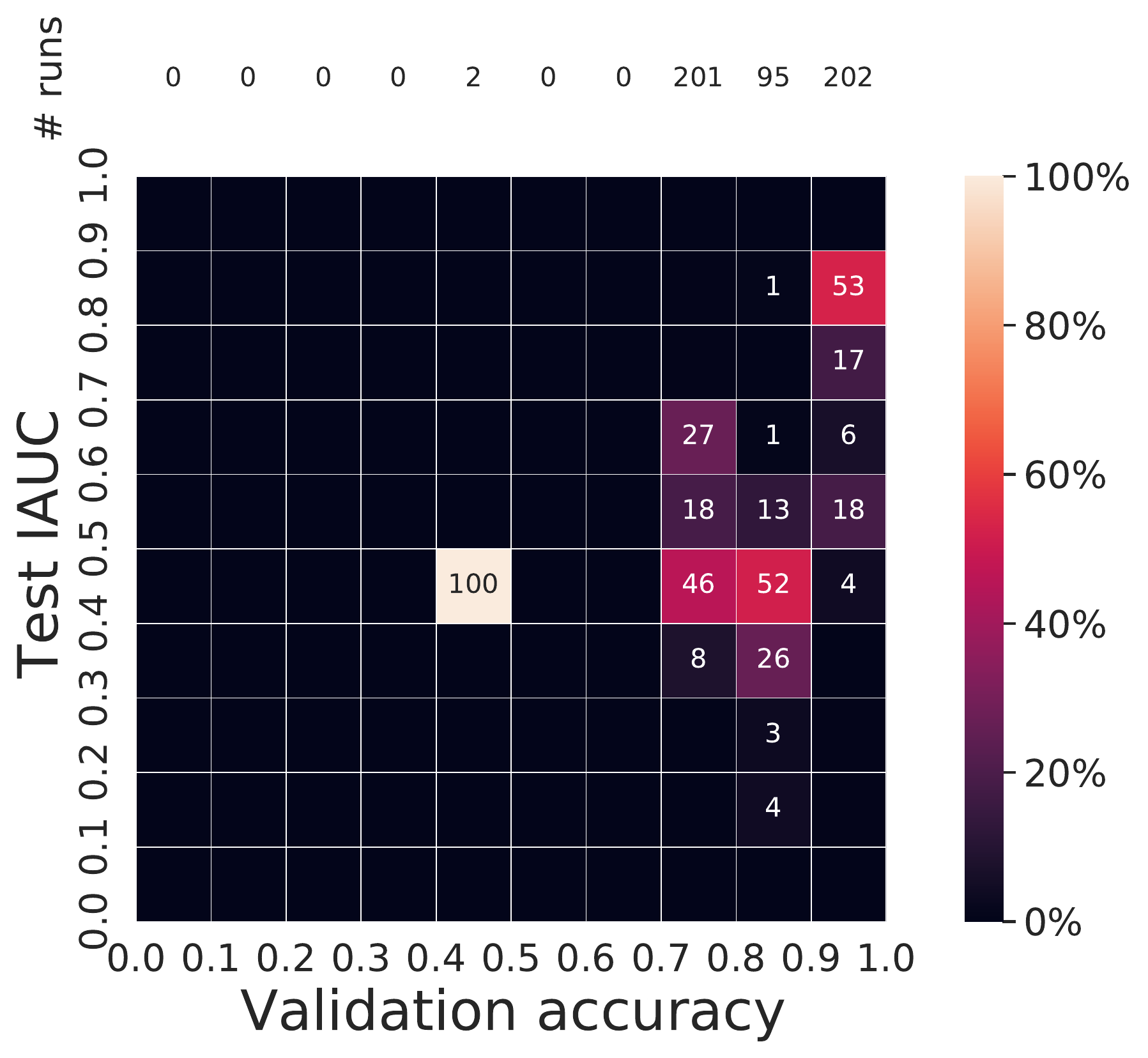}
\caption{Gaussian AND\label{fig:acc_vs_auroc_and}}
\end{subfigure}
\begin{subfigure}{0.31\textwidth}
\includegraphics[width=\textwidth]{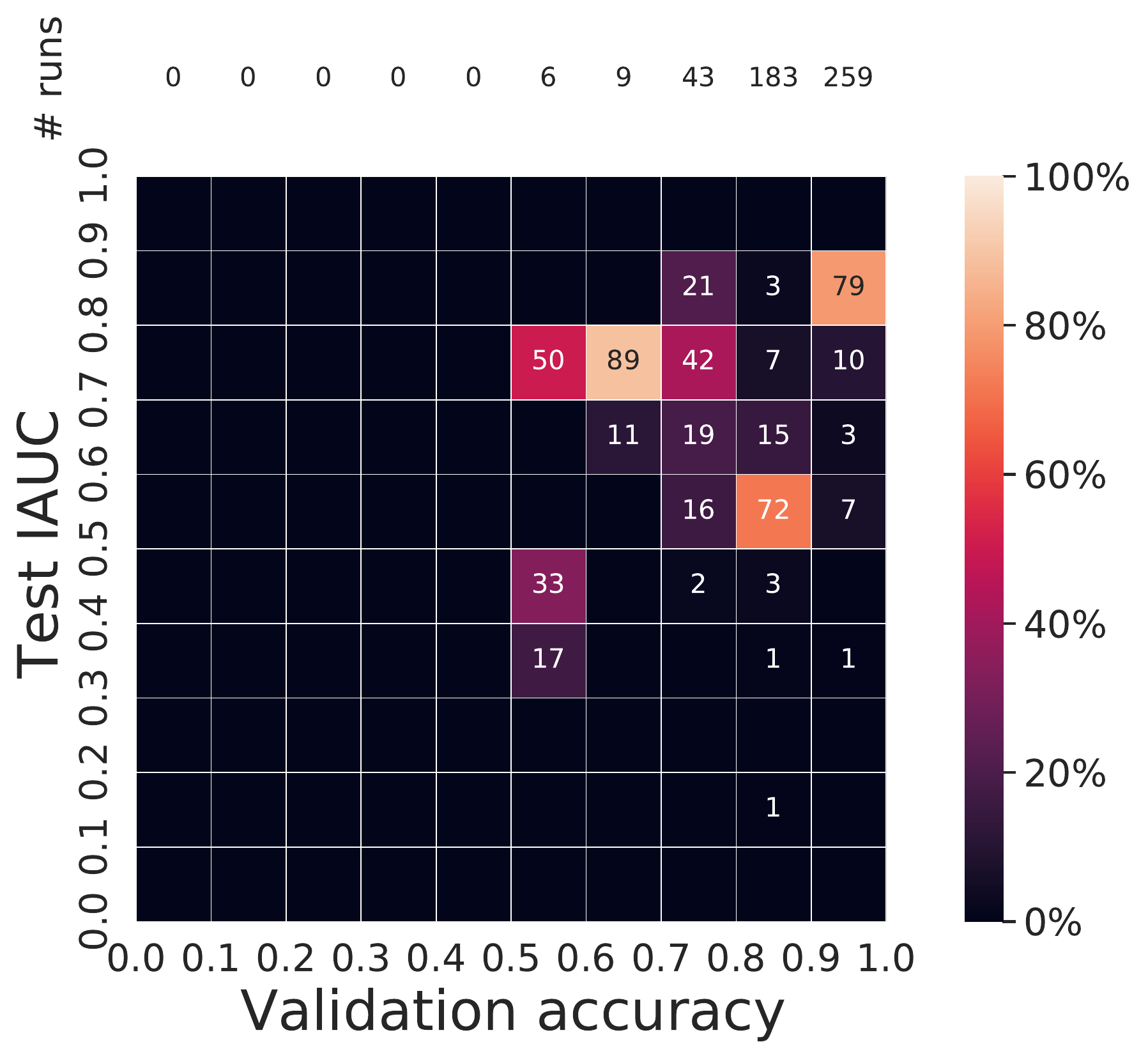}
\caption{Gaussian XOR\label{fig:acc_vs_auroc_xor}}
\end{subfigure}
\begin{subfigure}{0.31\textwidth}
\includegraphics[width=\textwidth]{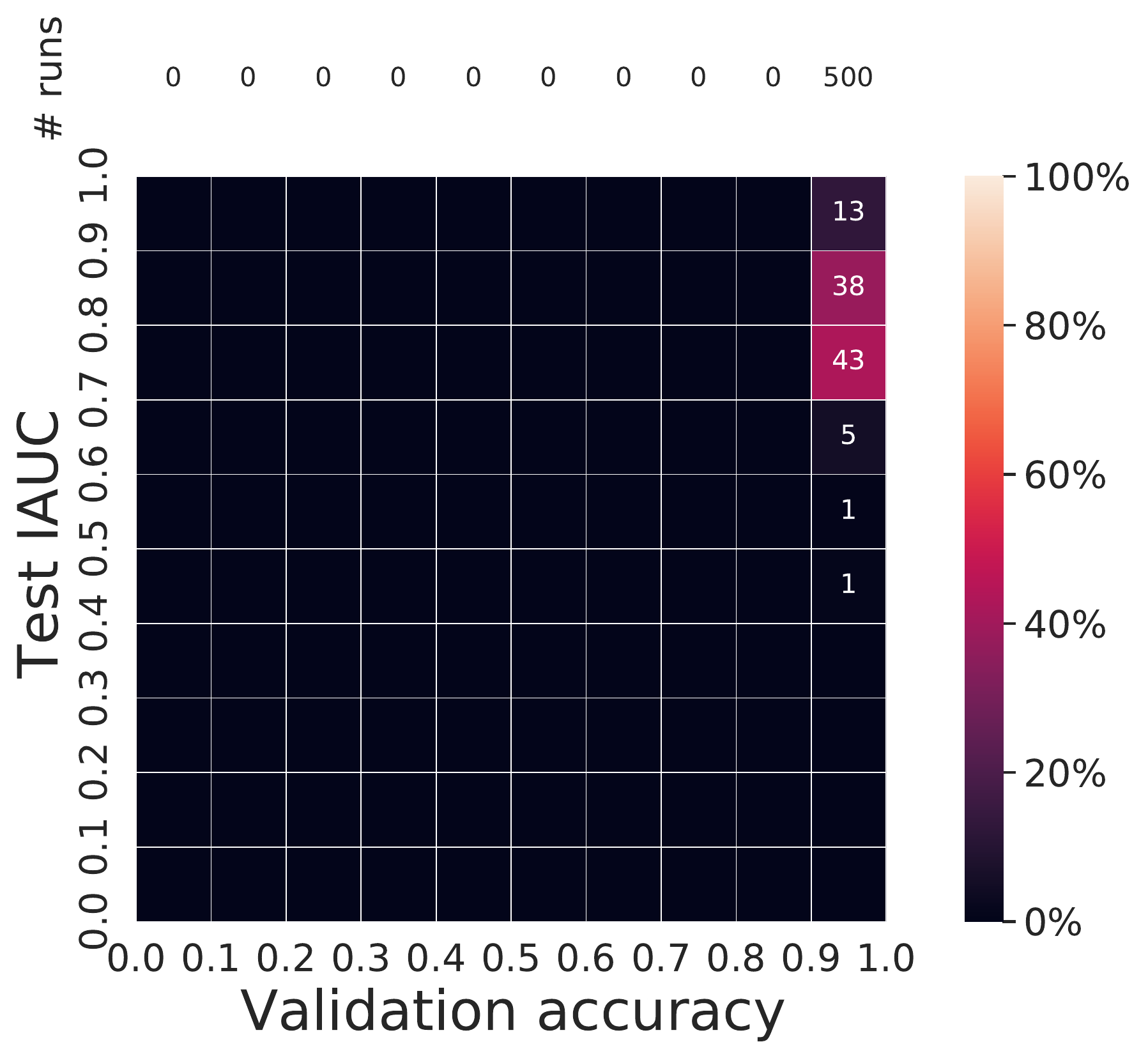}
\caption{MNIST MIL\label{fig:acc_vs_auroc_mnist_mil}}
\end{subfigure}
\begin{subfigure}{0.31\textwidth}
\includegraphics[width=\textwidth]{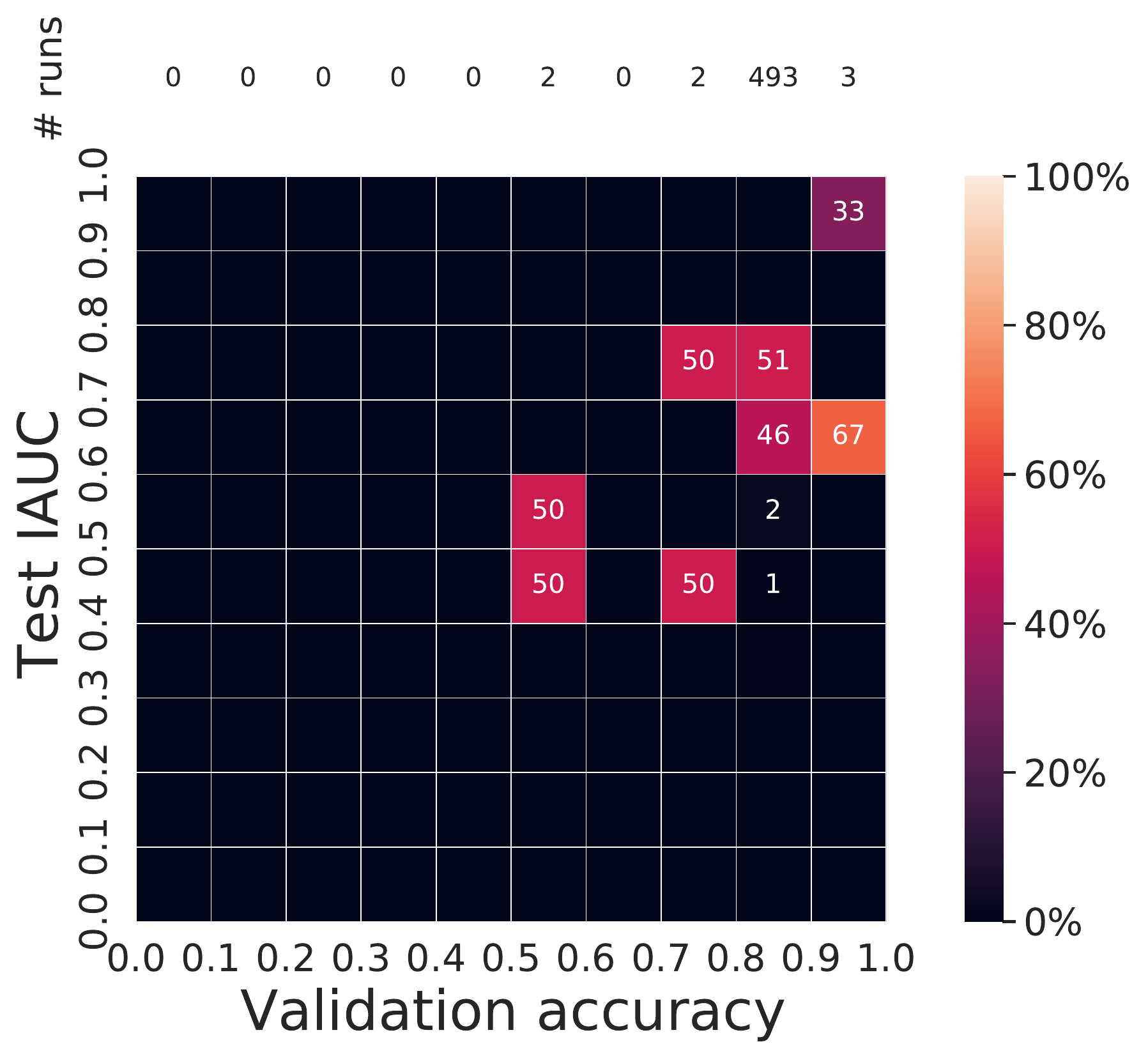}
\caption{MNIST AND\label{fig:acc_vs_auroc_mnist_and}}
\end{subfigure}
\begin{subfigure}{0.31\textwidth}
\includegraphics[width=\textwidth]{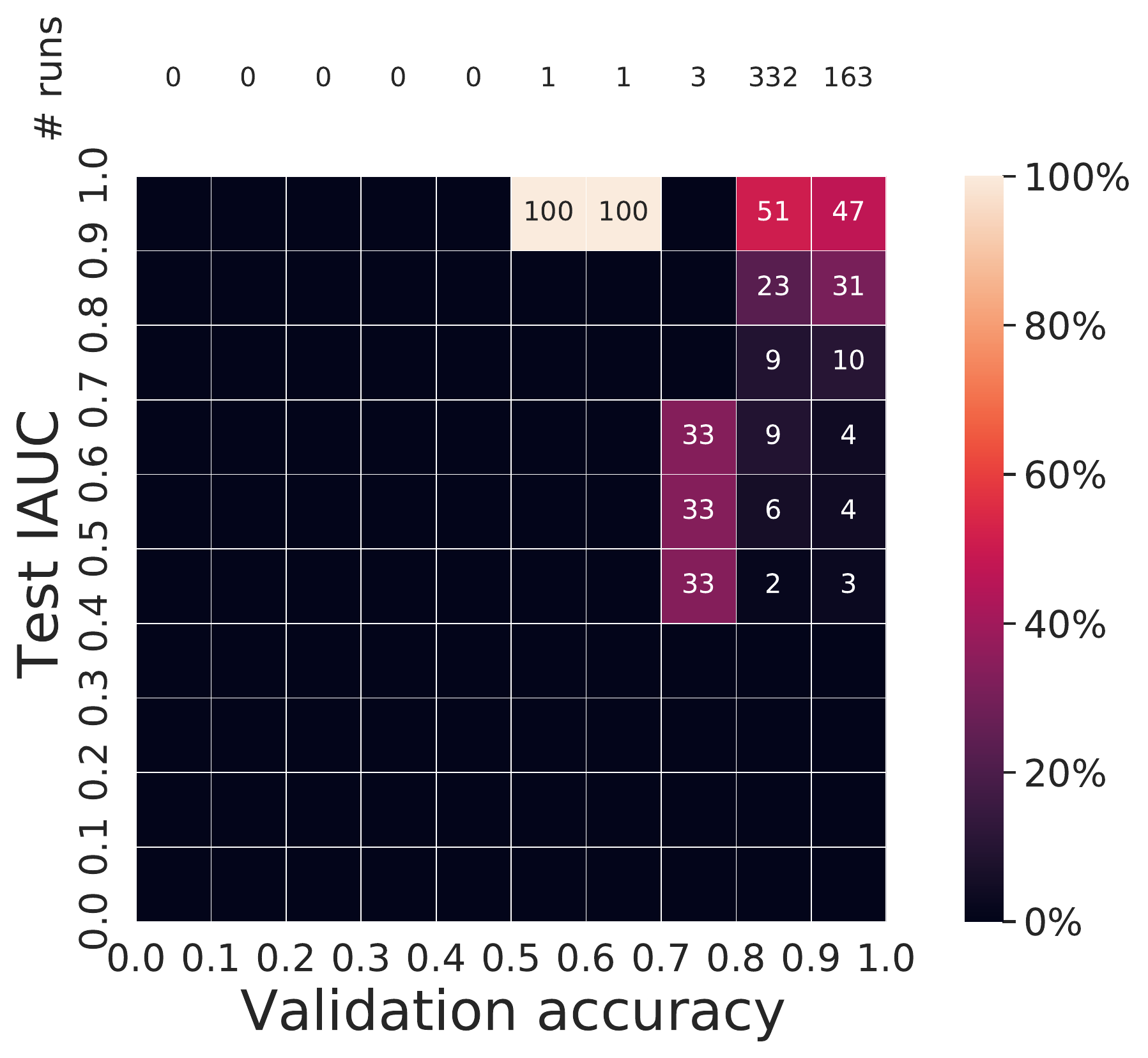}
\caption{MNIST XOR\label{fig:acc_vs_auroc_mnist_xor}}
\end{subfigure}
\begin{subfigure}{0.31\textwidth}
\includegraphics[width=\textwidth]{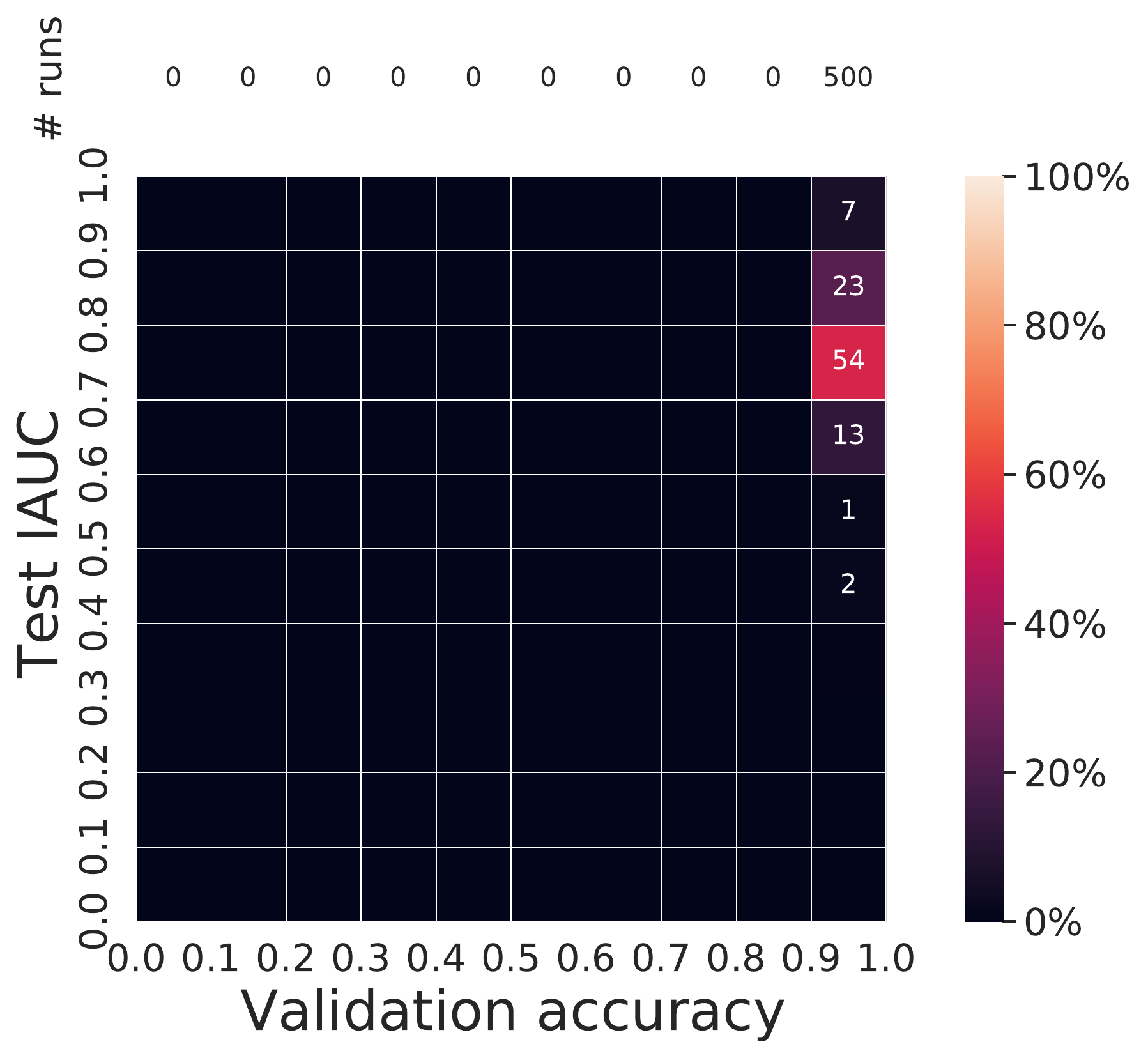}
\caption{CyTOF MIL\label{fig:acc_vs_auroc_sc_mil}}
\end{subfigure}
\begin{subfigure}{0.31\textwidth}
\includegraphics[width=\textwidth]{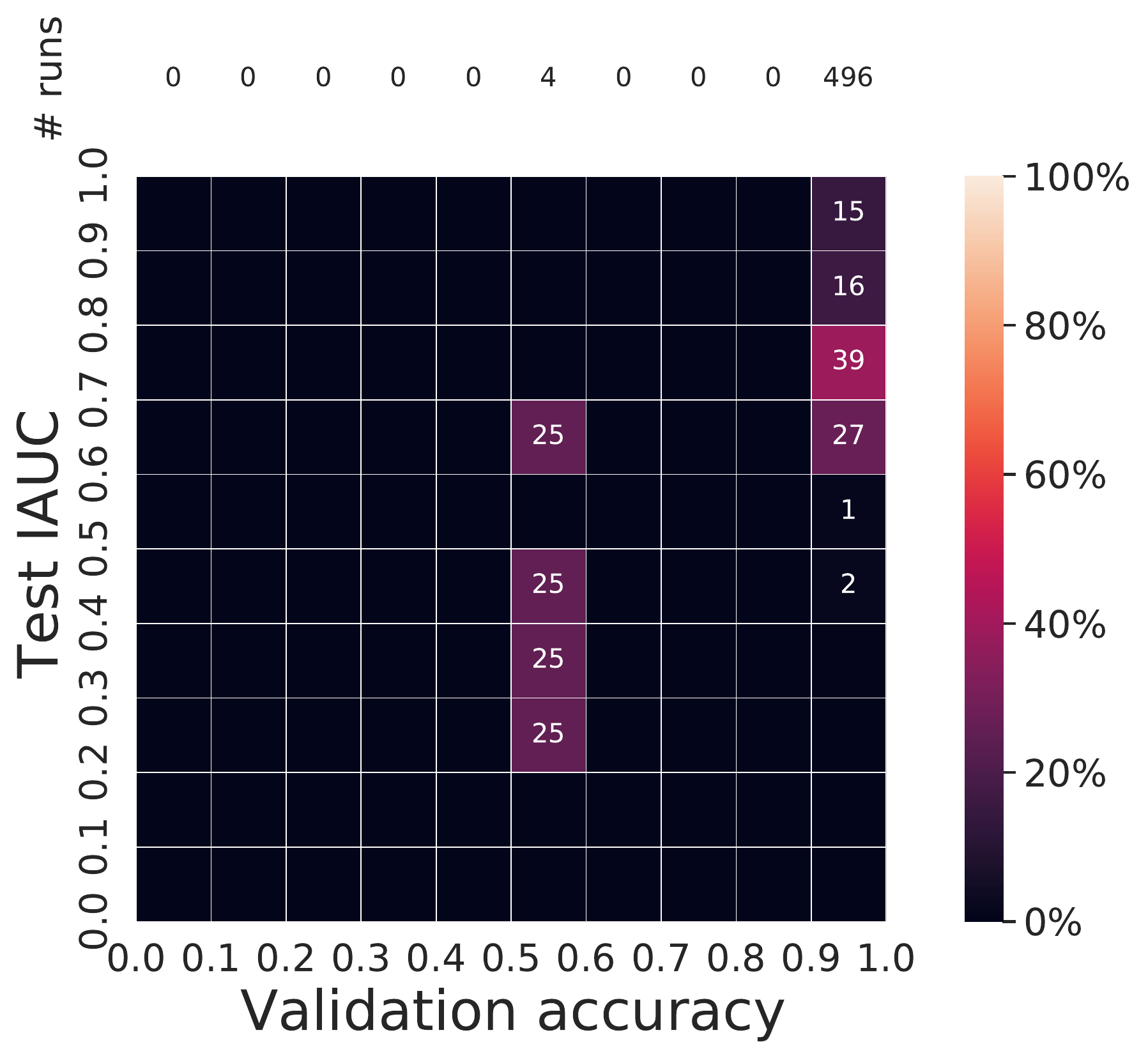}
\caption{CyTOF AND\label{fig:acc_vs_auroc_sc_and}}
\end{subfigure}
\begin{subfigure}{0.31\textwidth}
\includegraphics[width=\textwidth]{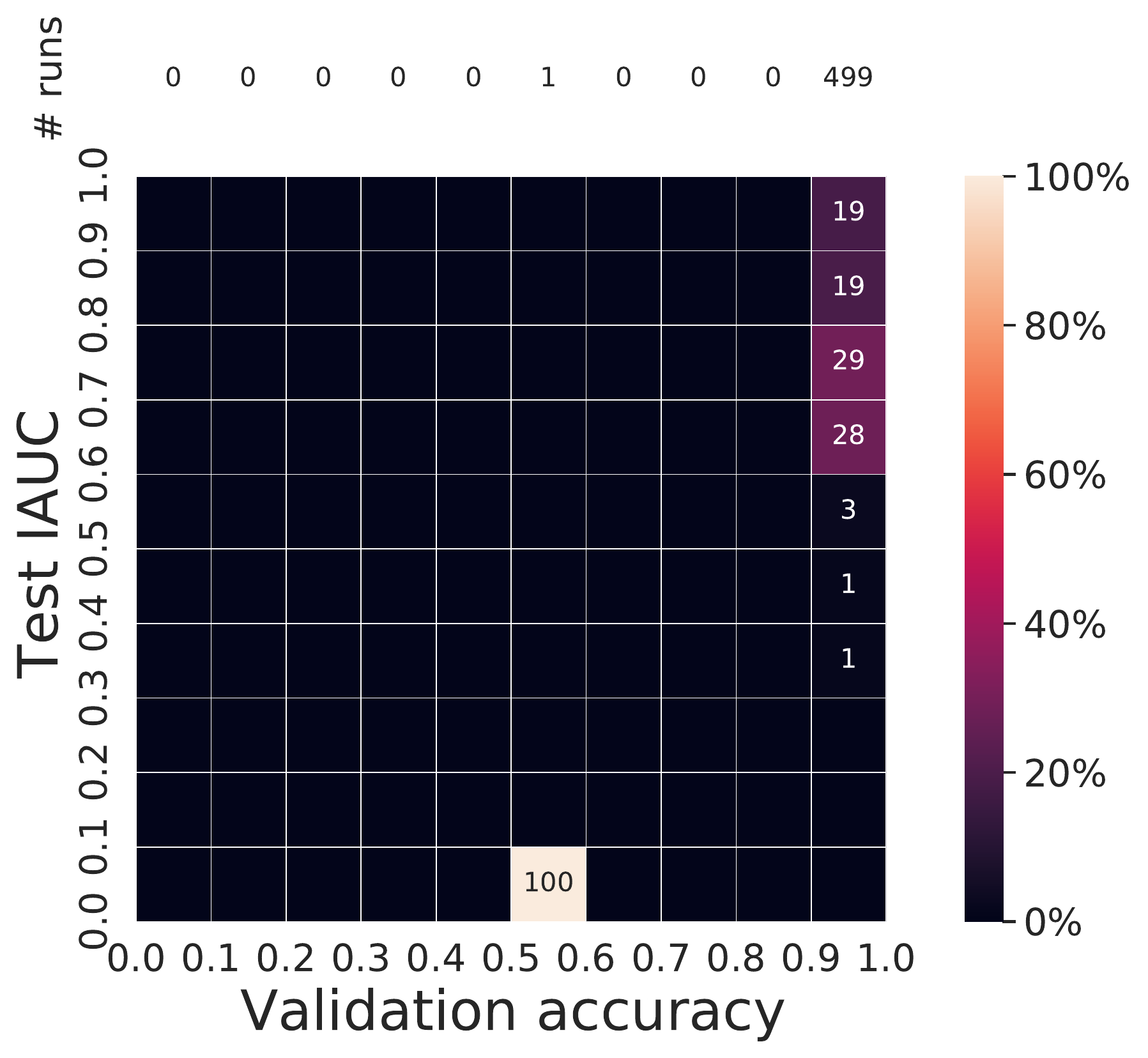}
\caption{CyTOF XOR\label{fig:acc_vs_auroc_sc_xor}}
\end{subfigure}
\end{multicols}

\caption{Relationship between validation accuracy and test IAUC for top configurations, separated by problem and data modality.
Models are binned by validation accuracy and IAUC and each bin displays the fraction of total models \textit{per column} (\textit{i.e.} per accuracy bin). The total number of models in each column is reported at the top.
\label{fig:acc_vs_auroc_top_config_detail}}
\end{figure*}


\twocolumn
\section{Ensembling}
\label{sec:ensembles-bad-prop}
Proportion of bad ensembles for single- and multi-configuration ensembles.
Bad ensembles are characterised by an IAUC of 0.65 or below. 
For each ensemble size, 30 different ensembles were produced.
In the single-configuration plots, the light grey lines show the results for the individual configurations while the black line shows their average. 
In the multi-configuration case, the process was repeated five times.
The 95\% confidence interval is indicated by the grey area.
\begin{figure}[hb]
\centering
\begin{subfigure}{0.22\textwidth}
\centering
\includegraphics[width=\textwidth]{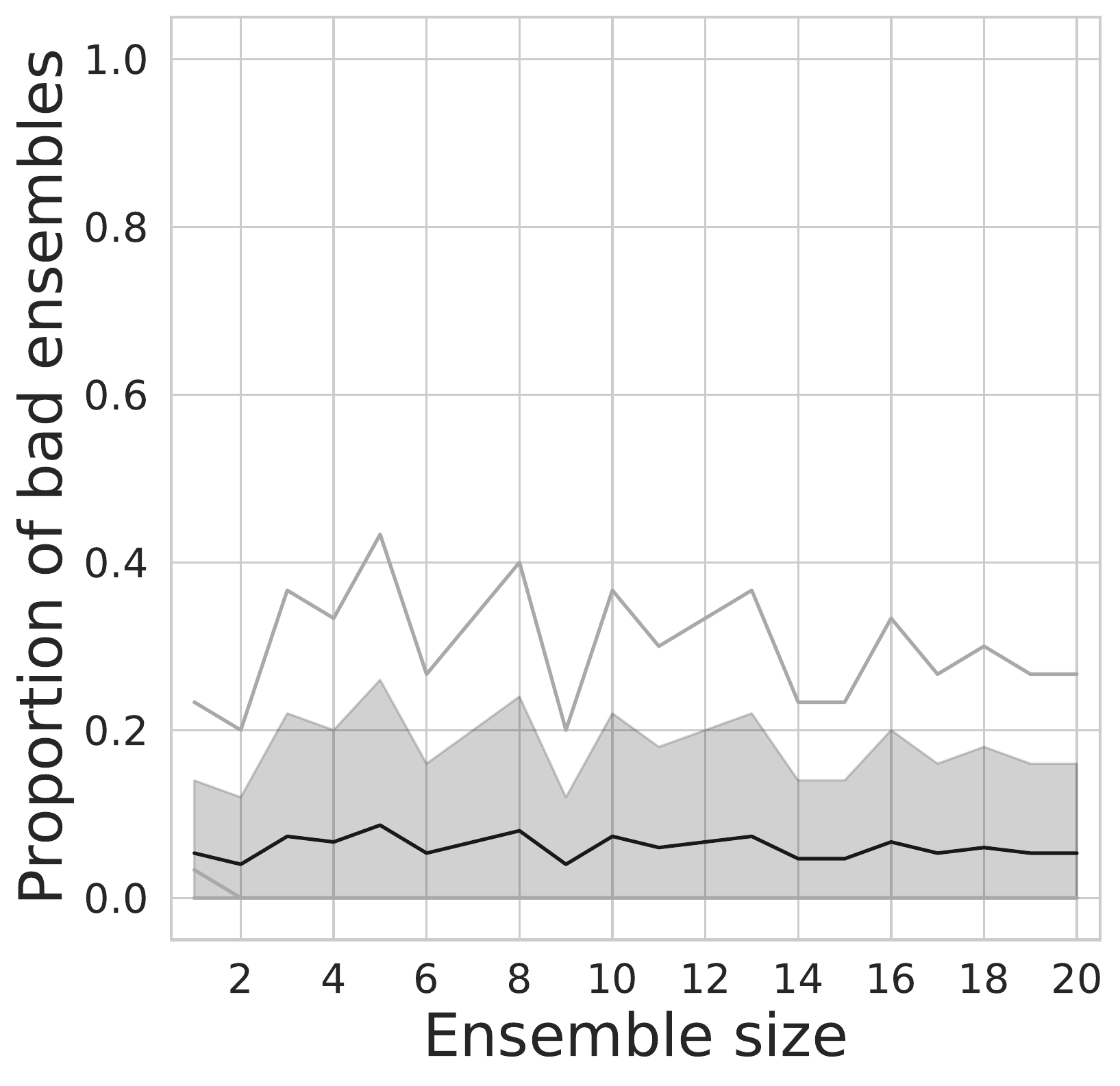}
\caption{Single-Configuration}
\end{subfigure}
\begin{subfigure}{0.22\textwidth}
\centering
\includegraphics[width=\textwidth]{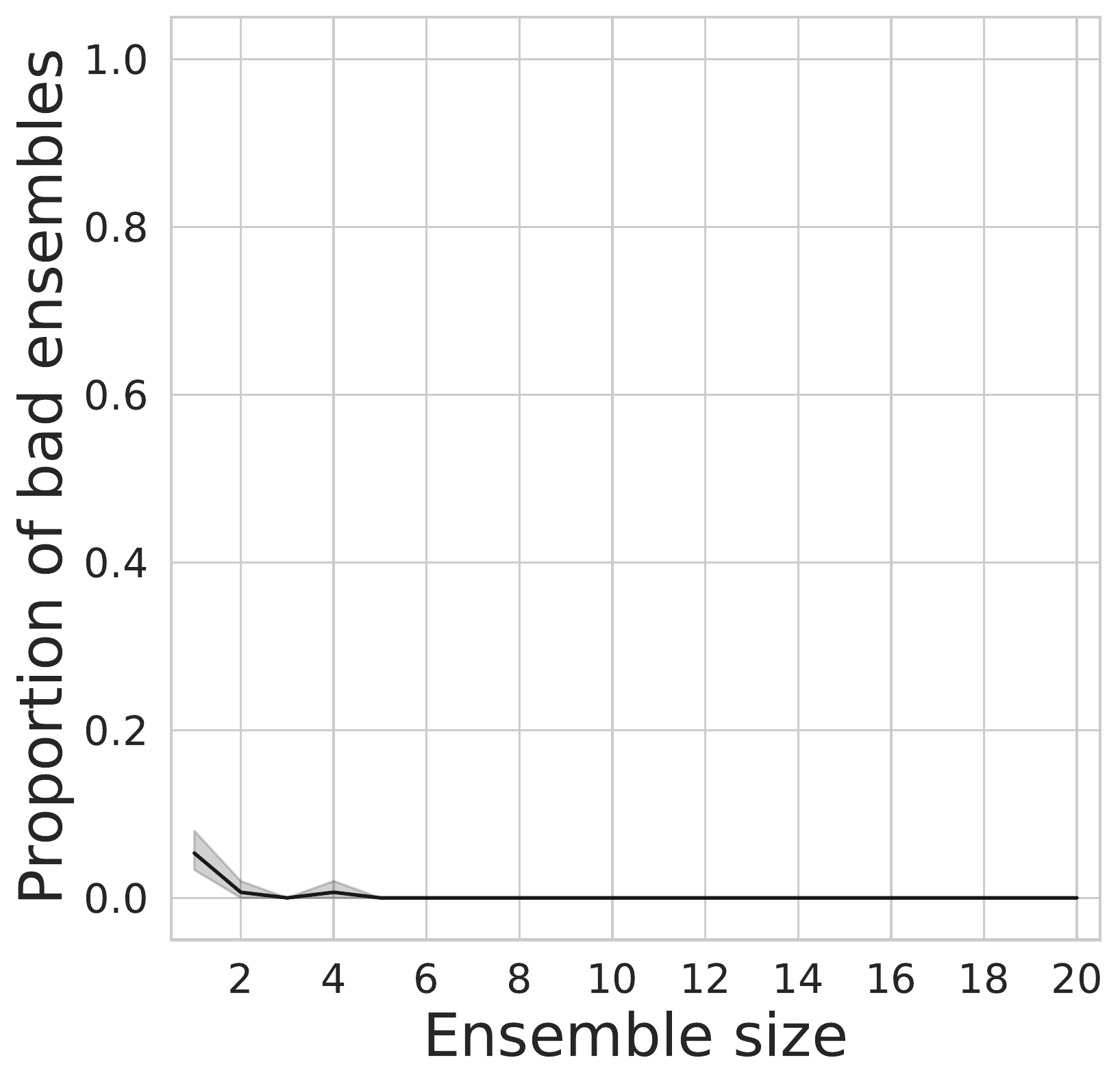}
\caption{Multi-Configuration}
\end{subfigure}

\caption[]{Gaussian MIL\label{fig:gaussian-mil-ensembles} }
\end{figure}

\begin{figure}[hb]
\centering
\begin{subfigure}{0.22\textwidth}
\centering
\includegraphics[width=\textwidth]{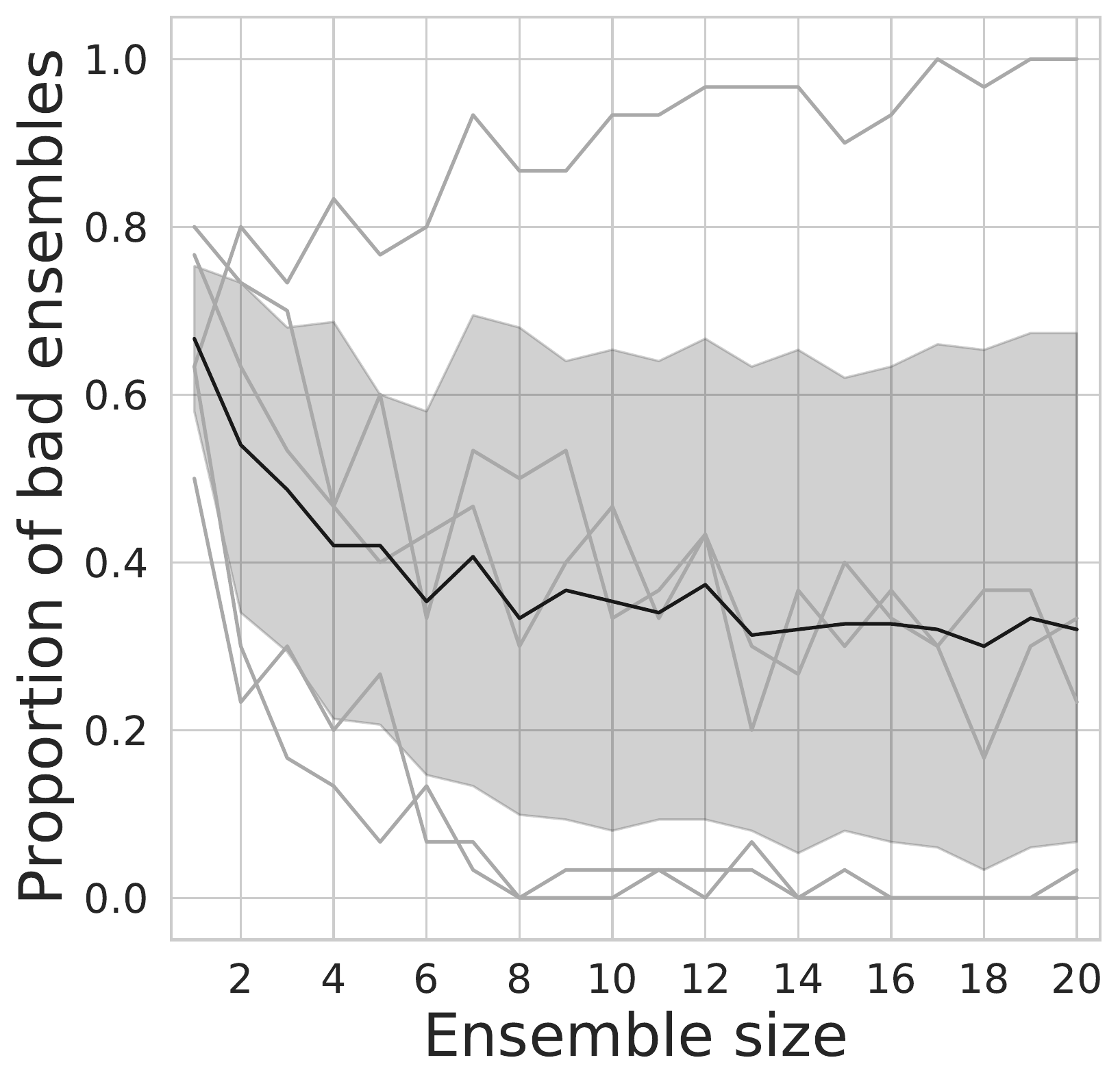}
\caption{Single-Configuration}
\end{subfigure}
\begin{subfigure}{0.22\textwidth}
\centering
\includegraphics[width=\textwidth]{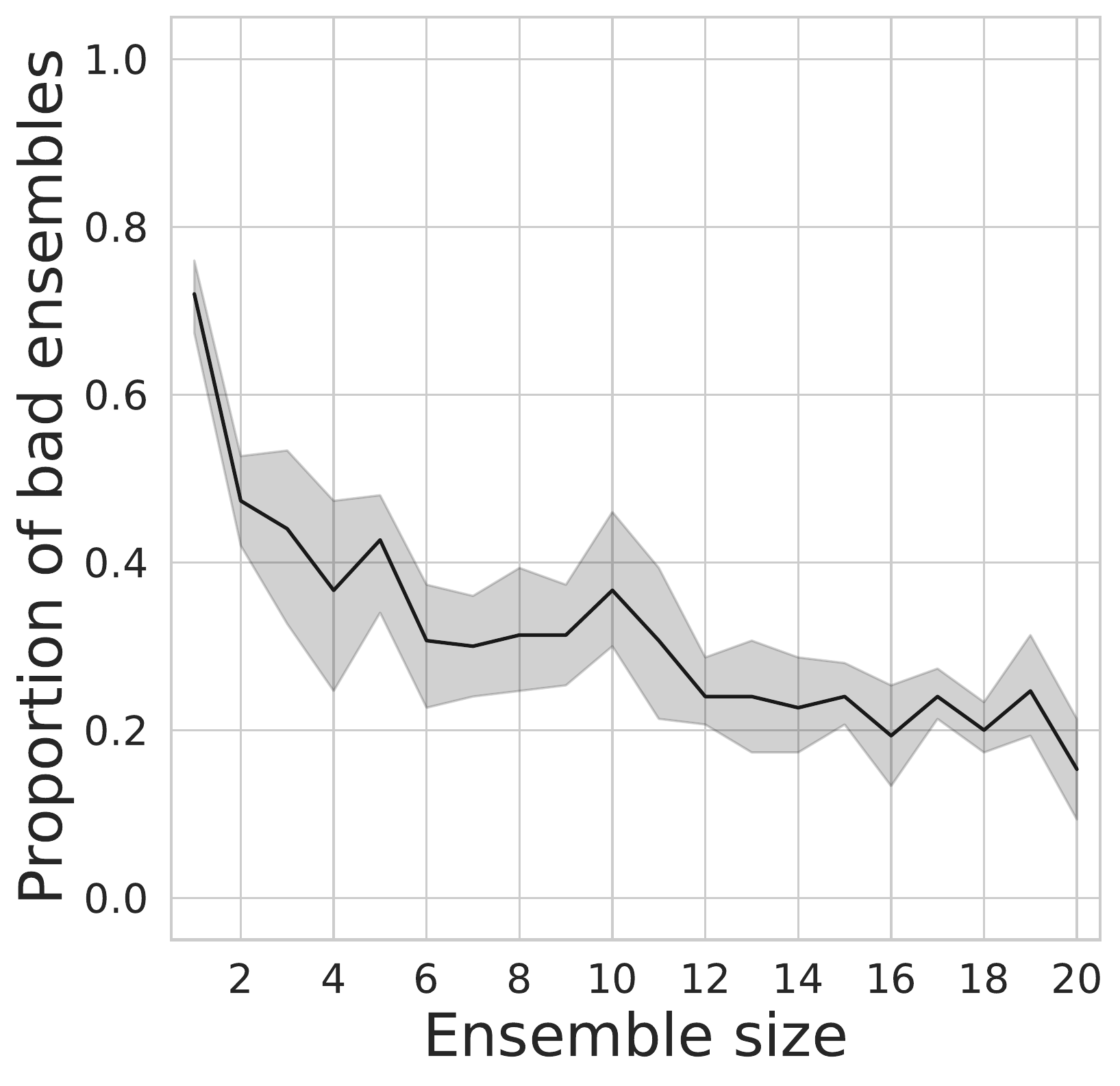}
\caption{Multi-Configuration}
\end{subfigure}

\caption[]{Gaussian AND \label{fig:gaussian-and-ensembles} }
\end{figure}

\begin{figure}[hb]
\centering
\begin{subfigure}{0.22\textwidth}
\centering
\includegraphics[width=\textwidth]{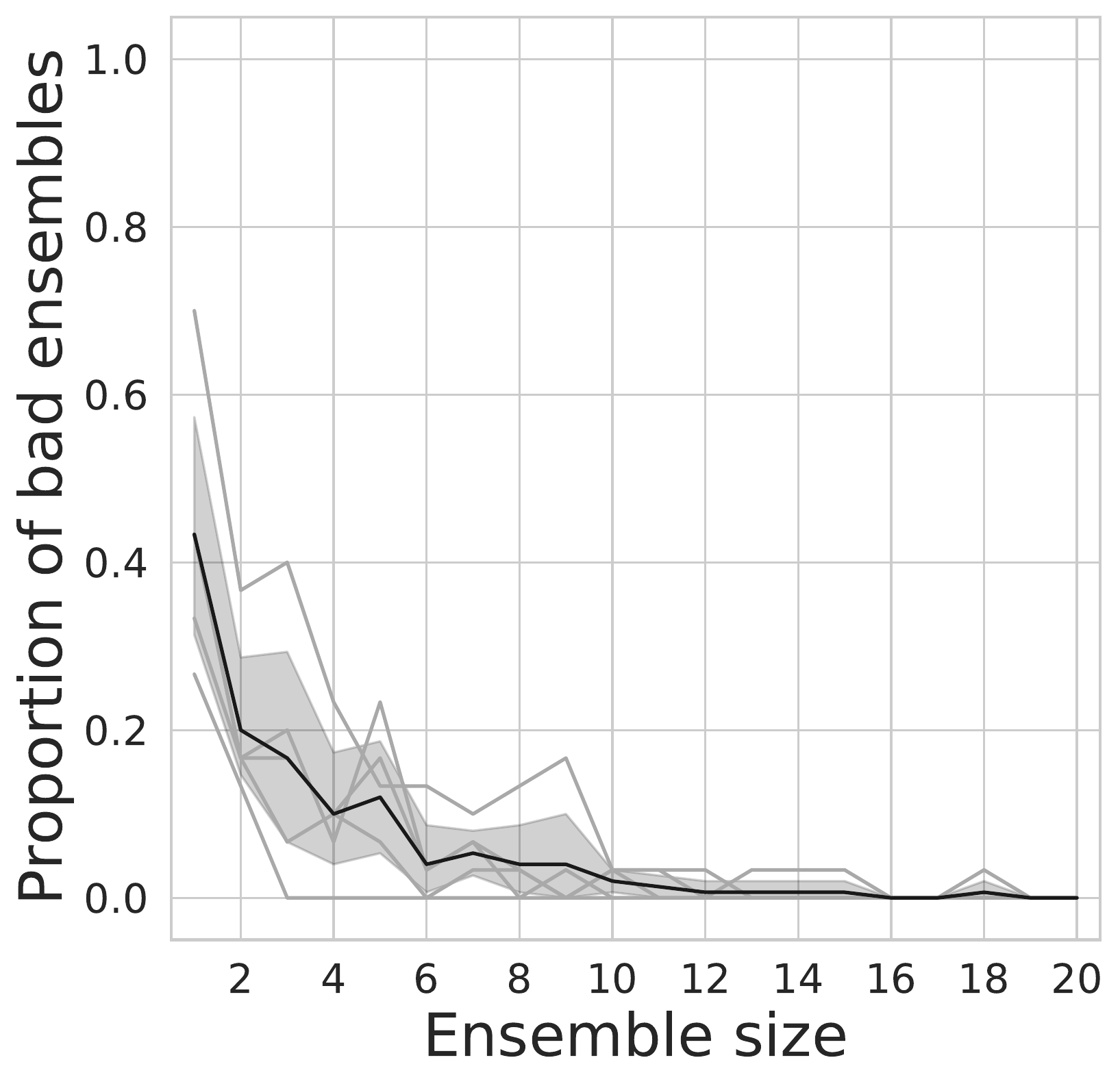}
\caption{Single-Configuration}
\end{subfigure}
\begin{subfigure}{0.22\textwidth}
\centering
\includegraphics[width=\textwidth]{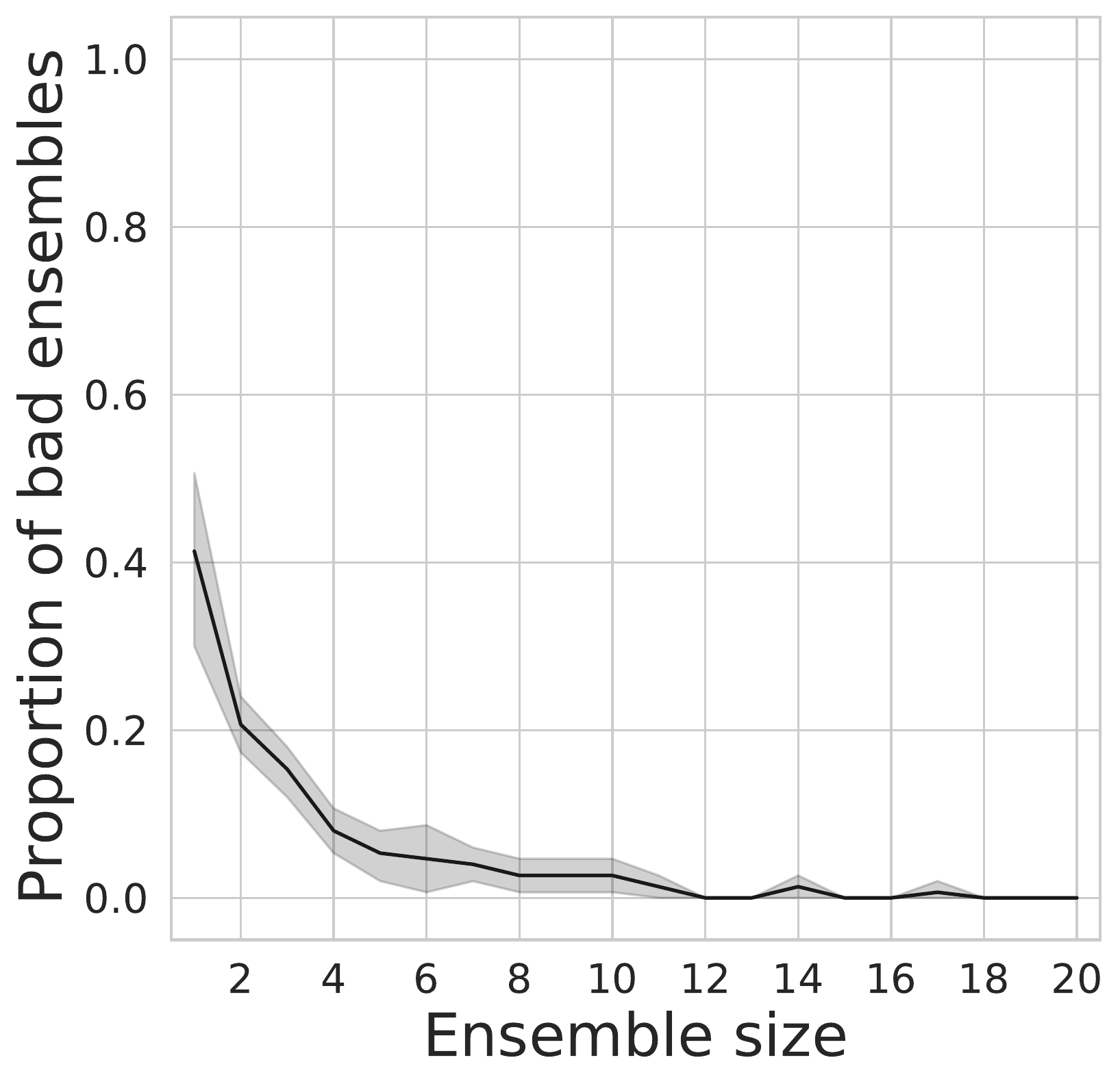}
\caption{Multi-Configuration}
\end{subfigure}

\caption[]{Gaussian XOR \label{fig:gaussian-and-ensembles} }
\end{figure}

\begin{figure}[hb]
\centering
\begin{subfigure}{0.22\textwidth}
\centering
\includegraphics[width=\textwidth]{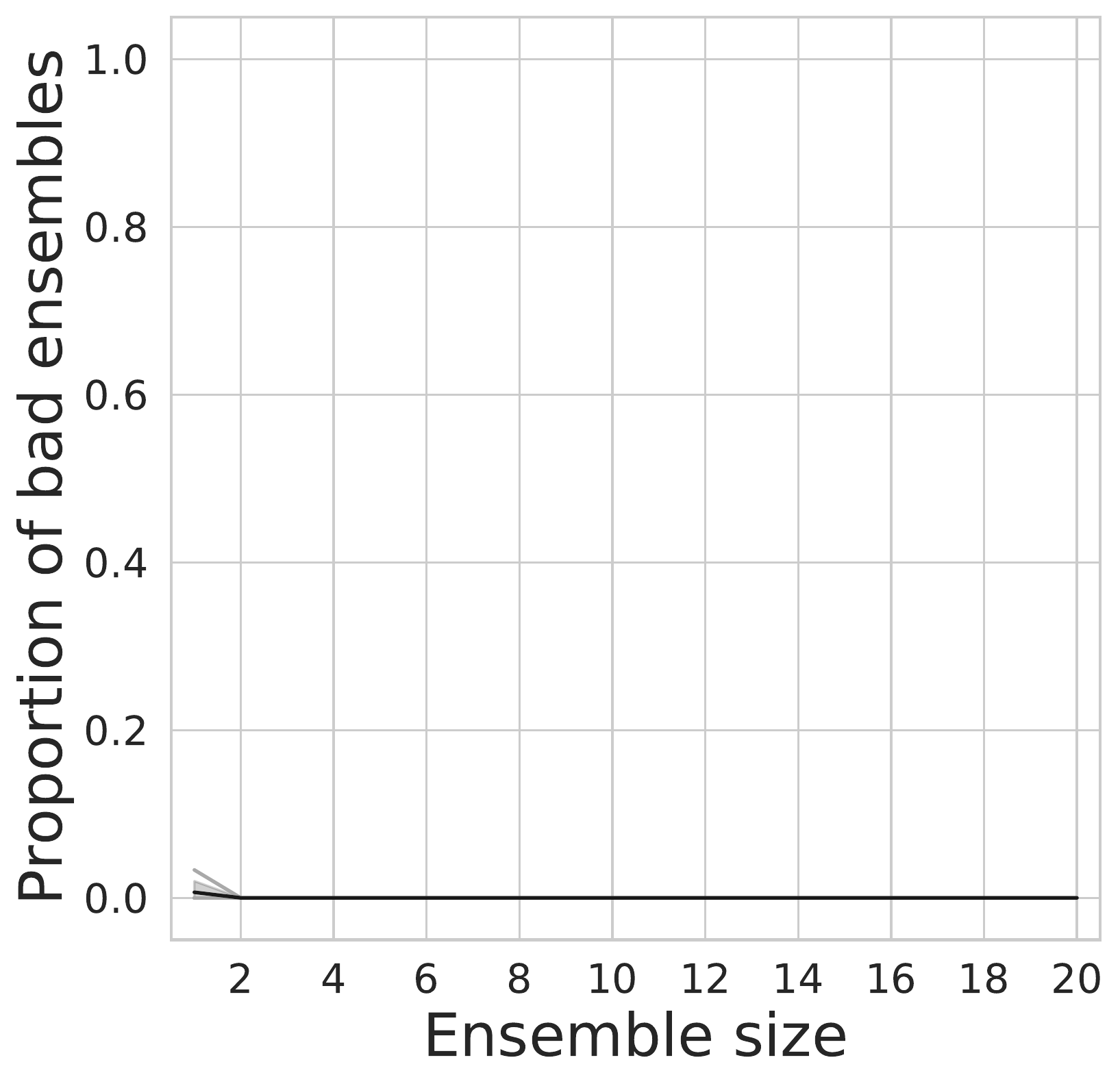}
\caption{Single-Configuration}
\end{subfigure}
\begin{subfigure}{0.22\textwidth}
\centering
\includegraphics[width=\textwidth]{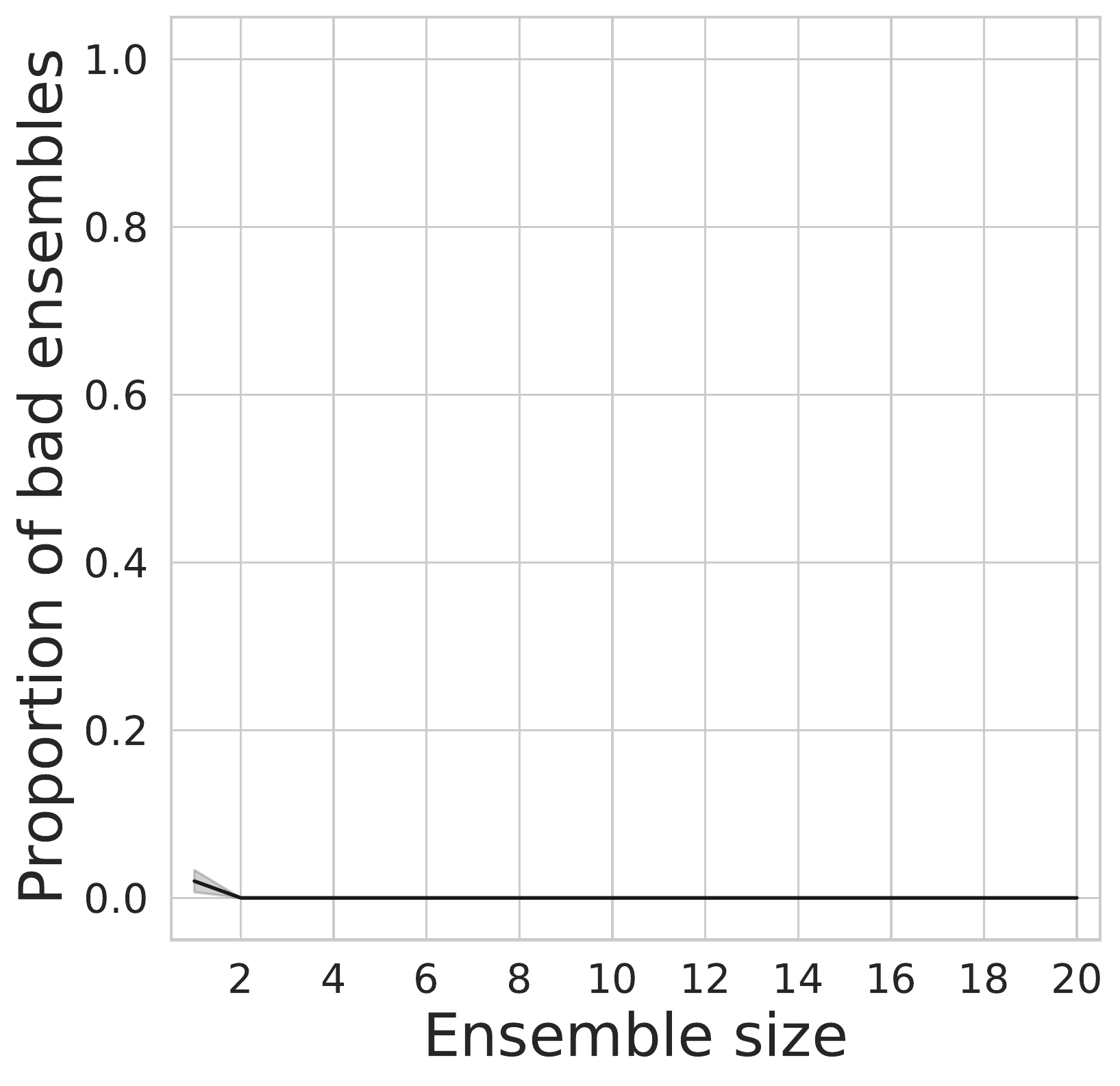}
\caption{Multi-Configuration}
\end{subfigure}

\caption[]{MNIST MIL \label{fig:gaussian-and-ensembles} }
\end{figure}

\begin{figure}[hb]
\centering
\begin{subfigure}{0.22\textwidth}
\centering
\includegraphics[width=\textwidth]{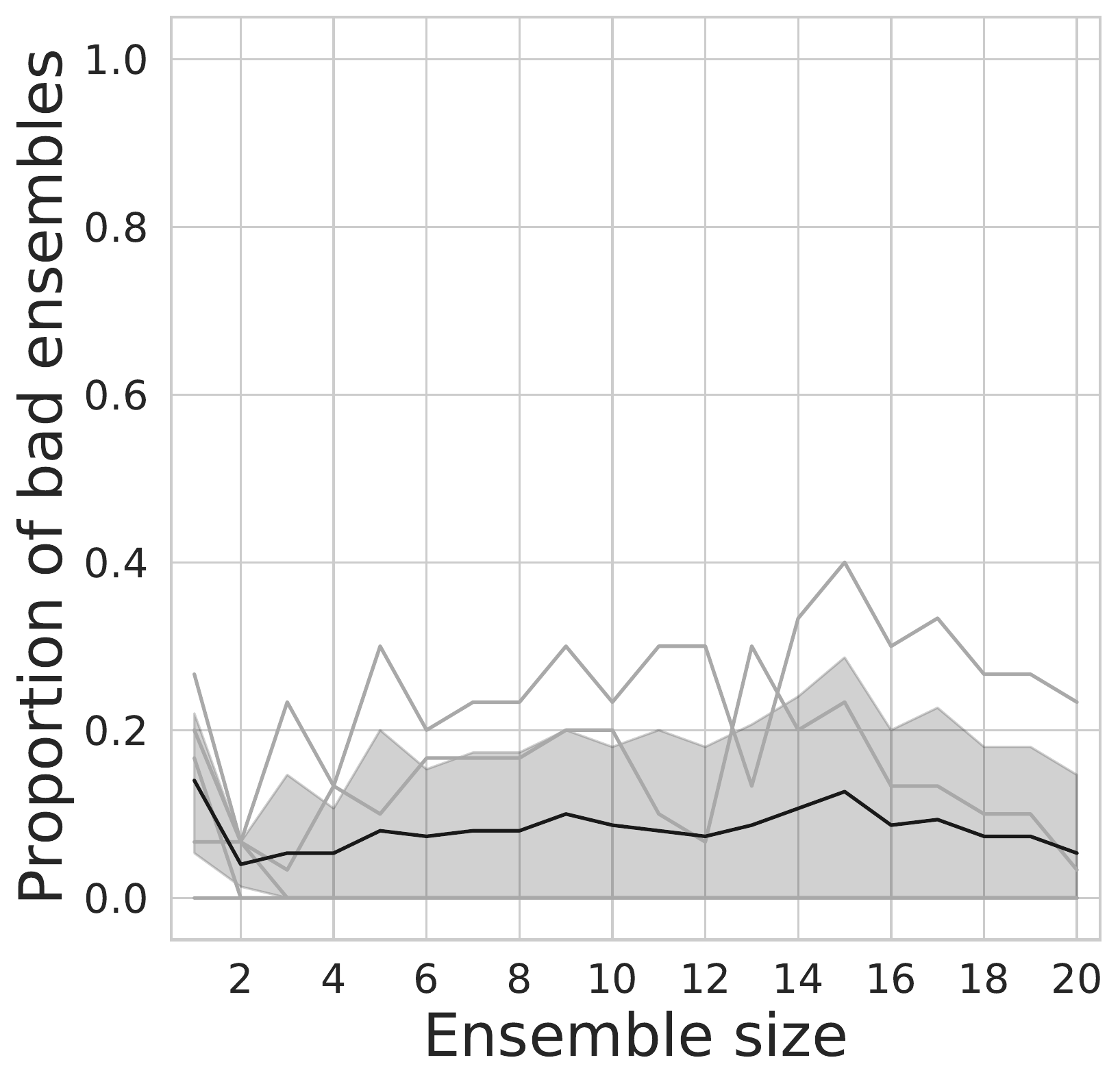}
\caption{Single-Configuration}
\end{subfigure}
\begin{subfigure}{0.22\textwidth}
\centering
\includegraphics[width=\textwidth]{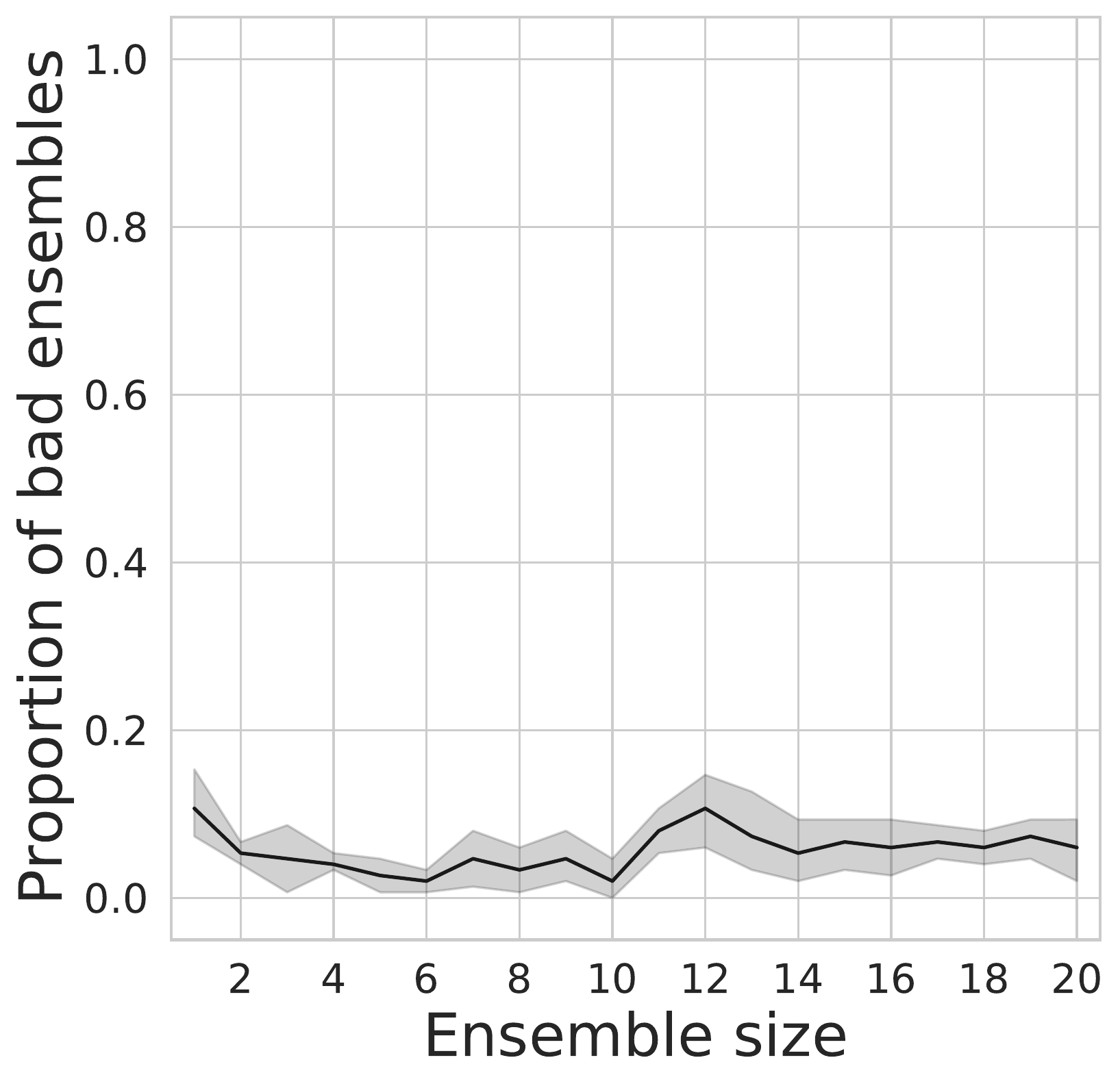}
\caption{Multi-Configuration}
\end{subfigure}

\caption[]{MNIST AND \label{fig:gaussian-and-ensembles} }
\end{figure}

\begin{figure}[hb]
\centering
\begin{subfigure}{0.22\textwidth}
\centering
\includegraphics[width=\textwidth]{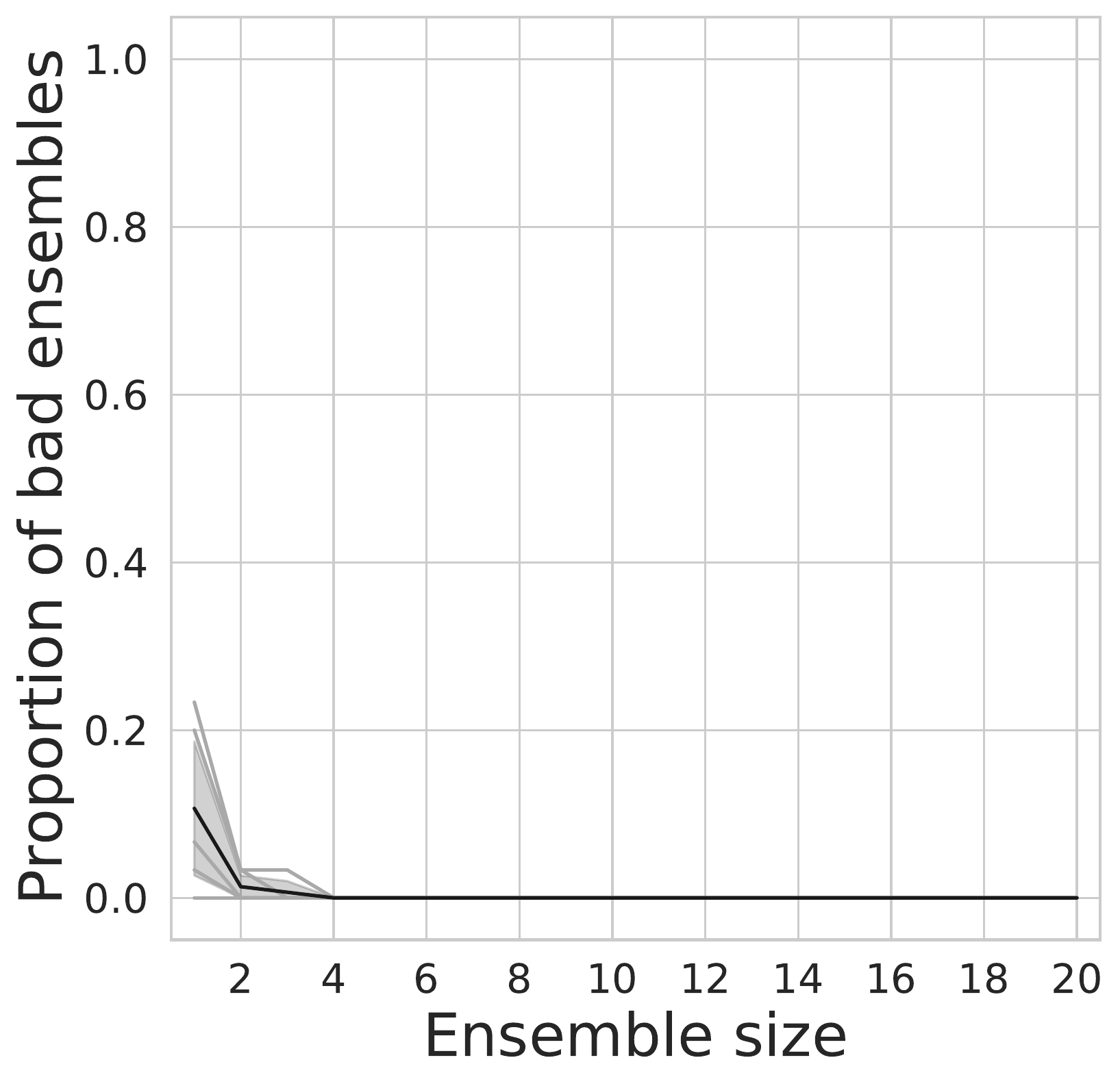}
\caption{Single-Configuration}
\end{subfigure}
\begin{subfigure}{0.22\textwidth}
\centering
\includegraphics[width=\textwidth]{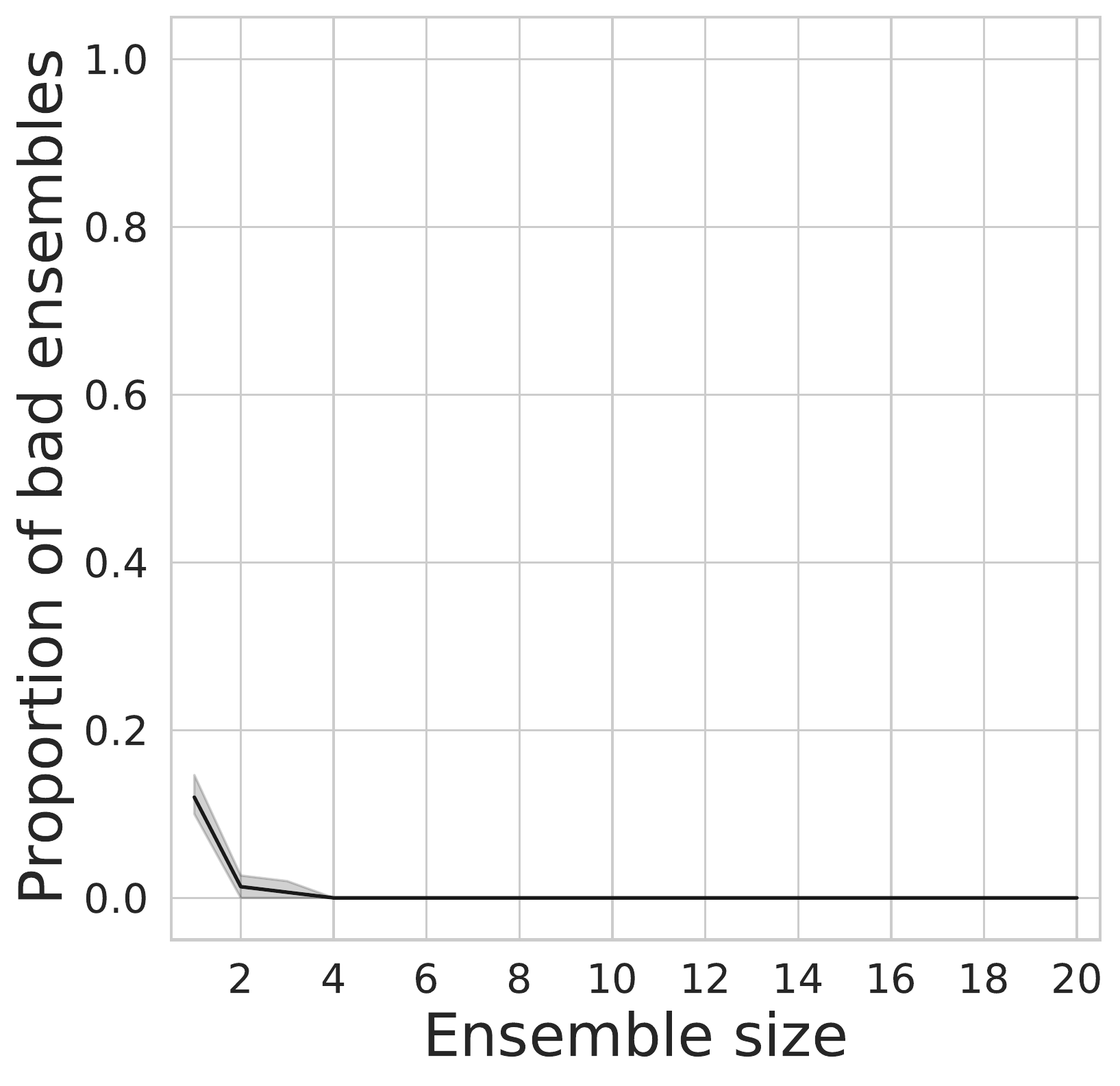}
\caption{Multi-Configuration}
\end{subfigure}

\caption[]{MNIST XOR \label{fig:gaussian-and-ensembles} }
\end{figure}


\end{document}